\documentclass[iicol]{sn-jnl}

\usepackage{graphicx}%
\usepackage{multirow}%
\usepackage{amsmath,amssymb,amsfonts}%
\usepackage{amsthm}%
\usepackage{mathrsfs}%
\usepackage[title]{appendix}%
\usepackage{xcolor}%
\usepackage{textcomp}%
\usepackage{manyfoot}%
\usepackage{booktabs}%
\usepackage{algorithm}%
\usepackage{algorithmicx}%
\usepackage{algpseudocode}%
\usepackage{listings}%
\usepackage{mathtools}
\usepackage{nicefrac}       
\usepackage{microtype}      
\usepackage{xcolor}         
\definecolor{rblue}{rgb}{0,0.5,1}

\usepackage{graphicx}

\usepackage{algorithm}
\usepackage{algorithmicx}
\usepackage[rightComments=true,italicComments=true,beginComment=//~]{algpseudocodex}

\algnewcommand{\IIf}[1]{\State\algorithmicif\ #1\ \algorithmicthen}
\algnewcommand{\EndIIf}{\unskip\ \algorithmicend\ \algorithmicif}

\definecolor{darkgray}{gray}{0.3}

\newcommand{\sComment}[1]{\Comment{\small #1}}

\usepackage{amsmath}
\usepackage{booktabs}
\usepackage{multirow}
\usepackage{colortbl}
\usepackage{booktabs}
\usepackage{hhline}
\usepackage{subcaption}
\usepackage{tikz}
\usepackage{lipsum}
\usepackage{wrapfig}
\usepackage{lipsum}  
\usepackage{paralist}
\usepackage{color}
\usepackage{rotating}
\usepackage{booktabs}

\definecolor{mydarkblue}{rgb}{0,0.08,0.45}
\definecolor{myfavblue}{rgb}{0.1176, 0.392, 1.0}
\usepackage{nicefrac}

\usepackage{listings}

\definecolor{dkgreen}{rgb}{0,0.6,0}
\definecolor{gray}{rgb}{0.5,0.5,0.5}
\definecolor{mauve}{rgb}{0.58,0,0.82}
\definecolor{lightgray}{HTML}{DDDDDD}

\usetikzlibrary{calc}
\usepackage{longtable}

\usepackage{amsmath}

\DeclareMathOperator*{\argmin}{arg\,min}
\newcommand{\repeatcommand}[2]{%
  \ifnum#1>0
    #2%
    \repeatcommand{\numexpr#1-1\relax}{#2}%
  \fi
}

\usepackage[english]{babel}
\usepackage{blindtext}

\theoremstyle{thmstyleone}%
\theoremstyle{thmstyletwo}%

\theoremstyle{thmstylethree}%

\usepackage{xcolor}

\definecolor{rblue}{rgb}{0,0.5,1}
\definecolor{awesome}{rgb}{1.0, 0.13, 0.32}
\definecolor{hollywoodcerise}{rgb}{0.96, 0.0, 0.63}
\definecolor{lasallegreen}{rgb}{0.03, 0.47, 0.19}
\definecolor{hanpurple}{rgb}{0.32, 0.09, 0.98}
\definecolor{green(pigment)}{rgb}{0.0, 0.65, 0.31}

\hypersetup{colorlinks=true,linkcolor={red},citecolor={hanpurple},urlcolor={magenta}}

\begin{document}

\title[Exploring Open-Set Domain Generalization Under Noisy Label]{Mitigating Label Noise using Prompt-Based Hyperbolic Meta-Learning in Open-Set Domain Generalization}


\author*[1,]{\fnm{Kunyu} \sur{Peng}}\email{kunyu.peng@kit.edu}
\author[1]{\fnm{Di} \sur{Wen}}\email{di.wen@kit.edu}
\author[1]{\fnm{M. Saquib} \sur{Sarfraz}}\email{saquib.sarfraz@kit.edu}
\author[1]{\fnm{Yufan} \sur{Chen}}\email{yufan.chen@kit.edu}
\author[1]{\fnm{Junwei} \sur{Zheng}}\email{junwei.zheng@kit.edu}
\author[1]{\fnm{David} \sur{Schneider}}\email{david.schneider@kit.edu}
\author*[2,]{\fnm{Kailun} \sur{Yang}}\email{kailun.yang@hnu.edu.cn}
\author[3]{\fnm{Jiamin} \sur{Wu}}\email{jiaminwu@cuhk.edu.hk}
\author[4]{\fnm{Alina} \sur{Roitberg}}\email{roitberg@uni-hildesheim.de}
\author[1]{\fnm{Rainer} \sur{Stiefelhagen}}
\email{rainer.stiefelhagen@kit.edu}
\affil[1]{\orgname{Karlsruhe Institute of Technology}, \city{Karlsruhe}, \country{Germany}}

\affil[2]{\orgname{Hunan University}, \city{Changsha}, \country{China}}
\affil[3]{\orgname{Shanghai AI Lab}, \city{Shanghai}, \country{China}}
\affil[4]{\orgname{University of Hildesheim}, \city{Hildesheim}, \country{Germany}}

\abstract{Open-Set Domain Generalization (OSDG) is a challenging task requiring models to accurately predict familiar categories while minimizing confidence for unknown categories to effectively reject them in unseen domains. While the OSDG field has seen considerable advancements, the impact of label noise—a common issue in real-world datasets—has been largely overlooked. Label noise can mislead model optimization, thereby exacerbating the challenges of open-set recognition in novel domains. In this study, we take the first step towards addressing Open-Set Domain Generalization under Noisy Labels (OSDG-NL) by constructing dedicated benchmarks derived from widely used OSDG datasets, including PACS and DigitsDG. We evaluate baseline approaches by integrating techniques from both label denoising and OSDG methodologies, highlighting the limitations of existing strategies in handling label noise effectively. To address these limitations, we propose HyProMeta, a novel framework that integrates hyperbolic category prototypes for label noise-aware meta-learning alongside a learnable new-category agnostic prompt designed to enhance generalization to unseen classes. Our extensive experiments demonstrate the superior performance of HyProMeta compared to state-of-the-art methods across the newly established benchmarks. The source code of this work is released at \url{https://github.com/KPeng9510/HyProMeta}.
}
\maketitle

\section{Introduction}
\label{sec1}

Open-Set Domain Generalization (OSDG) represents a critical generalization problem that combines the dual challenges of domain shift and category shift during the test phase. In this task, models are required to provide accurate predictions for categories encountered during training while assigning low confidence scores to unseen categories from new domains. OSDG is particularly relevant in real-world applications with diverse and dynamic conditions, such as healthcare~\cite{li2020domain}, security~\cite{busto2018open}, and autonomous driving~\cite{guo2022simt}, where new domains and categories frequently arise during deployment. Recent advancements in OSDG have often adopted meta-learning frameworks~\cite{wang2023generalizable, shu2021open}, where training involves simulating various cross-domain tasks to enhance generalization. These methods rely on precise a priori knowledge from source domains and known categories to learn discriminative embeddings that enable accurate predictions in new domains and new categories during testing.

The presence of label noise further exacerbates the challenges of OSDG, as it undermines the reliability of the a priori knowledge derived from the training set. Despite its importance, the issue of label noise in OSDG has been largely overlooked by the research community. While extensive efforts have been made to address label noise in general classification tasks, existing methods such as relabeling approaches~\cite{zhang2024badlabel,zheng2020error,li2024nicest}, data pruning techniques~\cite{kim2021fine,karim2022unicon,zhou2020robust,li2022selective}, and loss-based noise-agnostic methods~\cite{xu2024skeleton,lukasik2020does,zheng2021meta,yue2024ctrl} focus primarily on identifying and correcting noisy labels, pruning noisy datasets, or optimizing selectively based on loss values. However, these approaches do not account for the additional complexity of generalizing to unseen domains and distinguishing unseen categories, which are essential in OSDG.

To address this gap, we for the first time systematically investigate the impact of label noise on open-set domain generalization. In this work, we introduce two novel benchmarks for OSDG under noisy label settings, based on the widely-used PACS~\cite{li2017deeper} and DigitsDG~\cite{zhou2020deep} datasets. We evaluate a wide range of well-established methods from both the OSDG and noisy label learning fields as baselines. Our experiments reveal that these baseline approaches face significant limitations in this new setting. Specifically, OSDG methods are highly sensitive to label noise, while noisy label learning approaches struggle to handle the large domain shifts and label shifts inherent in the train-test split of the OSDG task.

To overcome these challenges, we propose a novel method, named HyProMeta, specifically designed to address the problem of open-set domain generalization under label noise. 
\begin{figure}[!t]
\centering 
\includegraphics[width=0.49\textwidth]{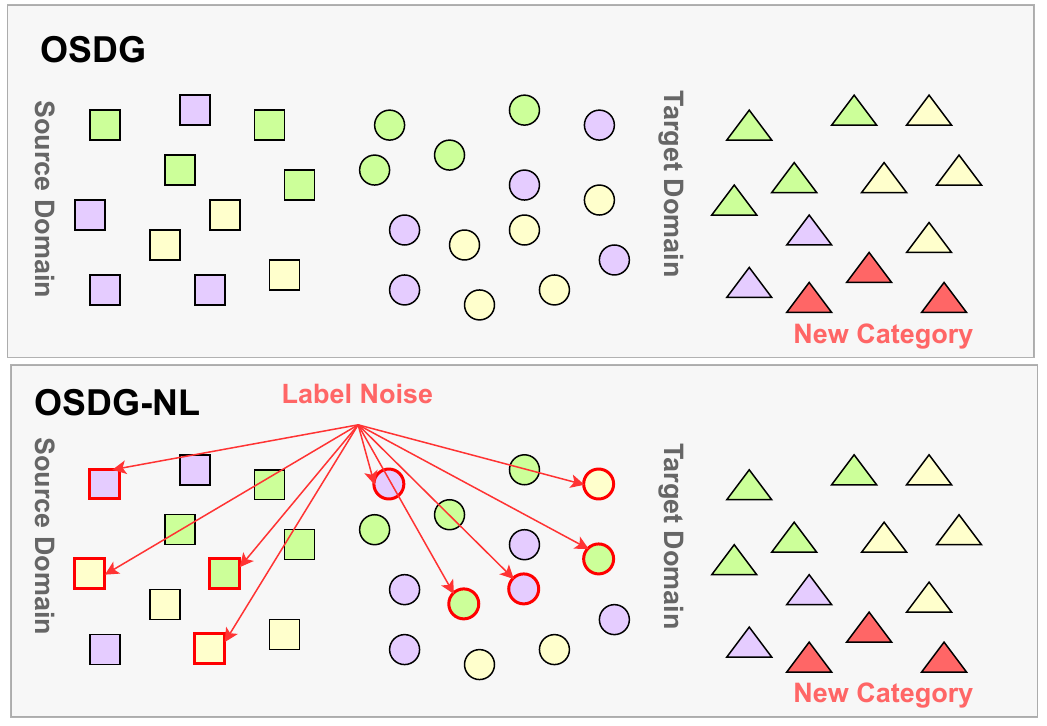}

\caption{Illustration of the Open-Set Domain Generalization under Noisy Label (OSDG-NL) task. Different shapes (\textit{e.g.}, circles, triangles, squares) symbolize distinct domains, while their respective colors represent various categories. A subset of these samples, outlined in red, indicates the presence of label noise. The objective of this task is to train a model capable of leveraging robust priors derived from source domains, despite label noise, to achieve precise classification of known categories while assigning low confidence scores to novel, unseen categories in a new target domain.
}
\label{fig:teaser}
\end{figure} 

HyProMeta incorporates a meta-learning framework designed to leverage samples with relatively clean labels alongside those potentially containing label noise, aiming to minimize losses on in-domain and cross-domain samples identified as clean together with the new corrected in-domain samples which are recognized with label noise. To estimate clean and noisy sample sets, categorical cluster centers in hyperbolic space are utilized to achieve effective separation, facilitating the meta-learning process. Samples identified as noisy are relabeled based on their nearest cluster center within the hyperbolic space.
To address challenges posed by out-of-distribution samples within one specific known category, where hard samples may exhibit significant hyperbolic uncertainty relative to their corresponding prototypes and confuse the label noise judgment, we introduce a learnable prompt for prompt-based mixed category augmentation to enhance model generalization towards new categories.
The learnable prompt, with dimensions identical to the input images, is incorporated into a data augmentation strategy. Specifically, two samples from different classes are mixed, and a fixed-size window on the mixed image is randomly selected and replaced with the corresponding region from the learnable prompt. We do not directly use similarity-based metric learning within the sample domain and category to avoid the model from overfitting on samples with label noise.
The resulting augmented sample is then classified as an additional category to represent out-of-distribution data, which aims at implicitly improving the generalizability of the model on unseen samples by delivering low confidence scores on the existing categories.

Our approach demonstrates state-of-the-art performance on two newly constructed benchmarks, highlighting its effectiveness in addressing noisy label learning across diverse domains.

The contributions of this paper are summarized as follows:
\begin{itemize} 
\item To the best of our knowledge, this is the first work to explore the task of Open-Set Domain Generalization under Noisy Labels (OSDG-NL). We establish two benchmarks for this purpose, leveraging methodologies from both the open-set domain generalization and label noise learning paradigms, using PACS~\cite{li2017deeper} and DigitsDG~\cite{zhou2020deep} datasets. 
\item To effectively tackle the challenges inherent to OSDG-NL, we propose a novel approach named HyProMeta, which employs hyperbolic category prototypes for label noise-aware meta-learning and introduces new category-aware prompt learning to enhance the model's generalizability to unseen categories and assist the meta-learning. 
\item The proposed HyProMeta approach achieves state-of-the-art performances on the two constructed benchmarks, demonstrating high generalizability across varying label noise ratios, noise types, and backbone architectures. 
\end{itemize}

\section{Related Work}
\label{sec2}

\subsection{Open-Set Domain Generalization}
Open-set domain generalization consists of two interconnected challenges targeting the generalizability of deep learning models, which are the domain generalization challenge and the open-set recognition challenge. 
Open-set recognition focuses on rejecting instances of unknown categories during inference by assigning them low confidence scores~\cite{geng2020recent,peng2024navigating} while domain generalization involves training a deep learning model on multiple source domains and enabling it to generalize effectively to unseen domains at test time~\cite{sun2016deep,wang2022generalizing}.

Different approaches are adopted to deal with these two different data shifts existing in the training and testing sets respectively. Open-set recognition is well-explored by the community through the proposal of techniques such as evidential learning~\cite{wang2024towards, zhao2023open, bao2021evidential}, logits calibration~\cite{peng2024navigating}, GAN-based approaches~\cite{kong2021opengan}, reconstruction-based methods~\cite{yoshihashi2019classification, huang2022class}, and reciprocal point-based models~\cite{chen2021adversarial, chen2020learning}, while domain generalization strategies aim to bridge the domain gap using a variety of techniques, including data augmentation~\cite{wang2020heterogeneous, nam2021reducing, zhou2020domain, guo2023aloft, zhou2020learning, li2021progressive, li2021simple}, domain-specific normalization~\cite{seo2020learning}, domain adversarial training~\cite{ganin2016domain}, and GAN-based methods~\cite{chen2022discriminative,li2024vocoder}.  

When tackling the aforementioned two challenges at the same time in open-set domain generalization, most of the research works explored how to achieve good generalizability towards different domains and adopted settings where the known categories are not equally distributed in each source domain~\cite{fu2020learning, singha2024unknown, bose2023beyond, chen2021adversarial, li2018deep, zhao2022adaptive}. 
Among them, Katsumata~\textit{et al.}~\cite{katsumata2021open} presented a metric learning-based approach to create a discriminative embedding space, aiding open-set domain generalization. 
Domain-augmented meta-learning is proposed by Shu~\textit{et al.}~\cite{shu2021open} to tackle the open-set domain generalization problem by introducing more diverse data distributions during training time. Bose~\textit{et al.}~\cite{bose2023beyond} introduced ODG-Net, which leverages GANs to synthesize data from a union of training domains, improving cross-domain performance. 
Wang~\textit{et al.}~\cite{wang2023generalizable} made notable contributions by formalizing new Open-Set Domain Generalization (OSDG) protocols, where all source domains share the same predefined seen category set.
Within these new proposed protocols, meta-learning-based approaches~\cite{peng2024advancing,wang2023generalizable} show promising performances when dealing with OSDG tasks.
In this work, we adopt the same OSDG settings proposed by Wang~\textit{et al.}~\cite{wang2023generalizable} on the leveraged DigitsDG and PACS datasets.
Our proposed new method, HyProMeta, also adopts meta-learning as the basic framework to deal with open-set domain generalization while incorporating the label noise agnostic learning enabled by hyperbolic space prototypes and new-category aware prompt learning to enhance the model's generalizability.

\subsection{Noisy Label Learning}
Optimizing models with accurate labels is crucial for most deep learning methods to enable them to learn precise a priori knowledge from the provided samples~\cite{xu2024skeleton}. 
In contrast, incorrect labels can misguide the optimization process~\cite{cheng2020weakly}. To address the negative impact of label noise, researchers have proposed various approaches from different perspectives. For instance, methods introduced in~\cite{xia2019anchor, tanno2019learning, zhu2021clusterability, zhu2022beyond, li2022estimating} aim to identify label noise by modeling the probability of label corruption. 
Other approaches, such as those in~\cite{song2019selfie, wei2021smooth, chen2021beyond}, focus on detecting label noise prior to training.

Alternatively, sample re-weighting techniques mitigate the influence of mislabeled samples by assigning lower weights to their loss values~\cite{liu2015classification}. Meta-learning-based methods~\cite{shu2019meta, wang2020training, zheng2021meta} have also shown promise in handling noisy labels by dynamically adapting the loss function and learning optimal training strategies.

Dynamic sample selection approaches adopt a semi-supervised framework to manage noisy data. These methods begin training with a subset of clean samples and incrementally incorporate mislabeled data in a controlled manner, enabling the model to adapt effectively over time~\cite{Han_Yao_Yu_Niu_Xu_Hu_Tsang_Sugiyama_2017, chen2019understanding, li2020dividemix}. For instance, TCL~\cite{huang2023twin} employs contrastive learning to develop robust representations and uses a Gaussian Mixture Model for label mapping. Similarly, PLM~\cite{zhao2024estimating} utilizes part-level labels to guide the model in extracting richer and more detailed information.

Several works have also explored noise-robust loss functions~\cite{liu2020peer, ma2020normalized, zhu2021second} to improve optimization in the presence of label noise. Other studies have focused on employing proper regularization techniques to minimize the adverse effects of noisy labels~\cite{wei2021open, cheng2021mitigating, liu2022robust}. For example, BadLabel~\cite{zhang2024badlabel} implements a label-flipping attack, creating indistinguishable loss values between clean and noisy labels. Meanwhile, LSL~\cite{kim2024learning} leverages structural label information to improve learning from noisy data.

Despite these advancements, most existing approaches for learning with label noise fail to address the challenges posed by domain shifts and label distribution shifts encountered at test time. This limitation presents significant robustness challenges for handling noisy labels in the field of Open-Set Domain Generalization (OSDG). In this work, we, for the first time, focus on the task of Open-Set Domain Generalization under Noisy Labels (OSDG-NL). We introduce two novel benchmarks based on the well-established OSDG datasets, namely PACS~\cite{li2017deeper} and DigitsDG~\cite{zhou2020deep}, by incorporating various levels of label noise. Additionally, we select diverse methods from the OSDG and label noise learning fields to serve as baselines. We further propose a novel approach, HyProMeta, which implements label noise-aware meta-learning and demonstrates superior performance compared to existing methods on the challenging OSDG-NL task.

\section{Benchmark}
\label{sec3}
In this section, we first provide a detailed description of the proposed OSDG-NL task in Section~\ref{sec:task_description}. The specific label noise settings are outlined in Section~\ref{sec:label_noise}. Subsequently, the datasets and baseline methods utilized in this study are presented in Section~\ref{sec:datasets} and Section~\ref{sec:baselines}, respectively.

\subsection{Task Description}
\label{sec:task_description}
In our task, various domains $\Omega_\mathcal{D} = \{\mathcal{D}_1, \mathcal{D}_2,..., \mathcal{D}_{N_S}\}$ are available, we follow the leave one out setting in~\cite{wang2023generalizable} by choosing $\mathcal{D}_m$ as the test domain, while the remaining domains from $\Omega_{\mathcal{D}_S} = \Omega_\mathcal{D}/\{\mathcal{D}_m\}$ serve as source domains during test phase. 
Let $\Omega_l$ denote the set of labels existing in a dataset. 
To achieve open-set recognition, $\Omega_u \subset \Omega_l$ are chosen as unseen categories during test phase, while $\Omega_k \subset \Omega_l$ are known during training, where $\Omega_l = \Omega_u \cup \Omega_k$. 
The sample from each source domain consists of data-label pairs $(\mathbf{x}_s, \mathbf{y}_s)$, where $\mathbf{x}_s$ denotes the data and $\mathbf{y}_s$ denotes the label. In the noisy label setting, the labels of seen categories in the source domains are perturbed and denoted by $\hat{\mathbf{y}}_s$. 
The goal of our task is to ask the model to give accurate classification results on seen categories for $\mathbf{X}_t^k = \{(\mathbf{x}_t^k, \mathbf{y}_t^k)|\mathbf{y}_t^k \in \Omega_k, \mathbf{X}_t^u \in \mathcal{D}_m\}$ while delivering low confidence score on unseen categories for $\mathbf{X}_t^u = \{(\mathbf{x}_t^u, \mathbf{y}_t^u)| \mathbf{y}_t^u \in \Omega_u, \mathbf{x}_t^u \in \mathcal{D}_m\}$ in the test set, based on the a priori knowledge provided by samples with label noise $\{\hat{\mathbf{y}}_s\}$ from known categories in the source domains provided by the train set. 

\subsection{Noisy Label Settings}
\label{sec:label_noise}
To construct the first OSDG-NL benchmark, we select two types of label noise, which are symmetric and asymmetric label noises.
Symmetric noise involves randomly reassigning training labels to other classes at predefined noise levels ($20\%$, $50\%$, $80\%$ in our work), simulating varying degrees of label corruption. 
In contrast, asymmetric noise considers the semantic similarity between classes, altering labels more likely to the most related categories (\textit{e.g.}, ``\textit{horse}'' might be mislabeled as ``\textit{dog}'').
We employ the BERT model~\cite{BERT} to extract semantic features based on the textual description of each class and measure the semantic categorical distance using cosine similarity. 
Then, we use the calculated similarity scores as weights for mislabeling.
For asymmetric noise, we set the label noise ratio as $50\%$.

\subsection{Datasets}
\label{sec:datasets}
\noindent\textbf{PACS:} 
The PACS~\cite{li2017deeper} is a well-established dataset in the OSDG field, which consists of $4$ distinct domains: \textit{Photo}, \textit{Art Painting}, \textit{Cartoon}, and \textit{Sketch}. 
These domains introduce large visual domain shifts, making PACS suitable for our task. The dataset includes $9,991$ images across $7$ shared categories (\textit{i.e.}, \textit{dog}, \textit{elephant}, \textit{giraffe}, \textit{guitar}, \textit{house}, \textit{horse}, \textit{person}).
PACS is widely used in domain generalization studies with a leave-one-domain-out protocol, which is also leveraged in our work, where the model is trained on three domains and tested on the other domain. We follow the open-set protocol introduced by Wang~\textit{et al.}~\cite{wang2023generalizable} by selecting the last $1$ category as the unseen category, where all the categories are rearranged into alphabet order according to their text descriptions.

\noindent\textbf{DigitsDG:} 
The Digits-DG~\cite{zhou2020deep} is another domain generalization dataset, comprising $4$ distinct domains, \textit{i.e.}, \textit{mnist}, \textit{mnist$_m$}, \textit{svhn}, and \textit{syn}. Each domain introduces distinct visual styles, ranging from grayscale handwritten digits to real-world and synthetic digit images, resulting in obvious domain shifts. The dataset includes $10$ shared classes (digits $0{\sim}9$). We follow the open-set protocol introduced by Wang~\textit{et al.}~\cite{wang2023generalizable} by selecting the last $4$ categories as unseen categories which are rearranged into alphabet order.

\noindent\textbf{DomainNet:} 
DomainNet~\cite{peng2019moment} is a large-scale benchmark dataset designed for domain adaptation and generalization tasks, containing approximately $600,000$ images across six distinct domains: \textit{Clipart}, \textit{Infograph}, \textit{Painting}, \textit{Quickdraw}, \textit{Real}, and \textit{Sketch}. It includes $345$ object categories shared among all domains, enabling evaluation of model performance under significant domain shifts. The first $182$ categories following alphabet order are selected as seen classes while the rest are selected as unseen classes.

\subsection{Baselines}
\label{sec:baselines}
We employ well-established approaches in both the open-set domain generalization field and the noisy label learning field to construct our OSDG-NL benchmarks.

\noindent\textbf{Open-set domain generalization baselines:}
We select $6$ baselines from the open-set domain generalization field, each exemplifying distinct strategies to address domain generalization challenges. These methods cover a wide spectrum of approaches, including meta-learning (MEDIC~\cite{wang2023generalizable}, EBiL-HaDS~\cite{peng2024advancing} and MLDG~\cite{shu2019meta}), adversarial learning (ARPL~\cite{chen2021adversarial}), data augmentation (MixStyle~\cite{zhou2020domain}), generative modeling (ODGNet~\cite{bose2023beyond}), and optimization techniques (SWAD~\cite{cha2021swad}), ensuring a comprehensive evaluation.
These baselines collectively provide a well-rounded comparison, covering key contributions of open-set domain generalization. Thereby, we would like to evaluate their performances when they are facing various label noise challenges.

\noindent\textbf{Noisy label learning baselines:}
We select $6$ recent representative approaches from the noisy label learning field on the general image classification task as another group of baselines in our benchmarks, including TCL~\cite{huang2023twin}, NPN~\cite{sheng2024adaptive}, BadLabel~\cite{zhang2024badlabel}, DISC~\cite{li2023disc}, LSL~\cite{kim2024learning}, and PLM~\cite{zhao2024estimating}, each addressing noisy labels through distinct strategies, \textit{e.g.}, sample selection-based contrastive learning, loss-based label denoising, and label correction. 
These methods provide a comprehensive basis for evaluating noisy label learning techniques.

\begin{figure*}[!t]
\centering 
\includegraphics[width=\textwidth]{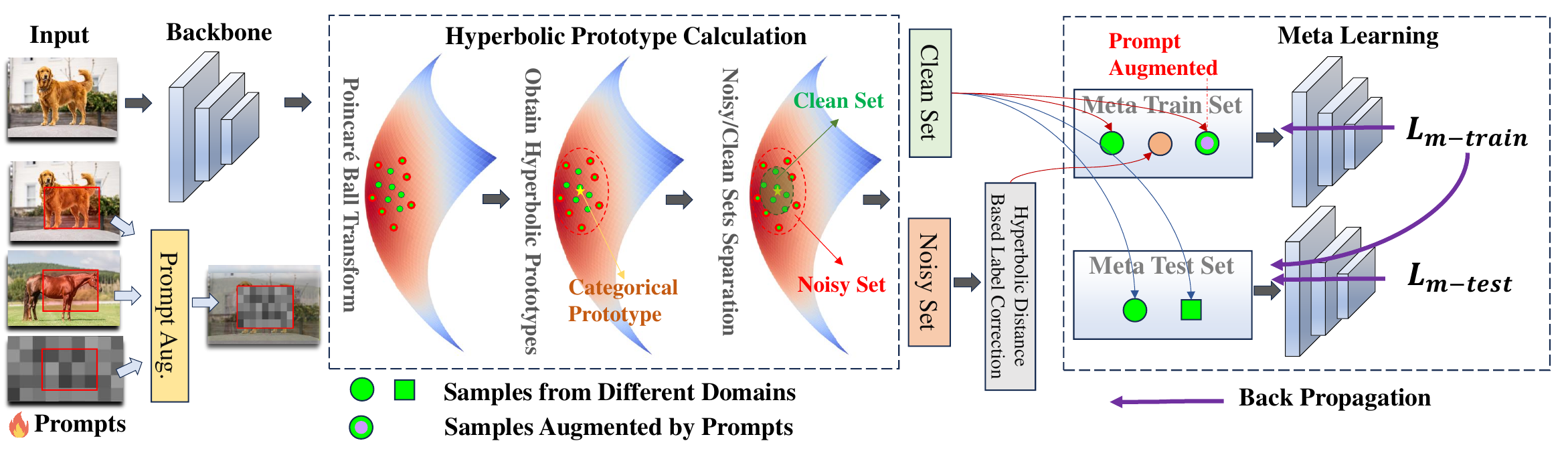}
\caption{Overview of the proposed HyProMeta framework for Open-Set Domain Generalization under Noisy Labels (OSDG-NL). The model first computes hyperbolic category-wise prototypes from source domain data and uses the mode of hyperbolic distances to partition samples into clean and noisy subsets. Clean samples are used directly for meta-training, while noisy samples are relabeled based on proximity to prototypes. A learnable prompt is introduced for category-aware augmentation, where mixed samples from different classes are combined with prompt patches and assigned to an auxiliary out-of-distribution class. The framework iteratively updates through meta-train and meta-test phases, enabling robust generalization to unseen categories and domains despite label noise.}
\label{fig:main}
\end{figure*} 
    
\section{Method}
\label{sec4}
Meta-learning proves to be highly effective for domain generalization by simulating a diverse range of cross-domain tasks during the training process, thereby facilitating the learning of adaptable representations capable of generalizing to previously unseen domains~\cite{li2018learning}. By emphasizing task-specific and category-specific learning dynamics, meta-learning enables models to efficiently adapt to novel domains and novel class distributions and effectively address the challenges posed by domain shifts under open-set conditions~\cite{wang2023generalizable,shu2021open}. 
In this work we improve the robustness of meta-learning for open-set domain generalization against label noise perturbation during the training time. In this section, we will introduce our proposed novel method, HyProMeta, which is composed of hyperbolic categorical prototype-based label noise-aware meta-learning (as introduced in Section~\ref{sec:hyperbolic}) and new category-aware prompt learning to improve the discriminative ability of the learned embeddings (as introduced in Section~\ref{sec:nca_prompts}).

\begin{algorithm*}[t]
    \caption{Training with HyProMeta.}
    \label{algorithm}
    \renewcommand{\thealgorithm}{}
    \begin{algorithmic}[1]

\Require
    $\Omega_{\mathcal{D}_S} \coloneqq \text{Set of source domains}$, 
    \quad $\mathcal{C} \coloneqq \text{Set of known classes}, \quad C \coloneqq |\mathcal{C}|$, \\
    $k \in \{1,\ldots,C\} \coloneqq \text{Class index}$, \quad $\mathbf{M}_{\alpha} \coloneqq \text{Neural network backbone with parameters } \alpha$, \\ 
    $\mathbf{H}_{\beta} \coloneqq \text{Classification head with parameters $\beta$}$, \\
    $CE \coloneqq \text{Cross entropy loss function}$, \quad
    $\mathbf{p}_l \coloneqq \text{Learnable image-wise prompt}$\\

\While{not converged}
\State $\{\mathcal{D}_{s_i}, \mathcal{D}_{s_j}\} \leftarrow \Omega_{\mathcal{D}_S}$ \sComment{Select two random source domains}
\Statex
\LComment{Hyperbolic Distance Based Label Correction}
\State $\{\mathbf{P}_k^{s}, \hat{d}_k^{s}\} \leftarrow \text{HyperbolicPrototypesAndThresholds}(\{\mathcal{D}_{s_i}, \mathcal{D}_{s_j}\})$ \sComment{Domain- and category-specific}

\State $d_k^{s} \leftarrow \text{ComputeHyperbolicDistances}(\{\mathcal{D}_{s_i}, \mathcal{D}_{s_j}\}, \mathbf{P}_k^{s})$ \sComment{Calculate distances to prototypes}
\State $\Omega_c, \Omega_n \leftarrow \{x | d_k^{s}(x) < \hat{d}_k^{s}\}, \{x | d_k^{s}(x) \geq \hat{d}_k^{s}\}$  \sComment{Split into clean and noisy set.}

\State $\mathbf{y}_n^* \leftarrow \argmin_k d_k^{s_i}(\Omega_n)$ \sComment{Correct noisy labels using nearest prototype}
\State $\Omega_n^* \leftarrow \{\Omega_n, \mathbf{y}_n^*\}$ \sComment{Create corrected noisy set}

\Statex 
\LComment{Meta-train on first domain}
\State $\{\mathbf{B}^{s_i}_n, \mathbf{B}^{s_i}_c\} \leftarrow \text{SampleBatches}(\Omega_n^*, \Omega_c^{s_i})$ \sComment{Sample from noisy and clean sets}
\State $\hat{\mathbf{B}}^{s_i}_c \leftarrow \text{SampleDifferentClasses}(\Omega_c^{s_i}, \text{Classes}(\mathbf{B}^{s_i}_c))$ \sComment{Sample batch with different classes}

\State $\mathbf{M}_{crop} \leftarrow \text{GenerateRandomMask}()$ \sComment{Generate random crop mask}
\State $\mathbf{B}_{mixed} \leftarrow \text{Mix}(\mathbf{B}^{s_i}_c, \hat{\mathbf{B}}^{s_i}_c)$ \sComment{Mix clean samples}
\State $\hat{\mathbf{B}}^{s_i}_{a} \leftarrow \text{PromptAug}(\mathbf{B}_{mixed}, \mathbf{p}_l, \mathbf{M}_{crop})$ \sComment{Apply prompt augmentation}
\State $\mathbf{y}_a \leftarrow (C+1) \cdot \mathbf{1}$ \sComment{Assign unknown class label to augmented samples}

\State $L_{m-train} \leftarrow CE(\mathbf{H}_{\beta}(\mathbf{M}_{\alpha}(\mathbf{B}^{s_i}_c)), \mathbf{y}_{c})$ \sComment{Clean samples loss}
\State $L_{m-train} \leftarrow L_{m-train} + CE(\mathbf{H}_{\beta}(\mathbf{M}_{\alpha}(\hat{\mathbf{B}}^{s_i}_a)), \mathbf{y}_{a})$ \sComment{Augmented samples loss}
\State $L_{m-train} \leftarrow L_{m-train} + CE(\mathbf{H}_{\beta}(\mathbf{M}_{\alpha}(\mathbf{B}^{s_i}_n)), \mathbf{y}_{n}^*)$ \sComment{Corrected noisy samples loss}
\State $\text{UpdateParameters}(L_{m-train})$ \sComment{Update based on meta-train loss}
\Statex

\LComment{Meta-test on both domains}

\State $\mathbf{B}_c^{s_i} \leftarrow \text{SampleBatch}(\Omega_c^{s_i})$ \sComment{Sample from first domain}
\State $\mathbf{B}_c^{s_j} \leftarrow \text{SampleBatch}(\Omega_c^{s_j})$ \sComment{Sample from second domain}
\State $L_{m-test} \leftarrow CE(\mathbf{H}_{\beta}(\mathbf{M}_{\alpha}(\mathbf{B}_c^{s_i})), \mathbf{y}_{c}^{s_i})$ \sComment{First domain loss}
\State $L_{m-test} \leftarrow L_{m-test} + CE(\mathbf{H}_{\beta}(\mathbf{M}_{\alpha}(\mathbf{B}_c^{s_j})), \mathbf{y}_{c}^{s_j})$ \sComment{Second domain loss}
\Statex

\State $\text{UpdateParameters}(L_{m-test} + L_{m-train})$ \sComment{Final parameter update}
\EndWhile
\end{algorithmic}
\end{algorithm*}

\subsection{Hyperbolic Categorical Prototype-Based Meta Learning}
\label{sec:hyperbolic}

\noindent\textbf{Hyperbolic Prototype Calculation:} Hyperbolic space enhances the generalization capabilities of deep learning models by efficiently modeling hierarchical information and enabling compact representations of complex relationships. Its intrinsic geometric properties facilitate the preservation of both global and local data structures, reducing overfitting and improving robustness to unseen data~\cite{cui2024rethinking_few_shot,liu2020hyperbolic}.

Leveraging these generalizability advantages, we utilize hyperbolic space to generate category prototypes from the extracted embeddings of training samples during meta-learning.

A hyperbolic space is defined as a complete and connected mutually isometric Riemannian manifold, which is characterized by a constant negative sectional curvature~\cite{fang2023Poincare,absil2008optimization}. The Poincaré ball model is highly effective in hyperbolic space modeling, preserving both global and local geometric relationships~\cite{mishne2023numerical}. We select the Poincaré ball model, which is introduced in Eq.~\ref{eq:1}.
\begin{equation}
\label{eq:1}
    \mathcal{P}^{n}=\left\{\mathbf{z} \in \mathcal{R}^{n} \mid |\mathbf{z}|^{2}<r^2\right\},
\end{equation}
where $r$ denotes the radius, $\mathbf{z}$ denotes the high-dimensional representations, and $n$ denotes the dimension of Poincaré ball representations.

We transfer the learned representation of the samples from the known categories of source domains onto the hyperbolic space to obtain the categorical prototype. The distance between two samples in the hyperbolic space can be measured using the Riemannian metric, which is denoted by Eq.~\ref{eq:2},
\begin{equation}
\label{eq:2}
    d_\mathcal{G}(\mathbf{z}_1, \mathbf{z}_2)=2r \tanh ^{-1}\left(\frac{\left\|-\mathbf{z}_1 \oplus \mathbf{z}_2\right\|}{r}\right),
\end{equation}
where $\oplus$ is the differentiable Möbius addition as detailed in Eq.~\ref{eq:3}, $\mathbf{z}_1$ and $\mathbf{z}_2$ denote two embeddings provided by the leveraged model, on which we would like to calculate the hyperbolic distances.
\begin{equation}
\label{eq:3}
\begin{split}
    &\mathbf{z}_1 \oplus \mathbf{z}_2 = 
    \\
    &\frac{(1 + 2\gamma\langle \mathbf{z}_1, \mathbf{z}_2\rangle+ \gamma\left\| \mathbf{z}_1 \right\|^2)\mathbf{z}_1 +
    (1 - \gamma\left\|\mathbf{z}_1 \right\|^2)\mathbf{z}_2}{1 + 2\gamma\langle \mathbf{z}_1, \mathbf{z}_2\rangle + \gamma^2\left\| \mathbf{z}_1 \right\|^2\left\| \mathbf{z}_2 \right\|^2},
\end{split}
\end{equation}
where $\gamma = \frac{1}{r^2}$.
The mapping from Euclidean space to hyperbolic space can be represented by Eq.~\ref{eq_4}, and the base point $\mathbf{q}$ is set to zeros. $\lambda_z$ denotes a scale factor.
\begin{equation}
\label{eq_4}
    \Phi_h(\mathbf{z}) = \mathbf{q} \oplus (tanh(\frac{1}{\sqrt{\gamma}}\frac{\lambda_z\left\| \mathbf{z} \right\|}{2})) \frac{\mathbf{z}}{\sqrt{\gamma}\left\|\mathbf{z} \right\|}.
\end{equation}

Given input $\mathbf{x}$, we first pass it through the leveraged deep learning backbone $\mathbf{M}_{\alpha}$ to obtain high dimensional embeddings $\mathbf{z}$, where $\mathbf{z} = \mathbf{M}_{\alpha}(\mathbf{x})$. Then we project the resultant embedding into hyperbolic space using $\hat{\mathbf{z}} = \Phi_h(\mathbf{z})$. We calculate the hyperbolic category center according to Eq.~\ref{eq:center} for each domain $s$ and category $k$, where $s \in \Omega_s$ and $k \in \Omega_k$.
\begin{equation}
\label{eq:center}
    \mathbf{e}_k^s = \sum_{i=1}^{N_{k}}{\Phi_h(\mathbf{M}_{\alpha}(\hat{\mathbf{z}}_k^{(i, s)}))}, k \in \left\{1, ..., C \right\},
\end{equation}
where $\mathbf{e}_k$ denotes the hyperbolic center of category $k$. $N_{k}$ denotes the number of samples inside category $k$. The hyperbolic category center is calculated with a fixed epoch step $N_{epoch}$. After the calculation of the domain-agnostic category prototypes in the hyperbolic space, we achieve different data partitions for meta-test and meta-train assignment using the hyperbolic distance of each sample to its corresponding domain-agnostic category prototype. The distance can be represented by Eq.~\ref{eq:6}.
\begin{equation}
\label{eq:6}
    d_k^{(i,s)} = d_{\mathcal{G}}(\mathbf{z}_k^{(i, s)}, \mathbf{e}_k^s),
\end{equation}
where $i$ indicates the sample index, $k$ denotes the category, and $s$ denotes one source domain.
Then, we calculate the mode of distances within each domain for each category, denoted by $\hat{d}_k^s$, where $\hat{d}_k^s = \text{Mode}(\{d_k^{(i,s)}\})$, where the distances are binned to their upper bound using binned resolution of $\delta
$. We use the mode of hyperbolic distances for clean/noisy partitioning because it effectively captures the dominant distribution of clean samples within each class and domain, which typically forms a compact cluster. Unlike mean or median, the mode is less affected by noisy outliers and takes the majority structure of clean-label samples into consideration, which enables more robust thresholding under severe label noise.
The samples that have smaller hyperbolic distances to their corresponding domain-agnostic category prototypes are assigned to the clean set, while the others are assigned to the noisy set, where we utilize the hyperbolic distances as the measurement of label uncertainty. 
The samples from the noisy set are relabeled by the category of their nearest hyperbolic prototype.
Then we use the aforementioned partitions as described in Alg.~\ref{algorithm} to achieve label noise agnostic meta-learning to handle the OSDG-NL challenge. 

\noindent\textbf{Meta Task Assignment:} Our proposed meta-learning framework employs hyperbolic space prototypes to dynamically distinguish between clean and noisy labels during training, addressing the challenges posed by Open-Set Domain Generalization under Noisy Labels (OSDG-NL). Specifically, hyperbolic distances to category-specific prototypes are computed to partition training samples into clean and noisy subsets. Labels of noisy samples are corrected by aligning them with the nearest hyperbolic prototype, ensuring that the label noise is mitigated effectively. This approach exploits the hierarchical and geometric properties of hyperbolic space, enabling compact representation of data relationships and preserving both global and local structures.

To further enhance generalization capabilities, the method integrates a learnable prompt for category-aware data augmentation as an additional category, effectively capturing intra-class variability and reinforcing the model’s robustness to distribution shifts, which will be introduced in detail in the next subsection. 

The training process iteratively alternates between the meta-train and meta-test phases. During the meta-train phase, the model is optimized on a combination of clean samples, corrected noisy samples, and augmented OOD samples. This multi-faceted loss formulation ensures that the model learns robust representations across varied input scenarios. The meta-test phase evaluates the model's performance on clean samples from different source domains, enabling assessment of its generalization across domain shifts. The iterative parameter updates integrate information from both phases, ensuring the model learns to handle noisy labels while maintaining the ability to generalize to unseen domains and categories.

This framework is particularly advantageous for OSDG-NL by mitigating label noise and improving domain generalization. Through leveraging hyperbolic prototypes, the method robustly separates noisy and clean data while providing reliable label corrections. The inclusion of a learnable prompt facilitates enhanced data augmentation, targeting intra-class compactness and OOD generalizability. Consequently, the proposed approach systematically addresses the dual challenges of label noise and domain shifts, resulting in improved performance across diverse and dynamic real-world scenarios. Next, a detailed introduction on how to achieve prompt-based augmentation is presented.

\subsection{New Category Aware Prompt Learning (NCA-Prompt)}
\label{sec:nca_prompts}
We have previously introduced the hyperbolic prototype-based meta-learning framework, which separates meta-train and meta-test phases by dynamically partitioning data into clean and noisy subsets based on their respective hyperbolic distances from category prototypes.
While this framework effectively manages label shifts induced by label noise, it is also necessary to address distribution shifts that occur within a single class and domain. Specifically, the model must demonstrate high confidence when encountering OOD data that remain within the same category and domain.

Existing approaches, such as Deep Metric Learning (DML)~\cite{wang2017deep}, focus on achieving compact intra-class representations. However, DML methods are inherently sensitive to label noise. Forcing samples with noisy or incorrect labels to align closely in the feature space can degrade the performance of our hyperbolic prototype-based meta-learning framework, as it inadvertently propagates noise and hinders the model's ability to distinguish between clean and noisy samples. Therefore, an alternative strategy is required to achieve robust intra-class compactness during the meta-train process while being resilient to noisy labels.

Enhancing the model's generalizability to OOD samples beyond mere label shifts is equally critical. While MixStyle~\cite{zhou2020domain} has shown that domain-mixed samples can improve generalization across different domains via data augmentation, it does not explicitly target the intra-class and intra-domain generalizability required in our scenario. To bridge this gap, we propose a novel category-aware prompt learning mechanism to complement hyperbolic prototype-based meta-learning, particularly in the context of open-set domain generalization under noisy labels.

For each input-label pair, denoted as $\mathbf{z}_i = (\mathbf{x}_i, \hat{\mathbf{y}}_i)$ where $\mathbf{x}_i \in \mathcal{R}^{H \times W \times 3}$, we select another sample from a different category and domain, denoted as $\mathbf{z}_j = (\mathbf{x}_j, \hat{\mathbf{y}}_j)$. The two images are combined via simple averaging to generate a mixed sample by $\mathbf{z}^* = \frac{\mathbf{z}_i + \mathbf{z}_j}{2}$,

Additionally, we introduce a learnable prompt $\mathbf{p}_l \in \mathcal{R}^{H \times W \times 3}$, which has the same spatial dimensions as the input image. The learnable prompt is used to augment the mixed samples in order to enrich the data distribution for different categories during the training phase.
A random cropping window is selected on the mixed sample $\mathbf{z}^*$, and the corresponding region in $\mathbf{z}^*$ is replaced with the same region from $\mathbf{p}_l$. This operation is formally expressed as Eq.~\ref{eq:7}.
\begin{equation}
\label{eq:7}
    \mathbf{z}^*\left[\mathbf{M}_{crop}\right] := \mathbf{p}_l\left[\mathbf{M}_{crop} \right],
\end{equation}
where $\mathbf{M}_{\text{crop}}$ represents the cropped region mask.

To generate a rectangle prompt region, two random points within the image dimensions are sampled to define the top-left and bottom-right corners. The rectangle's area is calculated and compared against the total image area to ensure that its relative size falls within the specified range $\left[rel_{min},rel_{max}\right]$. This ensures that the selected region for each object occupies a controllable proportion of the image, maintaining both variability and constraint in the prompt generation.

To further enforce the model's awareness of OOD samples, the augmented mixed sample $\mathbf{z}^*$ is assigned an additional pseudo-category, labeled as $C+1$, where $C$ is the total number of original categories. By doing so, the model is encouraged to learn more compact intra-class representations through its exposure to realistic OOD samples, thereby improving its robustness to distribution shifts.

Our experimental results validate the effectiveness of this approach in enhancing the hyperbolic prototype-based meta-learning framework for the OSDG-NL task. 
The proposed distribution-aware cropped prompt learning strategy not only addresses intra-class compactness but also facilitates the model's generalizability to unseen categories, further strengthening its performance in open-set scenarios.

\section{Experiments}
\label{sec5}
In this section, we first illustrate the implementation details of our approach in Section~\ref{sec:impl} and evaluation metrics in Section~\ref{sec:metrics}. Then we deliver the analysis on the constructed benchmarks and our proposed method in Section~\ref{sec:benchmarks}. 
Next, we conduct analysis towards the effect brought by different pretraining strategies in Section~\ref{sec:pretraining} and the analysis on the label-clean/noisy set partition in Section~\ref{sec:clean_noisy}, followed by the analysis on the effect brought by different textual encoder for asymmetric label noise generation, ablation studies of different modules (in Section~\ref{sec:abl}), analysis on the t-SNE visualizations (in Section~\ref{sec:tsne}), and analysis on the confidence score towards seen and unseen categories in the test domain (in Section~\ref{sec:conf}). The ablation of NCA-Prompt, HYB-Meta, and hyperparameter $N_{epoch}$ are analyzed in Section~\ref{sec:nca}, Section~\ref{sec:hyb}, and Section~\ref{sec:hyper}. 

\subsection{Implementation Details}
\label{sec:impl}
All experiments are conducted using PyTorch 2.0 and a single NVIDIA A100 GPU. 
Training is capped at \(1 \times 10^4\) steps, employing the SGD optimizer with a Learning Rate (LR) of \(1 \times 10^{-3}\) and a batch size of $16$. 
A learning rate decay of \(1 \times 10^{-1}\) is applied after \(8 \times 10^3\) meta-training steps. $\delta$ is chosen as $1 \times 10^{-2}$. 
The experiments use a worker number of $4$, and the value of \(\gamma\) is fixed at \(2 \times 10^{-5}\). For feature extraction, the ConvNet~\cite{zhou2021domain} is employed as the backbone network on the DigitsDG dataset, following the setup outlined in~\cite{zhou2021domain}. 
$\left[rel_{min},rel_{max}\right]$ is set as $\left[0.2,0.6\right]$. The curvature, $\lambda_z$, and radius of the Poincaré ball are selected as $-5 \times 10^{-2}$, $2.0$, and $9.99$, the dimension of $z$ is $512$.
Hyperparameter $N_{epoch}$ is set as $500$, which is chosen according to the ablation. OSCR is chosen as the major evaluation metric, while H-score and close-set accuracy (ACC) are chosen as secondary metrics following~\cite{wang2023generalizable}.

\subsection{Evaluation Metrics}
\label{sec:metrics}
The \textit{Acc} metric denotes the closed-set accuracy evaluated on seen categories and serves to measure the correctness of classification within the known label space. In contrast, the \textit{H-score} and \textit{OSCR} are widely adopted in the OSDG field~\cite{wang2022generalizing,peng2024advancing} to evaluate open-set recognition performance. Since the \textit{H-score} requires a manually selected threshold derived from source domain validation sets to distinguish between seen and unseen categories, it is treated as a secondary metric in our evaluation. Instead, we adopt \textit{OSCR}, initially proposed by MEDIC~\cite{wang2023generalizable}, as our primary evaluation metric, as it does not depend on a fixed threshold and provides a more flexible and robust assessment of open-set recognition.
To compute the \textit{H-score}, a threshold ratio $\sigma$ is first selected to distinguish samples from seen and unseen categories. If the predicted confidence score for a sample falls below $\sigma$, it is classified as an unseen category. The accuracy for samples predicted as seen categories is then computed based on their ground-truth seen labels and denoted as $Acc_k$. For unseen category evaluation, a binary classification approach is applied: samples from seen categories are labeled as $1$ and those from unseen categories as $0$, and the classification accuracy is denoted as $Acc_u$. The final H-score is calculated as:
\begin{equation}
H_{score} = \frac{2Acc_uAcc_k }{Acc_u + Acc_k }.
\end{equation}
The \textit{OSCR} metric evaluates the quality of confidence scores through a moving threshold mechanism, combining accuracy with AUROC-like behavior. Unlike AUROC, however, OSCR only considers correctly predicted samples at each threshold, blending the evaluation principles of both H-score and AUROC to better reflect model performance in the OSDG setting.

\begin{table*}[t!]
\caption{Results (\%) of PACS on ResNet18~\cite{he2016deep}. The open-set ratio is $6{:}1$ and symmetric label noise with a ratio of $20\%$ is selected.}
\label{tab:pacs_res18_noise_20}
\centering
\resizebox{1.\linewidth}{!}{
\begin{tabular}{l|ccc|ccc|ccc|ccc|ccc}
\toprule
& \multicolumn{3}{c|}{\textbf{Photo (P)}} & \multicolumn{3}{c|}{\textbf{Art (A)}} & \multicolumn{3}{c|}{\textbf{Cartoon (C)}} & \multicolumn{3}{c|}{\textbf{Sketch (S)}} & \multicolumn{3}{c}{\textbf{Avg}} \\
\textbf{Method} & Acc & H-score & \cellcolor{gray!25}OSCR & Acc & H-score & \cellcolor{gray!25}OSCR & Acc & H-score & \cellcolor{gray!25}OSCR & Acc & H-score & \cellcolor{gray!25}OSCR & Acc & H-score & \cellcolor{gray!25}OSCR \\

\midrule

TCL~\cite{huang2023twin}  & 58.32 &	59.21&\cellcolor{gray!25}51.72  & 53.66 & 48.28 & \cellcolor{gray!25}42.91 & 46.78 & 38.29 & \cellcolor{gray!25}31.74 & 31.55 & 22.88 & \cellcolor{gray!25}24.30 & 47.58 & 42.17 & \cellcolor{gray!25}37.67 \\

NPN~\cite{sheng2024adaptive}  & 64.30 & 70.87 & \cellcolor{gray!25}61.99 & 51.66 & 52.10 &  \cellcolor{gray!25}45.40 & 55.65 & 44.88 & \cellcolor{gray!25}38.64 & 35.58 & 22.35 & \cellcolor{gray!25}25.86 & 51.80 & 47.55 & \cellcolor{gray!25}42.97 \\

BadLabel~\cite{zhang2024badlabel}  &54.93  &55.73  & \cellcolor{gray!25}48.24 & 53.72 & 53.25 & \cellcolor{gray!25}46.55 & 50.23 & 55.36 & \cellcolor{gray!25}45.70 & 31.55 & 21.84 & \cellcolor{gray!25}\textbf{28.38} & 47.61 & 46.55 & \cellcolor{gray!25}42.22 \\

DISC~\cite{li2023disc} &53.47  &56.13  & \cellcolor{gray!25}47.22 & 54.47 & 47.46 & \cellcolor{gray!25}43.48 & 53.27 & 53.97 & \cellcolor{gray!25}44.33 & 24.01 & 16.75 & \cellcolor{gray!25}11.52 & 46.31 & 43.58 & \cellcolor{gray!25}36.64 \\
LSL~\cite{kim2024learning}  & 58.97 & 58.93 & \cellcolor{gray!25}52.15 & 49.97 & 48.17 & \cellcolor{gray!25}39.20 & 47.50 & 44.07 & \cellcolor{gray!25}34.63 & 30.59 & 12.44 & \cellcolor{gray!25}16.81 & 46.76 & 40.90 & \cellcolor{gray!25}35.70 \\
PLM~\cite{zhao2024estimating}  & 55.57 &42.33  & \cellcolor{gray!25}38.27 & 41.78 & 43.09 & \cellcolor{gray!25}32.95 & 45.75 & 40.44 & \cellcolor{gray!25}33.26 & 33.27 & 12.11 & \cellcolor{gray!25}15.46 & 38.33 & 34.49 & \cellcolor{gray!25}29.99 \\
\bottomrule
\end{tabular}

}
\end{table*}

\begin{table*}[t!]
\caption{Results (\%) of PACS on ResNet18~\cite{he2016deep}.
The open-set ratio is $6{:}1$ and symmetric label noise with a ratio of $50\%$ is selected.}

\label{tab:pacs_res18_50}
\centering
\resizebox{1.\linewidth}{!}{
\begin{tabular}{l|ccc|ccc|ccc|ccc|ccc}
\toprule
& \multicolumn{3}{c|}{\textbf{Photo (P)}} & \multicolumn{3}{c|}{\textbf{Art (A)}} & \multicolumn{3}{c|}{\textbf{Cartoon (C)}} & \multicolumn{3}{c|}{\textbf{Sketch (S)}} & \multicolumn{3}{c}{\textbf{Avg}} \\
\textbf{Method} & Acc & H-score & \cellcolor{gray!25}OSCR & Acc & H-score & \cellcolor{gray!25}OSCR & Acc & H-score & \cellcolor{gray!25}OSCR & Acc & H-score & \cellcolor{gray!25}OSCR & Acc & H-score & \cellcolor{gray!25}OSCR \\

\midrule

TCL~\cite{huang2023twin}  & 54.68 & 52.40 & \cellcolor{gray!25}46.51 & 52.78 & 22.65 & \cellcolor{gray!25}30.94 & 47.19 & 37.73 & \cellcolor{gray!25}34.89 & 26.33 & 9.83 & \cellcolor{gray!25}7.62 & 45.25 & 30.65 & \cellcolor{gray!25}29.99 \\

NPN~\cite{sheng2024adaptive} & 48.38 & 38.12 & \cellcolor{gray!25}33.55 & 35.71 & 32.33 & \cellcolor{gray!25}24.47 & 38.94 & 26.88 & \cellcolor{gray!25}18.60 & 26.93 & 27.59 & \cellcolor{gray!25}18.96 & 37.49 & 31.23 & \cellcolor{gray!25}23.90 \\

BadLabel~\cite{zhang2024badlabel} & 46.20 & 57.45 & \cellcolor{gray!25}45.07 & 45.34 & 47.29 & \cellcolor{gray!25}37.89 & 35.17 & 43.35 & \cellcolor{gray!25}32.14 & 28.40 & 26.95 & \cellcolor{gray!25}15.71 & 38.78 & 43.76 & \cellcolor{gray!25}32.70 \\

DISC~\cite{li2023disc} & 52.52 & 56.07 & \cellcolor{gray!25}50.55 & 46.84 & 31.91 & \cellcolor{gray!25}30.35 & 28.47 & 28.28 & \cellcolor{gray!25}19.97 & 30.83 & 25.63 & \cellcolor{gray!25}\textbf{24.78} & 39.67 & 35.47 & \cellcolor{gray!25}31.41 \\

LSL~\cite{kim2024learning}  & 41.36 & 30.83 & \cellcolor{gray!25}20.27 & 42.28 & 39.78 & \cellcolor{gray!25}31.40 & 42.39 & 37.59 & \cellcolor{gray!25}30.89 & 26.90 & 15.40 & \cellcolor{gray!25}7.42 & 38.23 & 30.90 & \cellcolor{gray!25}22.50 \\
PLM~\cite{zhao2024estimating}  &55.57  &42.33  & \cellcolor{gray!25}38.27 & 39.21 & 27.81 & \cellcolor{gray!25}24.93 & 33.01 & 27.81 & \cellcolor{gray!25}21.49 & 25.52 & 6.65 & \cellcolor{gray!25}13.41 & 38.33 & 26.15 & \cellcolor{gray!25}24.53 \\
\bottomrule
\end{tabular}
}
\end{table*}

\begin{table*}[t!]
\caption{Results (\%) of PACS on ResNet18~\cite{he2016deep}.
The open-set ratio is $6{:}1$ and symmetric label noise with a ratio of $80\%$ is selected.}

\label{tab:pacs_res18_80}
\centering
\resizebox{1.\linewidth}{!}{
\begin{tabular}{l|ccc|ccc|ccc|ccc|ccc}
\toprule
& \multicolumn{3}{c|}{\textbf{Photo (P)}} & \multicolumn{3}{c|}{\textbf{Art (A)}} & \multicolumn{3}{c|}{\textbf{Cartoon (C)}} & \multicolumn{3}{c|}{\textbf{Sketch (S)}} & \multicolumn{3}{c}{\textbf{Avg}} \\
\textbf{Method} & Acc & H-score & \cellcolor{gray!25}OSCR & Acc & H-score & \cellcolor{gray!25}OSCR & Acc & H-score & \cellcolor{gray!25}OSCR & Acc & H-score & \cellcolor{gray!25}OSCR & Acc & H-score & \cellcolor{gray!25}OSCR \\

\midrule

TCL~\cite{huang2023twin}  &31.58&	25.81&\cellcolor{gray!25}17.39  & 27.08 & \textbf{26.60} & \cellcolor{gray!25}16.45 & 27.69 & 27.17 & \cellcolor{gray!25}17.66 & 21.20 & 8.52 & \cellcolor{gray!25}14.87 & 26.89 & 22.03 & \cellcolor{gray!25}16.59 \\

NPN~\cite{sheng2024adaptive} & 17.21 & 12.49 & \cellcolor{gray!25}10.18 & 24.27 & 12.67 & \cellcolor{gray!25}13.87 & 22.85 & 12.99 & \cellcolor{gray!25}10.85 & 19.66 & 4.31 & \cellcolor{gray!25}11.63 & 21.00 & 10.62 & \cellcolor{gray!25}11.63 \\

BadLabel~\cite{zhang2024badlabel}  & 22.62 & 14.11 & \cellcolor{gray!25}22.62 & 15.95 & 14.50 & \cellcolor{gray!25}10.16 & 19.39 & 24.34 & \cellcolor{gray!25}14.95 & 26.13 & 14.14 & \cellcolor{gray!25}17.96 & 21.02  & 16.77 & \cellcolor{gray!25}16.42 \\

DISC~\cite{li2023disc} & 22.05 & 19.53 & \cellcolor{gray!25}15.27 & 24.77 & 23.65 & \cellcolor{gray!25}17.53 & 27.13 & 22.49 & \cellcolor{gray!25}14.76 & 16.03 & 12.86 & \cellcolor{gray!25}10.00 & 22.05 & 19.63 & \cellcolor{gray!25}14.39 \\

LSL~\cite{kim2024learning}  & 18.58 & 22.82 & \cellcolor{gray!25}13.89 & 23.64 & 16.71 & \cellcolor{gray!25}14.56 & 15.37 & 15.84 & \cellcolor{gray!25}7.94 & 21.68 & 1.92 & \cellcolor{gray!25}8.26 & 19.82 & 14.32 & \cellcolor{gray!25}11.16 \\
PLM~\cite{zhao2024estimating}  &24.39  &8.71  & \cellcolor{gray!25}9.41 & 30.08 & 24.67 & \cellcolor{gray!25}17.85 &  20.94& 13.16 & \cellcolor{gray!25}12.21 & 21.76 & 21.26 & \cellcolor{gray!25}11.74 & 24.29 & 16.95 & \cellcolor{gray!25}12.80 \\
\bottomrule
\end{tabular}
}
\end{table*}

\begin{table*}[t!]
\caption{Results (\%) of PACS on ResNet18~\cite{he2016deep}.
The open-set ratio is $6{:}1$ and asymmetric label noise with a ratio of $50\%$ is selected.}

\label{tab:pacs_res18_50_a}
\centering
\resizebox{1.\linewidth}{!}{
\begin{tabular}{l|ccc|ccc|ccc|ccc|ccc}
\toprule
& \multicolumn{3}{c|}{\textbf{Photo (P)}} & \multicolumn{3}{c|}{\textbf{Art (A)}} & \multicolumn{3}{c|}{\textbf{Cartoon (C)}} & \multicolumn{3}{c|}{\textbf{Sketch (S)}} & \multicolumn{3}{c}{\textbf{Avg}} \\
\textbf{Method} & Acc & H-score & \cellcolor{gray!25}OSCR & Acc & H-score & \cellcolor{gray!25}OSCR & Acc & H-score & \cellcolor{gray!25}OSCR & Acc & H-score & \cellcolor{gray!25}OSCR & Acc & H-score & \cellcolor{gray!25}OSCR \\

\midrule

TCL~\cite{huang2023twin}  & 15.83&3.33	&\cellcolor{gray!25}10.69    & 35.21 & 30.54 & \cellcolor{gray!25}20.96 & 26.46 & 24.24 & \cellcolor{gray!25}14.88 & 20.35 & 1.2 & \cellcolor{gray!25}2.63 & 24.46 & 14.83 & \cellcolor{gray!25}12.29 \\

NPN~\cite{sheng2024adaptive} & 44.43 & 37.82 & \cellcolor{gray!25}28.94 & 38.65 & 39.48 & \cellcolor{gray!25}23.11 & 31.92 & 20.12 & \cellcolor{gray!25}13.49 & 23.24 & 11.25 & \cellcolor{gray!25}4.29 & 34.56 & 27.17 & \cellcolor{gray!25}17.46 \\

BadLabel~\cite{zhang2024badlabel} & 37.16 & 44.56 & \cellcolor{gray!25}35.70 & 26.89 & 31.98 & \cellcolor{gray!25}24.77 & 31.10 & 34.52 & \cellcolor{gray!25}25.33 & 13.88 & 16.21 & \cellcolor{gray!25}20.38 & 27.26 & 31.82 & \cellcolor{gray!25}26.54 \\

DISC~\cite{li2023disc} & 46.20 & 42.31 & \cellcolor{gray!25}38.82 & 44.28 & 42.69 & \cellcolor{gray!25}34.41 & 42.70 & 35.22 & \cellcolor{gray!25}26.28 & 32.77 & 2.77 & \cellcolor{gray!25}22.37 & 41.49 & 30.75 & \cellcolor{gray!25}30.47 \\

LSL~\cite{kim2024learning}  & 27.30 & 19.63 & \cellcolor{gray!25}14.32 & 24.27 & 11.56 & \cellcolor{gray!25}13.18 & 25.63 & 14.77 & \cellcolor{gray!25}12.04 & 22.21 & 16.31 & \cellcolor{gray!25}25.07 & 24.85 & 15.57 & \cellcolor{gray!25}16.15 \\
PLM~\cite{zhao2024estimating}  &17.37  &23.20  & \cellcolor{gray!25}14.13 & 25.45 & 24.44 & \cellcolor{gray!25}15.04 & 20.06 & 9.68 & \cellcolor{gray!25}11.93 & 20.48 & 2.60 & \cellcolor{gray!25}3.24 & 20.84 & 14.98 & \cellcolor{gray!25}11.09 \\
\bottomrule
\end{tabular}

}
\end{table*}

\begin{table*}[t!]
\caption{Results (\%) of PACS on ViT-Base~\cite{dosovitskiy2021an}.
The open-set ratio is $6{:}1$ and symmetric label noise with a ratio of $20\%$ is selected.}
\label{tab:pacs_vit_20}
\centering
\resizebox{1.\linewidth}{!}{
\begin{tabular}{l|ccc|ccc|ccc|ccc|ccc}
\toprule
& \multicolumn{3}{c|}{\textbf{Photo (P)}} & \multicolumn{3}{c|}{\textbf{Art (A)}} & \multicolumn{3}{c|}{\textbf{Cartoon (C)}} & \multicolumn{3}{c|}{\textbf{Sketch (S)}} & \multicolumn{3}{c}{\textbf{Avg}} \\
\textbf{Method} & Acc & H-score & \cellcolor{gray!25}OSCR & Acc & H-score & \cellcolor{gray!25}OSCR & Acc & H-score & \cellcolor{gray!25}OSCR & Acc & H-score & \cellcolor{gray!25}OSCR & Acc & H-score & \cellcolor{gray!25}OSCR \\

\midrule

TCL~\cite{huang2023twin}  & 67.77 & 79.17 & \cellcolor{gray!25}67.64 & 64.35 & 63.45 & \cellcolor{gray!25}56.91 & 51.32 & 50.24 & \cellcolor{gray!25}42.30 & 26.43 & \textbf{28.25} & \cellcolor{gray!25}20.54 & 52.47 & 55.28 & \cellcolor{gray!25} 46.85 \\

NPN~\cite{sheng2024adaptive}  & 63.97 & 71.61 & \cellcolor{gray!25}62.53 & 57.22 & 51.43 & \cellcolor{gray!25}45.95 & 47.65 & 40.11 & \cellcolor{gray!25}32.85 & 21.86 & 13.01 & \cellcolor{gray!25}12.71 & 47.68 & 44.04 & \cellcolor{gray!25} 38.51 \\

BadLabel~\cite{zhang2024badlabel}  & 62.66 & 64.81 & \cellcolor{gray!25}57.87 & 51.59 & 60.49 & \cellcolor{gray!25}49.13 & 50.85 & 58.42 & \cellcolor{gray!25}47.75 & 31.15 & 18.54 & \cellcolor{gray!25}29.40 & 49.06 & 50.57 & \cellcolor{gray!25} 46.04 \\

DISC~\cite{li2023disc} & 66.69 & 71.98 & \cellcolor{gray!25}65.67 & 54.10 & 29.80 & \cellcolor{gray!25}41.07 & 53.48 & 41.72 & \cellcolor{gray!25}40.83 & 34.57 & 16.94 & \cellcolor{gray!25}24.16 & 52.21 & 40.11 & \cellcolor{gray!25} 42.93 \\

LSL~\cite{kim2024learning}  & 67.04 & 71.46 & \cellcolor{gray!25}63.82 & 63.10 & 63.42 & \cellcolor{gray!25}58.01 & 53.27 & 54.92 & \cellcolor{gray!25}47.14 & 28.44 & 4.37 & \cellcolor{gray!25}27.51 & 52.96 & 48.54 & \cellcolor{gray!25} 49.12 \\
PLM~\cite{zhao2024estimating}  & 65.73 & 61.01 & \cellcolor{gray!25}59.76 &65.73  & 61.01 & \cellcolor{gray!25}59.76 & 52.71 & 48.17 & \cellcolor{gray!25}46.32 & 27.57 & 17.94 & \cellcolor{gray!25}20.62 & 52.94 & 47.03 & \cellcolor{gray!25} 46.62 \\
\bottomrule
\end{tabular}

}
\end{table*}

\begin{table*}[t!]
\caption{Results (\%) of PACS on ViT-Base~\cite{dosovitskiy2021an}.
The open-set ratio is $6{:}1$ and symmetric label noise with a ratio of $50\%$ is selected.}

\label{tab:pacs_vit_50}
\centering
\resizebox{1.\linewidth}{!}{
\begin{tabular}{l|ccc|ccc|ccc|ccc|ccc}
\toprule
& \multicolumn{3}{c|}{\textbf{Photo (P)}} & \multicolumn{3}{c|}{\textbf{Art (A)}} & \multicolumn{3}{c|}{\textbf{Cartoon (C)}} & \multicolumn{3}{c|}{\textbf{Sketch (S)}} & \multicolumn{3}{c}{\textbf{Avg}} \\
\textbf{Method} & Acc & H-score & \cellcolor{gray!25}OSCR & Acc & H-score & \cellcolor{gray!25}OSCR & Acc & H-score & \cellcolor{gray!25}OSCR & Acc & H-score & \cellcolor{gray!25}OSCR & Acc & H-score & \cellcolor{gray!25}OSCR \\

\midrule

TCL~\cite{huang2023twin}  & 68.17 & 62.70 & \cellcolor{gray!25}67.56 & 64.48 & 67.84 & \cellcolor{gray!25}61.30 & 45.80 & 35.84 & \cellcolor{gray!25}35.68 & 22.29 & 9.28 & \cellcolor{gray!25}3.54 & 50.19 & 43.92 & \cellcolor{gray!25} 42.02 \\

NPN~\cite{sheng2024adaptive}  & 27.71 & 10.79 & \cellcolor{gray!25}8.48 & 39.34 &  33.88 & \cellcolor{gray!25}32.40 & 40.39 & 44.98 & \cellcolor{gray!25}34.61 & 20.80 & 26.62 & \cellcolor{gray!25}19.84 & 32.06 & 29.07 & \cellcolor{gray!25} 23.83 \\

BadLabel~\cite{zhang2024badlabel}  & 51.62 & 61.25 & \cellcolor{gray!25}50.41 & 32.52 & 37.26 & \cellcolor{gray!25}26.67 & 43.58 & 55.85 & \cellcolor{gray!25}42.45 & 31.60 & 28.24 & \cellcolor{gray!25}26.93 & 39.83 & 45.65 & \cellcolor{gray!25} 36.62\\

DISC~\cite{li2023disc} & 57.35 & 52.08 & \cellcolor{gray!25}44.73 & 38.02 & 37.36 & \cellcolor{gray!25}28.09 & 30.84 & 31.69 & \cellcolor{gray!25}22.68 & 20.72 & 16.32 & \cellcolor{gray!25}16.20 & 36.73 & 34.36 & \cellcolor{gray!25} 27.93 \\

LSL~\cite{kim2024learning}  & 65.99 & 68.68 & \cellcolor{gray!25}62.18 & 55.66 & 49.85 & \cellcolor{gray!25}44.97 & 49.20 & 44.94 & \cellcolor{gray!25}36.86 & 29.90 & 6.91 & \cellcolor{gray!25}19.18 & 50.19 & 42.60 & \cellcolor{gray!25} 40.80 \\
PLM~\cite{zhao2024estimating}  & 57.67 & 50.81 & \cellcolor{gray!25}51.87 & 51.41 & 46.57 & \cellcolor{gray!25}41.82 & 39.66 & 39.47 & \cellcolor{gray!25}31.89 & 19.95 & 12.93 & \cellcolor{gray!25}15.75 & 42.17 & 37.45 & \cellcolor{gray!25} 35.33 \\
\bottomrule
\end{tabular}

}
\end{table*}

\begin{table*}[t!]
\caption{Results (\%) of PACS on ViT-Base~\cite{dosovitskiy2021an}.
The open-set ratio is $6{:}1$ and symmetric label noise with a ratio of $80\%$ is selected.}

\label{tab:pacs_vit_80}
\centering
\resizebox{1.\linewidth}{!}{
\begin{tabular}{l|ccc|ccc|ccc|ccc|ccc}
\toprule
& \multicolumn{3}{c|}{\textbf{Photo (P)}} & \multicolumn{3}{c|}{\textbf{Art (A)}} & \multicolumn{3}{c|}{\textbf{Cartoon (C)}} & \multicolumn{3}{c|}{\textbf{Sketch (S)}} & \multicolumn{3}{c}{\textbf{Avg}} \\
\textbf{Method} & Acc & H-score & \cellcolor{gray!25}OSCR & Acc & H-score & \cellcolor{gray!25}OSCR & Acc & H-score & \cellcolor{gray!25}OSCR & Acc & H-score & \cellcolor{gray!25}OSCR & Acc & H-score & \cellcolor{gray!25}OSCR \\

\midrule

TCL~\cite{huang2023twin}  & 22.70 & 25.47 & \cellcolor{gray!25}16.55 & 22.26 & 23.14 & \cellcolor{gray!25}14.42 & 19.75 & 14.23 & \cellcolor{gray!25}6.53 & 28.52 & 10.92 & \cellcolor{gray!25}11.54 & 23.31 & 18.44 & \cellcolor{gray!25} 12.26 \\

NPN~\cite{sheng2024adaptive} & 15.27 & 15.29 & \cellcolor{gray!25}7.9 & 16.07 & 10.88 & \cellcolor{gray!25}11.29 & 19.96 & 17.19 & \cellcolor{gray!25}8.06 & 20.48 & 1.12 & \cellcolor{gray!25}18.43 & 17.95 & 11.12 & \cellcolor{gray!25} 11.42 \\

BadLabel~\cite{zhang2024badlabel}  & 13.33 & 20.96 & \cellcolor{gray!25}13.32 & 24.77 & 32.36 & \cellcolor{gray!25}21.98 & 23.98 & 32.76 & \cellcolor{gray!25}21.91 & 21.60 & 2.36 & \cellcolor{gray!25}\textbf{21.31} & 20.92 & 22.11 & \cellcolor{gray!25} 19.63 \\

DISC~\cite{li2023disc} &22.37 & 1.92 & \cellcolor{gray!25}11.07 & 24.58 & 19.66 & \cellcolor{gray!25}12.81 & 24.08 & 13.40 & \cellcolor{gray!25}14.20 & 20.03 & 4.56 & \cellcolor{gray!25}12.61 & 22.77 & 9.89 & \cellcolor{gray!25} 12.67 \\

LSL~\cite{kim2024learning}  & 10.02 & 3.64 & \cellcolor{gray!25}5.89 & 26.27 & 10.83 & \cellcolor{gray!25}17.29 & 31.87 & 27.63 & \cellcolor{gray!25}19.75 & 25.39 & 8.09 & \cellcolor{gray!25}7.83 & 23.39 & 12.55 & \cellcolor{gray!25} 12.69 \\
PLM~\cite{zhao2024estimating}  & 29.32 & 5.93 & \cellcolor{gray!25}1.34 & 26.77 & 28.58 & \cellcolor{gray!25}18.39 & 25.27 & 25.69 & \cellcolor{gray!25}16.89 & 24.22 & 3.33 & \cellcolor{gray!25}2.25 & 26.4 & 15.88 & \cellcolor{gray!25} 9.72 \\
\bottomrule
\end{tabular}
}
\end{table*}

\begin{table*}[t!]
\caption{Results (\%) of PACS on ViT-Base~\cite{dosovitskiy2021an}. The open-set ratio is $6{:}1$ and asymmetric label noise with a ratio of $50\%$ is selected.}

\label{tab:pacs_vit_50a}
\centering
\resizebox{1.\linewidth}{!}{
\begin{tabular}{l|ccc|ccc|ccc|ccc|ccc}
\toprule
& \multicolumn{3}{c|}{\textbf{Photo (P)}} & \multicolumn{3}{c|}{\textbf{Art (A)}} & \multicolumn{3}{c|}{\textbf{Cartoon (C)}} & \multicolumn{3}{c|}{\textbf{Sketch (S)}} & \multicolumn{3}{c}{\textbf{Avg}} \\
\textbf{Method} & Acc & H-score & \cellcolor{gray!25}OSCR & Acc & H-score & \cellcolor{gray!25}OSCR & Acc & H-score & \cellcolor{gray!25}OSCR & Acc & H-score & \cellcolor{gray!25}OSCR & Acc & H-score & \cellcolor{gray!25}OSCR \\

\midrule

TCL~\cite{huang2023twin}  & 51.21 & 56.51 & \cellcolor{gray!25}48.47 & 46.65 & 51.35 & \cellcolor{gray!25}40.80 & 31.92 & 30.94 & \cellcolor{gray!25}22.61 & 25.87 & 20.94 & \cellcolor{gray!25}12.04 & 38.91 & 39.94 & \cellcolor{gray!25}30.98 \\

NPN~\cite{sheng2024adaptive}  & 27.63 & 8.86 & \cellcolor{gray!25}6.32 & 32.40 & 19.00 & \cellcolor{gray!25}15.72 & 21.66 & 16.87 & \cellcolor{gray!25}14.25 & 20.48 & 26.15 & \cellcolor{gray!25}16.50 & 25.54 & 17.72 & \cellcolor{gray!25}13.20 \\

BadLabel~\cite{zhang2024badlabel}  & 46.12 & 58.96 & \cellcolor{gray!25}45.77 & 35.33 & 34.30 & \cellcolor{gray!25}34.83 & 26.92 & 39.52 & \cellcolor{gray!25}26.18 & 20.70 & \textbf{31.32} & \cellcolor{gray!25}20.66 & 32.27 & 41.03 & \cellcolor{gray!25}31.86 \\

DISC~\cite{li2023disc} & 47.17 & 17.08 & \cellcolor{gray!25}9.77 & 24.64 & 19.96 & \cellcolor{gray!25}10.75 & 20.22 & 12.84 & \cellcolor{gray!25}9.22 & 23.93 & 4.31 & \cellcolor{gray!25}15.10 & 28.99 & 13.55 & \cellcolor{gray!25}11.21 \\

LSL~\cite{kim2024learning}  & 49.52 & 5.63 & \cellcolor{gray!25}15.99 & 37.52 & 34.68 & \cellcolor{gray!25}26.68 & 28.16 & 27.60 & \cellcolor{gray!25}18.13 & 28.16 & 27.60 & \cellcolor{gray!25}18.13 & 35.84 & 23.88 & \cellcolor{gray!25}19.73 \\
PLM~\cite{zhao2024estimating}  & 20.51 & 23.01 & \cellcolor{gray!25}18.15 & 20.17 & 13.41 & \cellcolor{gray!25}8.08 & 37.85 & 33.88 & \cellcolor{gray!25}26.43 & 20.59 & 24.16 & \cellcolor{gray!25}17.26 & 24.78 & 23.62 & \cellcolor{gray!25}17.48 \\
\bottomrule
\end{tabular}

}
\end{table*}

\begin{table*}[t!]
\caption{Results ($\%$) of DigitsDG on ConvNet~\cite{zhou2021domain}, where symmetric label noise with a ratio of $20\%$ is selected.}
\label{tab:dg_20}
\centering
\resizebox{1.\linewidth}{!}{
\begin{tabular}{l|ccc|ccc|ccc|ccc|ccc}
\toprule
& \multicolumn{3}{c|}{\textbf{mnist}} & \multicolumn{3}{c|}{\textbf{mnist$_m$}} & \multicolumn{3}{c|}{\textbf{syn}} & \multicolumn{3}{c|}{\textbf{svhn}} & \multicolumn{3}{c}{\textbf{Avg}} \\
\textbf{Method} & Acc & H-score & \cellcolor{gray!25}OSCR & Acc & H-score & \cellcolor{gray!25}OSCR & Acc & H-score & \cellcolor{gray!25}OSCR & Acc & H-score & \cellcolor{gray!25}OSCR & Acc & H-score & \cellcolor{gray!25}OSCR \\

\midrule
NPN~\cite{sheng2024adaptive} & 82.28 & 29.42 & \cellcolor{gray!25}70.68 & 32.56 & 28.22 & \cellcolor{gray!25}22.74 & 21.78 & 22.54 & \cellcolor{gray!25}14.41 & 17.33 & 11.95 & \cellcolor{gray!25}9.89 & 38.49 & 23.03 & \cellcolor{gray!25}29.43 \\

BadLabel~\cite{zhang2024badlabel}  & 63.25 & 52.49 & \cellcolor{gray!25}61.18 & 30.00 & 34.82 & \cellcolor{gray!25}27.42 & 50.17 & 51.19 & \cellcolor{gray!25}46.06 & 41.82 & 16.75 & \cellcolor{gray!25}41.82 & 46.31 & 38.81 & \cellcolor{gray!25}44.12 \\

\midrule 

ODGNet~\cite{bose2023beyond}& 90.33 & 50.84 & \cellcolor{gray!25}71.38 & 59.28 & 26.10 & \cellcolor{gray!25}\textbf{43.83} & 70.11 & 53.83 & \cellcolor{gray!25}49.54 & 55.81 & 32.71 & \cellcolor{gray!25}39.54 & 68.88 & 40.87 & \cellcolor{gray!25}51.07 \\

MLDG \cite{chen2021adversarial} & 90.67 & 27.60 & \cellcolor{gray!25}80.46 & 57.89 & 48.38 & \cellcolor{gray!25}42.77 & 60.33 & 39.95 & \cellcolor{gray!25}41.30 & 53.75 & 6.08 & \cellcolor{gray!25}35.20 & 65.66 & 30.50 & \cellcolor{gray!25}49.93 \\

MEDIC-cls~\cite{wang2023generalizable} & 22.08 & 8.71 & \cellcolor{gray!25}5.89 & 21.33 & 20.09 & \cellcolor{gray!25}9.94 & 23.24 & 9.83 & \cellcolor{gray!25}13.28 & 13.72 & 10.29 & \cellcolor{gray!25}6.63 & 20.09 & 12.23 & \cellcolor{gray!25}8.94 \\

MEDIC-bcls~\cite{wang2023generalizable} & 22.08 & 12.31 & \cellcolor{gray!25}5.57 & 21.33 & 16.19 & \cellcolor{gray!25}10.11 & 23.24 & 12.64 & \cellcolor{gray!25}5.41 & 13.72 & 13.47 & \cellcolor{gray!25}5.94 & 20.09 &  13.65 & \cellcolor{gray!25}6.76 \\
\bottomrule
\end{tabular}}
\end{table*}

\begin{table*}[t!]
\caption{Results (\%) of DigitsDG on ConvNet~\cite{zhou2021domain}, where symmetric label noise with a ratio of $50\%$ is selected.}

\label{tab:dg_50}
\centering
\resizebox{1.\linewidth}{!}{
\begin{tabular}{l|ccc|ccc|ccc|ccc|ccc}
\toprule
& \multicolumn{3}{c|}{\textbf{mnist}} & \multicolumn{3}{c|}{\textbf{mnist$_m$}} & \multicolumn{3}{c|}{\textbf{syn}} & \multicolumn{3}{c|}{\textbf{svhn}} & \multicolumn{3}{c}{\textbf{Avg}} \\
\textbf{Method} & Acc & H-score & \cellcolor{gray!25}OSCR & Acc & H-score & \cellcolor{gray!25}OSCR & Acc & H-score & \cellcolor{gray!25}OSCR & Acc & H-score & \cellcolor{gray!25}OSCR & Acc & H-score & \cellcolor{gray!25}OSCR \\

\midrule

NPN~\cite{sheng2024adaptive} & 68.11 & 25.80 & \cellcolor{gray!25}49.14 & 28.31 & 28.29 & \cellcolor{gray!25}18.65 & 45.78 & 32.17 & \cellcolor{gray!25}31.15 & 28.42 & 1.38 & \cellcolor{gray!25}16.37 & 42.66 & 21.91 & \cellcolor{gray!25}28.83 \\

BadLabel~\cite{zhang2024badlabel}  & 63.31 &35.12  & \cellcolor{gray!25}57.47& 42.28 & 42.69 & \cellcolor{gray!25}38.44 & 21.36 & 25.07 & \cellcolor{gray!25}18.30 & 19.83 & 4.88 & \cellcolor{gray!25}19.28 & 36.70 & 26.94 & \cellcolor{gray!25}33.37 \\

\midrule

ODGNet~\cite{bose2023beyond}& 71.25 & 22.26 & \cellcolor{gray!25}49.81 & 59.22 & 36.88 & \cellcolor{gray!25}44.86 & \textbf{61.39} & \textbf{43.65} & \cellcolor{gray!25}\textbf{42.41} & 51.75 & 19.70 & \cellcolor{gray!25}35.38 & 60.90 & 30.62 & \cellcolor{gray!25}43.12 \\

MLDG~\cite{shu2019meta} & 62.72 & \textbf{54.21} & \cellcolor{gray!25}50.52 & 48.94 & 41.54 & \cellcolor{gray!25}35.35 & 43.53  & 33.15 & \cellcolor{gray!25}29.09 & 24.61 & 6.48 & \cellcolor{gray!25}14.29 & 44.95 & 33.85 & \cellcolor{gray!25}32.31 \\

MEDIC-cls~\cite{wang2023generalizable} & 22.39 & 0.17 & \cellcolor{gray!25}2.81 & 25.89 & 22.88 & \cellcolor{gray!25}12.16 & 12.78 & 12.10 & \cellcolor{gray!25}4.73 & 7.61 & 5.70 & \cellcolor{gray!25}2.35 & 17.17 & 10.21 & \cellcolor{gray!25}5.51 \\
\bottomrule
\end{tabular}

}
\end{table*}

\begin{table*}[t!]
\caption{Results (\%) of DigitsDG on ConvNet~\cite{zhou2021domain}, where symmetric label noise with a ratio of $80\%$ is selected..}
\label{tab:dg_80}
\centering
\resizebox{1.\linewidth}{!}{
\begin{tabular}{l|ccc|ccc|ccc|ccc|ccc}
\toprule
& \multicolumn{3}{c|}{\textbf{mnist}} & \multicolumn{3}{c|}{\textbf{mnist$_m$}} & \multicolumn{3}{c|}{\textbf{syn}} & \multicolumn{3}{c|}{\textbf{svhn}} & \multicolumn{3}{c}{\textbf{Avg}} \\
\textbf{Method} & Acc & H-score & \cellcolor{gray!25}OSCR & Acc & H-score & \cellcolor{gray!25}OSCR & Acc & H-score & \cellcolor{gray!25}OSCR & Acc & H-score & \cellcolor{gray!25}OSCR & Acc & H-score & \cellcolor{gray!25}OSCR \\

\midrule

NPN~\cite{sheng2024adaptive} & 16.67 & 0.01 & \cellcolor{gray!25}9.65 & 18.61 & 13.38 & \cellcolor{gray!25}10.15 & 17.78 & 18.12 & \cellcolor{gray!25}9.29 & 16.67 & 0.55 & \cellcolor{gray!25}8.21 & 17.43 & 8.02 & \cellcolor{gray!25}9.33 \\

BadLabel~\cite{zhang2024badlabel} & 18.58 & 5.84 & \cellcolor{gray!25}7.39 &17.39  & 15.13 & \cellcolor{gray!25}8.99 & 16.67 & 0.41 & \cellcolor{gray!25}8.36 & 17.11 & 0.22 & \cellcolor{gray!25}8.44 & 17.44 & 5.40 & \cellcolor{gray!25}8.30 \\

\midrule

ODGNet~\cite{bose2023beyond} & 16.19 & 1.51  & \cellcolor{gray!25}10.37 & 17.28 & 11.80 & \cellcolor{gray!25}10.64 & 18.47 & 17.04 & \cellcolor{gray!25}8.49 & 16.72 & 14.89 & \cellcolor{gray!25}8.59 &17.17 & 11.31 & \cellcolor{gray!25} 9.52\\

MLDG~\cite{shu2019meta} & 16.06 & 6.70 & \cellcolor{gray!25}9.69 & 18.58 & 3.26 & \cellcolor{gray!25}9.27 & 16.94 & 6.81 & \cellcolor{gray!25}8.20 & 17.33 & 6.88 & \cellcolor{gray!25}\textbf{8.84} & 17.23 & 5.91 & \cellcolor{gray!25} 9.00\\

MEDIC-cls~\cite{wang2023generalizable} & 21.17 & 3.51 & \cellcolor{gray!25}11.37 & 18.75 & 16.52 & \cellcolor{gray!25}7.77 & 15.81 & 4.48 & \cellcolor{gray!25}7.54 & 17.11  & 4.86 & \cellcolor{gray!25}8.64 & 18.21 & 7.34 & \cellcolor{gray!25}8.83 \\

MEDIC-bcls~\cite{wang2023generalizable} & 21.17 & 7.21 & \cellcolor{gray!25}8.32 & 16.83 & 13.25 & \cellcolor{gray!25}8.81 & 15.81 & 4.44 & \cellcolor{gray!25}7.70 & 17.11 & 4.96 & \cellcolor{gray!25}8.38 & 17.73 & 7.47 & \cellcolor{gray!25}8.30 \\
\bottomrule
\end{tabular}

}
\end{table*}

\begin{table*}[h!]
\caption{Results (\%) of DigitsDG on ConvNet~\cite{zhou2021domain}, where asymmetric label noise with a ratio of $50\%$ is selected.}

\label{tab:dg_50a}
\centering
\resizebox{1.\linewidth}{!}{
\begin{tabular}{l|ccc|ccc|ccc|ccc|ccc}
\toprule
& \multicolumn{3}{c|}{\textbf{mnist}} & \multicolumn{3}{c|}{\textbf{mnist$_m$}} & \multicolumn{3}{c|}{\textbf{syn}} & \multicolumn{3}{c|}{\textbf{svhn}} & \multicolumn{3}{c}{\textbf{Avg}} \\
\textbf{Method} & Acc & H-score & \cellcolor{gray!25}OSCR & Acc & H-score & \cellcolor{gray!25}OSCR & Acc & H-score & \cellcolor{gray!25}OSCR & Acc & H-score & \cellcolor{gray!25}OSCR & Acc & H-score & \cellcolor{gray!25}OSCR \\

\midrule

NPN~\cite{sheng2024adaptive} & 71.08 & 24.46 & \cellcolor{gray!25}\textbf{61.45} & 54.58 & 41.21 & \cellcolor{gray!25}38.79 & 55.92 & 42.95 & \cellcolor{gray!25}35.03 & 51.97 & 11.28 & \cellcolor{gray!25}34.23 &58.39 & 29.98 & \cellcolor{gray!25}42.38 \\

BadLabel~\cite{zhang2024badlabel}  & 53.00 & 38.77 & \cellcolor{gray!25}43.34 & 33.94 & 27.79 & \cellcolor{gray!25}30.06 & 37.64 & 44.59 & \cellcolor{gray!25}33.47 & 27.61 & \textbf{20.90} & \cellcolor{gray!25}26.18 & 38.05 & 33.01 & \cellcolor{gray!25}33.26 \\

\midrule

ODGNet~\cite{bose2023beyond} & 53.67 & 48.23 & \cellcolor{gray!25}41.31 & 39.36 & 27.19 & \cellcolor{gray!25}28.08 & 51.17 & 24.92 & \cellcolor{gray!25}34.91 & 39.64 & 5.51 & \cellcolor{gray!25}25.32 &45.96&	26.46	  & \cellcolor{gray!25}32.41   \\

MLDG~\cite{shu2019meta} & 68.17 & 23.34 & \cellcolor{gray!25}55.68 & 56.47 & \textbf{40.89} & \cellcolor{gray!25}40.97 & 56.31 & 41.94 & \cellcolor{gray!25}38.02 & 47.81 & 11.35 & \cellcolor{gray!25}31.60 &57.19 &	29.38	  & \cellcolor{gray!25}41.57  \\

MEDIC-cls~\cite{wang2023generalizable} & 19.86 & 14.41 & \cellcolor{gray!25}11.22 & 19.75 & 15.93 & \cellcolor{gray!25}8.03 & 9.83 & 8.28 & \cellcolor{gray!25}2.88 & 16.31 & 5.78 & \cellcolor{gray!25}7.36 &16.44  &	11.10	  & \cellcolor{gray!25}7.37 \\
\bottomrule
\end{tabular}
}
\end{table*}

\subsection{Analysis on the Benchmarks}
\label{sec:benchmarks}
\noindent\textbf{Analysis on benchmark on PACS:} 
We construct our benchmark according to the aforementioned label noise setting in Section~\ref{sec:label_noise} on both of the PACS and DigitsDG datasets, where we deploy ResNet18~\cite{he2016deep} (Table~\ref{tab:pacs_res18_noise_20} to Table~\ref{tab:pacs_res18_50_a}) and ViT-Base~\cite{dosovitskiy2021an} (Table~\ref{tab:pacs_vit_20} to Table~\ref{tab:pacs_vit_50a}) on the PACS dataset and ConvNet~\cite{zhou2021domain} on the DigitsDG dataset (Table~\ref{tab:dg_20} to Table~\ref{tab:dg_50a}).

First, we observe that with the increase of the label noise ratio, there are degradations of the OSDG performance on all the leveraged baselines, according to the Table~\ref{tab:pacs_res18_noise_20}, Table~\ref{tab:pacs_res18_50}, and Table~\ref{tab:pacs_res18_80}, where symmetric label noise with ratios of $20\%$, $50\%$, and $80\%$ are used on PACS and ResNet18~\cite{he2016deep}. Among all the approaches derived from the label noise learning field, BadLabel~\cite{zhang2024badlabel}, NPN~\cite{sheng2024adaptive}, and DISC~\cite{li2023disc} show promising performances when tackling OSDG-NL task, where BadLabel~\cite{zhang2024badlabel} delivers $47.61\%$, $38.78\%$, $21.02\%$, and $27.26\%$ of closed-set accuracy, $46.55\%$, $43.76\%$, $16.77\%$, and $31.82\%$ in terms of H-score, and $42.22\%$, $32.70\%$, $16.42\%$, and $26.54\%$ in terms of OSCR on PACS with symmetric label noise ratio $20\%$ (Table~\ref{tab:pacs_res18_noise_20}), $50\%$ (Table~\ref{tab:pacs_res18_50}), $80\%$ (Table~\ref{tab:pacs_res18_80}) and asymmetric label noise ratio $50\%$ (Table~\ref{tab:pacs_res18_50_a}) on ResNet18~\cite{he2016deep} backbone, respectively.

Among all the approaches from the OSDG field, GAN-based method, \textit{i.e.}, ODGNet~\cite{bose2023beyond}, and meta-learning-based methods, \textit{i.e.}, MLDG~\cite{shu2019meta} and MEDIC~\cite{wang2023generalizable}, show good performance when dealing with various adopted label noises, where MEDIC~\cite{wang2023generalizable} delivers $52.66\%$, $50.27\%$, $26.40\%$, and $37.79\%$ of closed-set accuracy, $49.25\%$, $40.24\%$, $16.25\%$, and $31.73\%$ in terms of H-score, and $42.76\%$, $34.62\%$, $11.17\%$, and $23.95\%$ in terms of OSCR on PACS with symmetric label noise ratio $20\%$ (Table~\ref{tab:pacs_res18_noise_20}), $50\%$ (Table~\ref{tab:pacs_res18_50}), $80\%$ (Table~\ref{tab:pacs_res18_80}) and asymmetric label noise ratio $50\%$ (Table~\ref{tab:pacs_res18_50_a}) on the ResNet18 backbone using binary classification head for each category (bcls), respectively. 
EBiL-HaDS-cls~\cite{peng2024advancing} and EBiL-HaDS-bcls~\cite{peng2024advancing} perform reasonably well under $20\%$ and $50\%$ symmetric noise, particularly in the \textit{Photo} domain, but both variants exhibit notable degradation as noise severity increases. Under $80\%$ symmetric noise and $50\%$ asymmetric noise, their average OSCR values fall below $28\%$, indicating limited robustness in high-noise scenarios.

The label noise learning approach, BadLabel~\cite{zhang2024badlabel}, outperforms OSDG approaches in the scenario where a large label noise ratio is adopted. 
Similar observations can also be found in the experiments conducted with ViT-Base~\cite{dosovitskiy2021an} backbone shown in Table~\ref{tab:pacs_vit_20} to Table~\ref{tab:pacs_vit_50a}.

When we compare Table~\ref{tab:pacs_res18_50} and Table~\ref{tab:pacs_res18_50_a}, we find that asymmetric label noise is much more challenging compared with the symmetric label noise under the same label noise ratio $50\%$. Among all the leveraged baselines using ResNet18~\cite{he2016deep} as the backbone, DISC~\cite{li2023disc} shows $30.47\%$ in terms of OSCR under asymmetric label noise but still yields $0.94\%$ performance degradation when compared with the symmetric scenario.
MLDG~\cite{chen2021adversarial} achieves better performance compared with the other leveraged baselines on ViT-Base~\cite{dosovitskiy2021an} backbone when facing asymmetric label noise with a ratio of $50\%$, where $40.70\%$ in terms of OSCR is delivered, as shown in Table~\ref{tab:pacs_vit_50a}.
The aforementioned observation highlights the challenge posed by label noise in the OSDG task, a problem that existing label noise learning methods and OSDG approaches have yet to effectively address.

The proposed approach, HyProMeta, surpasses all baseline methods by employing label noise-agnostic meta-learning and new-category-aware prompt learning. Specifically, category prototypes derived from the hyperbolic space guide data partitioning during the meta-task assignment, while learnable prompt enhances mixed-category samples, thereby improving the generalizability of the resulting model.
HyProMeta achieves superior performance, outperforming the strongest baselines by significant margins across various backbones and label noise ratios on PACS. For instance, under a symmetric label noise ratio of $20\%$, HyProMeta surpasses NPN~\cite{sheng2024adaptive} by $4.32\%$, $8.69\%$, and $5.23\%$ in terms of closed-set accuracy, H-score, and OSCR, respectively, when using ResNet18~\cite{he2016deep} as the feature learning backbone, as shown in Table~\ref{tab:pacs_res18_noise_20}. 
HyProMeta consistently outperforms both EBiL-HaDS-cls~\cite{peng2024advancing} and EBiL-HaDS-bcls~\cite{peng2024advancing} across all noise settings on PACS with ResNet18~\cite{he2016deep}, achieving the highest average OSCR under $20\%$, $50\%$, $80\%$ symmetric, and $50\%$ asymmetric noise, with respective margins of up to $+5\%$ in terms of OSCR. The performance gap becomes especially prominent as noise severity increases, demonstrating HyProMeta’s superior robustness to label corruption and domain shift.

When adopting a more powerful backbone, \textit{i.e.}, ViT-Base~\cite{dosovitskiy2021an}, both the baseline methods and our proposed approach achieve further performance improvements. Specifically, HyProMeta achieves closed-set accuracy of $59.65\%$, $58.68\%$, $37.06\%$, and $49.99\%$; H-score of $60.06\%$, $59.33\%$, $29.09\%$, and $48.47\%$; and OSCR of $54.97\%$, $52.91\%$, $25.26\%$, and $43.44\%$ on PACS using ViT-Base~\cite{dosovitskiy2021an} as the backbone, significantly outperforming existing approaches, as presented in Table~\ref{tab:pacs_vit_20} to Table~\ref{tab:pacs_vit_50a}.

These results emphasize the robust generalizability of the proposed HyProMeta method across various types of label noise and feature learning backbones.

\noindent\textbf{Analysis on benchmark on DigitsDG:} To further assess the cross-dataset generalizability of our proposed approach and the baselines for the OSDG-NL task, we selected several baselines that demonstrated promising performance on the PACS dataset to construct a benchmark on another OSDG dataset, \textit{i.e.}, DigitsDG. The results, presented in Table~\ref{tab:dg_20} to Table~\ref{tab:dg_50a}, consider symmetric label noise with ratios of $20\%$, $50\%$, and $80\%$, as well as asymmetric label noise with a ratio of $50\%$.

We first observe that the MEDIC approach~\cite{wang2023generalizable} fails to perform well on DigitsDG when using ConvNet~\cite{zhou2021domain} as the backbone. This demonstrates that MEDIC relies heavily on high-quality cues from the source domains to ensure its effectiveness, particularly on small datasets and when employing lightweight backbones. Among all the baselines evaluated on the DigitsDG dataset, ODGNet~\cite{bose2023beyond} achieves the best performance under symmetric label noise, while NPN~\cite{sheng2024adaptive} demonstrates superior performance under asymmetric label noise.

\begin{table*}[h!]
\caption{Module ablation on PACS using ResNet18~\cite{he2016deep} under symmetric label noise ratio $50\%$.}

\label{tab:ablation_study_component}
\centering
\resizebox{1.\linewidth}{!}{
\begin{tabular}{l|ccc|ccc|ccc|ccc|ccc}
\toprule
& \multicolumn{3}{c|}{\textbf{Photo (P)}} & \multicolumn{3}{c|}{\textbf{Art Painting (A)}} & \multicolumn{3}{c|}{\textbf{Cartoon (C)}} & \multicolumn{3}{c|}{\textbf{Sketch (S)}} & \multicolumn{3}{c}{\textbf{Avg}} \\
\textbf{Method} & Acc & H-score & \cellcolor{gray!25}OSCR & Acc & H-score & \cellcolor{gray!25}OSCR & Acc & H-score & \cellcolor{gray!25}OSCR & Acc & H-score & \cellcolor{gray!25}OSCR & Acc & H-score & \cellcolor{gray!25}OSCR \\

\midrule
\multicolumn{16}{c}{\textbf{Symmetric label noise ratio $20\%$}}\\
\midrule
w/o HYB-Meta & 60.26 & 66.41 & \cellcolor{gray!25}56.94 & 57.54 & 46.41 & \cellcolor{gray!25}43.21 & 55.08 & 51.07 & \cellcolor{gray!25}45.10 & 28.63 & 19.45 & \cellcolor{gray!25}17.22 & 50.38 & 45.83 & \cellcolor{gray!25}40.62 \\
w/o NCA-Prompt  & 62.60 & 70.31 & \cellcolor{gray!25}60.86 & 53.85 & 49.84 & \cellcolor{gray!25}42.27 & 53.66& 48.95& \cellcolor{gray!25}42.18 & 37.38& 29.97 & \cellcolor{gray!25}22.59& 51.87 & 49.77 & \cellcolor{gray!25}41.98 \\
\midrule
Ours  & \textbf{66.00} & \textbf{76.84}  & \cellcolor{gray!25}\textbf{66.00} & \textbf{59.91} & \textbf{56.89} & \cellcolor{gray!25}\textbf{49.93} & \textbf{59.41} & \textbf{56.47} & \cellcolor{gray!25}\textbf{50.42} & \textbf{39.16} & \textbf{34.76} & \cellcolor{gray!25}\textbf{26.46} & \textbf{56.12} & \textbf{56.24} & \cellcolor{gray!25} \textbf{48.20}\\
\midrule
\multicolumn{16}{c}{\textbf{Symmetric label noise ratio $50\%$}}\\
\midrule
w/o HYB-Meta & 61.79 & 71.74 & \cellcolor{gray!25}62.99 & 57.16 & 51.41 & \cellcolor{gray!25}43.95 & 44.66&37.42 & \cellcolor{gray!25}30.50 & 21.78& 21.77 & \cellcolor{gray!25}13.96& 46.35 & 45.59 & \cellcolor{gray!25}37.85 \\
\bottomrule
\end{tabular}
}
\end{table*}
\begin{table*}[h!]
\caption{Ablation study of NCA-Prompt on symmetric label noise $50\%$ on PACS dataset using ResNet18~\cite{he2016deep}.}

\label{tab:abl_nca}
\centering
\resizebox{1.\linewidth}{!}{
\begin{tabular}{l|ccc|ccc|ccc|ccc|ccc}
\toprule
& \multicolumn{3}{c|}{\textbf{Photo (P)}} & \multicolumn{3}{c|}{\textbf{Art Painting (A)}} & \multicolumn{3}{c|}{\textbf{Cartoon (C)}} & \multicolumn{3}{c|}{\textbf{Sketch (S)}} & \multicolumn{3}{c}{\textbf{Avg}} \\
\textbf{Method} & Acc & H-score & \cellcolor{gray!25}OSCR & Acc & H-score & \cellcolor{gray!25}OSCR & Acc & H-score & \cellcolor{gray!25}OSCR & Acc & H-score & \cellcolor{gray!25}OSCR & Acc & H-score & \cellcolor{gray!25}OSCR \\

\midrule

Asynchronous Opt. & 62.84 & 69.12 & \cellcolor{gray!25}60.44 & 45.22 & 43.55 & \cellcolor{gray!25}36.16 & 50.28 & 49.15 & \cellcolor{gray!25}40.88 & 31.04 & 21.41 & \cellcolor{gray!25}15.72 & 47.35 & 45.81 & \cellcolor{gray!25}35.30 \\
Adversarial Opt. & 61.55 & 66.07 & \cellcolor{gray!25}58.77 & 51.66 & 52.73 & \cellcolor{gray!25}43.72 & 50.23 & 43.80 & \cellcolor{gray!25}35.43 & 31.60 & 23.99 & \cellcolor{gray!25}14.38 & 48.76 & 46.65 & \cellcolor{gray!25}38.08 \\
Fixed Crop. & 64.50 & 69.59 & \cellcolor{gray!25}62.00 & 57.16 & \textbf{53.74} & \cellcolor{gray!25}45.91 & 47.19 & 39.61& \cellcolor{gray!25}30.26 & 32.53 & 5.44 & \cellcolor{gray!25}19.08 & 50.35 & 42.10 & \cellcolor{gray!25}39.31 \\
\midrule
Ours   & \textbf{65.19} &\textbf{73.38}  & \cellcolor{gray!25}\textbf{63.79} & \textbf{60.85} & 52.51 & \cellcolor{gray!25}\textbf{46.97} & \textbf{51.99} & \textbf{49.55} & \cellcolor{gray!25}\textbf{41.71} & \textbf{39.06} & \textbf{33.53} & \cellcolor{gray!25}\textbf{23.44} & \textbf{54.27} & \textbf{52.24} & \cellcolor{gray!25}\textbf{43.98} \\
\bottomrule
\end{tabular}
}
\end{table*}

\begin{table*}[h!]
\caption{Ablation study of HYB-Meta on symmetric label noise $50\%$ on PACS dataset using ResNet18~\cite{he2016deep}.}

\label{tab:abl_hyb}
\centering
\resizebox{1.\linewidth}{!}{
\begin{tabular}{l|ccc|ccc|ccc|ccc|ccc}
\toprule
& \multicolumn{3}{c|}{\textbf{Photo (P)}} & \multicolumn{3}{c|}{\textbf{Art Painting (A)}} & \multicolumn{3}{c|}{\textbf{Cartoon (C)}} & \multicolumn{3}{c|}{\textbf{Sketch (S)}} & \multicolumn{3}{c}{\textbf{Avg}} \\
\textbf{Method} & Acc & H-score & \cellcolor{gray!25}OSCR & Acc & H-score & \cellcolor{gray!25}OSCR & Acc & H-score & \cellcolor{gray!25}OSCR & Acc & H-score & \cellcolor{gray!25}OSCR & Acc & H-score & \cellcolor{gray!25}OSCR \\

\midrule
w/o label corr. & 63.00 & 70.08 & \cellcolor{gray!25}61.75 & 55.28 & \textbf{55.01} & \cellcolor{gray!25}47.94 & 47.34 & 38.25& \cellcolor{gray!25}34.66 & 32.25& 19.69 & \cellcolor{gray!25}21.45 & 49.47 & 45.76 & \cellcolor{gray!25}41.45 \\
w/o cross domain & 60.26 & 67.26 & \cellcolor{gray!25}58.67 & 53.41 & 50.89 & \cellcolor{gray!25}43.02 & 49.66 & 48.29& \cellcolor{gray!25}39.77 &31.97 & 16.80 & \cellcolor{gray!25}14.55 & 48.83 & 45.81 & \cellcolor{gray!25}39.00 \\
w/ euc. prototype & 62.36 & 71.08 & \cellcolor{gray!25}61.05 & 55.35  & 52.36 & \cellcolor{gray!25}45.83 & 49.92 &45.70 & \cellcolor{gray!25}37.57 & 27.91& 27.50 & \cellcolor{gray!25}18.81 & 48.89 & 49.16 & \cellcolor{gray!25}40.82 \\
\midrule
Ours   & \textbf{65.19} &\textbf{73.38}  & \cellcolor{gray!25}\textbf{63.79} & \textbf{60.85} & 52.51 & \cellcolor{gray!25}\textbf{46.97} & \textbf{51.99} & \textbf{49.55} & \cellcolor{gray!25}\textbf{41.71} & \textbf{39.06} & \textbf{33.53} & \cellcolor{gray!25}\textbf{23.44} & \textbf{54.27} & \textbf{52.24} & \cellcolor{gray!25}\textbf{43.98} \\
\bottomrule
\end{tabular}
}
\end{table*}

\begin{table*}[t!]
\caption{Results ($\%$) of PACS using ViT-Base~\cite{dosovitskiy2021an} with three different pretraining strategies, \textit{i.e.}, ImageNet~\cite{deng2009imagenet} pretraining, CLIP~\cite{radford2021learning} pretraining, and random initialization. The open-set ratio is $6{:}1$. The average performances on all target domains are shown.}
\label{tab:pretraining_comparison}
\centering
\resizebox{0.95\linewidth}{!}{
\begin{tabular}{l|c|ccc|ccc|ccc|ccc}
\toprule
& & \multicolumn{3}{c|}{\textbf{20\% sym}} & \multicolumn{3}{c|}{\textbf{50\% sym}} & \multicolumn{3}{c|}{\textbf{80\% sym}} & \multicolumn{3}{c}{\textbf{50\% asym}} \\
\textbf{Method} &Init & Acc & H-score & \cellcolor{gray!25}OSCR & Acc & H-score & \cellcolor{gray!25}OSCR & Acc & H-score & \cellcolor{gray!25}OSCR & Acc & H-score & \cellcolor{gray!25}OSCR \\

\midrule

NPN~\cite{sheng2024adaptive}  & \multirow{8}{*}{\begin{sideways}ImageNet~\cite{deng2009imagenet}\end{sideways}} & 47.68 & 44.04 & \cellcolor{gray!25}38.51 & 32.06 & 29.07 & \cellcolor{gray!25}23.83 & 17.95 & 11.12 & \cellcolor{gray!25}11.42 & 25.54 & 17.72 & \cellcolor{gray!25}13.20 \\

BadLabel~\cite{zhang2024badlabel}  &  & 21.02 & 16.77 & \cellcolor{gray!25}16.42 & 39.83 & 45.65 & \cellcolor{gray!25}36.62 & 20.92 & 22.11 & \cellcolor{gray!25}19.63 & 32.27 & 41.03 & \cellcolor{gray!25}31.86 \\

ODGNet~\cite{bose2023beyond}  &  & 25.31 & 17.41 & \cellcolor{gray!25}13.00 & 55.12 & 53.43 & \cellcolor{gray!25}48.31 & 20.92 & 15.08 & \cellcolor{gray!25}8.27 & 40.69& 39.08 & \cellcolor{gray!25}31.92 \\

MEDIC-cls~\cite{wang2023generalizable}   &  & 26.40 & 19.78 & \cellcolor{gray!25}14.78 & 53.61 & 48.99 & \cellcolor{gray!25}47.08 & 28.97 & 23.03 & \cellcolor{gray!25}16.36 & 42.42 & 43.23 & \cellcolor{gray!25}35.49 \\

MEDIC-bcls~\cite{wang2023generalizable}  &  & 26.40 & 16.25 & \cellcolor{gray!25}11.17 & 53.61 & 40.86 & \cellcolor{gray!25}38.67 & 28.98 & 18.43 & \cellcolor{gray!25}13.10 & 42.42 & 40.01 & \cellcolor{gray!25}33.31 \\

EBiL-HaDS-cls~\cite{peng2024advancing}   & & 28.60 & 23.31 & \cellcolor{gray!25}19.08 & 52.68 & 45.77 & \cellcolor{gray!25}44.39 & 30.87 & 14.26 & \cellcolor{gray!25}18.71 & 40.49 & 40.39 & \cellcolor{gray!25}32.21 \\
\bottomrule
\end{tabular}}

\end{table*}

\begin{table*}[t!]
\caption{Results (\%) of PACS on ViT-Base~\cite{dosovitskiy2021an} using CLIP~\cite{radford2021learning} initialization.
The open-set ratio is $6{:}1$ and symmetric label noise with a ratio of $20\%$ is selected.}
\label{tab:pacs_vit_20_clip}
\centering
\resizebox{1.\linewidth}{!}{
\begin{tabular}{l|ccc|ccc|ccc|ccc|ccc}
\toprule
& \multicolumn{3}{c|}{\textbf{Photo (P)}} & \multicolumn{3}{c|}{\textbf{Art (A)}} & \multicolumn{3}{c|}{\textbf{Cartoon (C)}} & \multicolumn{3}{c|}{\textbf{Sketch (S)}} & \multicolumn{3}{c}{\textbf{Avg}} \\
\textbf{Method} & Acc & H-score & \cellcolor{gray!25}OSCR & Acc & H-score & \cellcolor{gray!25}OSCR & Acc & H-score & \cellcolor{gray!25}OSCR & Acc & H-score & \cellcolor{gray!25}OSCR & Acc & H-score & \cellcolor{gray!25}OSCR \\

\midrule

NPN~\cite{sheng2024adaptive}  & 56.22 & 43.92 & \cellcolor{gray!25}39.26 &  50.03& 34.16 & \cellcolor{gray!25}26.73 & 48.79 & 40.42 & \cellcolor{gray!25}34.48 & 46.30 & 16.23 & \cellcolor{gray!25}30.46  & 50.33 & 33.68 & \cellcolor{gray!25}32.73\\

BadLabel~\cite{zhang2024badlabel}  & 43.46 & 36.78 & \cellcolor{gray!25}29.75 & 36.90 & 34.68& \cellcolor{gray!25}24.97 & 48.84 & 40.99& \cellcolor{gray!25}32.98 & 38.79 & 23.69 & \cellcolor{gray!25}16.44 & 42.00 & 34.04 & \cellcolor{gray!25}26.04 \\

\midrule

ODGNet~\cite{bose2023beyond} & 50.08 & 50.50 & \cellcolor{gray!25}41.64 & 47.22 &41.17 & \cellcolor{gray!25}33.14 & 52.19 & 44.33 & \cellcolor{gray!25}35.13 & 34.28 & 28.91 & \cellcolor{gray!25}23.31 & 45.94 & 41.23 & \cellcolor{gray!25}33.31  \\

MEDIC-cls~\cite{wang2023generalizable} & 58.16 & 49.31 & \cellcolor{gray!25}42.22 & 55.91 & 49.01 & \cellcolor{gray!25}43.34 & 54.98 & 46.80 & \cellcolor{gray!25}39.62 & 48.13 & 36.13 & \cellcolor{gray!25}30.49 & 54.30 & 45.31 & \cellcolor{gray!25}38.92 \\

MEDIC-bcls~\cite{wang2023generalizable}  & 58.16 & 8.80 & \cellcolor{gray!25}34.59 & 55.91 & 12.98& \cellcolor{gray!25}20.06 & 54.98 & 18.84& \cellcolor{gray!25}33.74 & 48.13 & 0.00 & \cellcolor{gray!25}34.37 & 54.30 & 10.16 & \cellcolor{gray!25}30.72 \\
\bottomrule
\end{tabular}

}
\end{table*}

\begin{table*}[t!]
\caption{Results (\%) of PACS on ViT-Base~\cite{dosovitskiy2021an} using CLIP~\cite{radford2021learning} initialization.
The open-set ratio is $6{:}1$ and symmetric label noise with a ratio of $50\%$ is selected.}
\label{tab:pacs_vit_50_clip}
\centering
\resizebox{1.\linewidth}{!}{
\begin{tabular}{l|ccc|ccc|ccc|ccc|ccc}
\toprule
& \multicolumn{3}{c|}{\textbf{Photo (P)}} & \multicolumn{3}{c|}{\textbf{Art (A)}} & \multicolumn{3}{c|}{\textbf{Cartoon (C)}} & \multicolumn{3}{c|}{\textbf{Sketch (S)}} & \multicolumn{3}{c}{\textbf{Avg}} \\
\textbf{Method} & Acc & H-score & \cellcolor{gray!25}OSCR & Acc & H-score & \cellcolor{gray!25}OSCR & Acc & H-score & \cellcolor{gray!25}OSCR & Acc & H-score & \cellcolor{gray!25}OSCR & Acc & H-score & \cellcolor{gray!25}OSCR \\

\midrule

NPN~\cite{sheng2024adaptive}  & 14.70 & 19.23 & \cellcolor{gray!25}13.33 & 12.57 & 15.38 & \cellcolor{gray!25}7.98 & 31.77 & 14.05 & \cellcolor{gray!25}18.91 & 19.14 & 10.18 & \cellcolor{gray!25}18.22 & 19.55 & 14.96 & \cellcolor{gray!25}14.61 \\

BadLabel~\cite{zhang2024badlabel}  & 39.82 & 26.45 & \cellcolor{gray!25}29.16 & 30.96 & 30.06& \cellcolor{gray!25}20.34 & 42.55 & 30.39& \cellcolor{gray!25}27.73 & 30.88 & 22.22 & \cellcolor{gray!25}15.28 & 36.05 & 27.21 & \cellcolor{gray!25}23.13\\

\midrule

ODGNet~\cite{bose2023beyond} & 45.72 & 35.60 & \cellcolor{gray!25}33.19 & 42.03 & 35.95& \cellcolor{gray!25}27.33 & 47.60 & 41.28& \cellcolor{gray!25}33.24 & 34.07 & 29.39 & \cellcolor{gray!25}18.68 & 42.36 & 35.56 & \cellcolor{gray!25}28.11\\

MEDIC-cls~\cite{wang2023generalizable} &45.32  & 32.01 & \cellcolor{gray!25}25.66 & 36.52 & 28.95& \cellcolor{gray!25}22.26 & 45.49 & 37.13& \cellcolor{gray!25}28.22 & 44.92 & 16.23 & \cellcolor{gray!25}18.73 &  43.06 & 28.58 & \cellcolor{gray!25}23.72 \\

MEDIC-bcls~\cite{wang2023generalizable}  & 45.32 & 5.95 & \cellcolor{gray!25}28.05 & 36.52 &20.04 & \cellcolor{gray!25}15.61 & 45.49 & 8.66& \cellcolor{gray!25}23.90 & 44.92 & 0.00 & \cellcolor{gray!25}23.50 &  43.06 & 8.66 & \cellcolor{gray!25}22.78\\
EBiL-HaDS-cls~\cite{peng2024advancing}   & 54.12 & 16.73 & \cellcolor{gray!25}45.65 & 50.59 & 19.31& \cellcolor{gray!25}39.42 &56.47  &13.08 & \cellcolor{gray!25}39.39 & 49.52 & 49.31 & \cellcolor{gray!25}40.52 & 52.68 &24.61 & \cellcolor{gray!25}41.25 \\
\bottomrule
\end{tabular}

}
\end{table*}

\begin{table*}[t!]
\caption{Results (\%) of PACS on ViT-Base~\cite{dosovitskiy2021an} using CLIP~\cite{radford2021learning} initialization.
The open-set ratio is $6{:}1$ and symmetric label noise with a ratio of $80\%$ is selected.}
\label{tab:pacs_vit_80_clip}
\centering
\resizebox{1.\linewidth}{!}{
\begin{tabular}{l|ccc|ccc|ccc|ccc|ccc}
\toprule
& \multicolumn{3}{c|}{\textbf{Photo (P)}} & \multicolumn{3}{c|}{\textbf{Art (A)}} & \multicolumn{3}{c|}{\textbf{Cartoon (C)}} & \multicolumn{3}{c|}{\textbf{Sketch (S)}} & \multicolumn{3}{c}{\textbf{Avg}} \\
\textbf{Method} & Acc & H-score & \cellcolor{gray!25}OSCR & Acc & H-score & \cellcolor{gray!25}OSCR & Acc & H-score & \cellcolor{gray!25}OSCR & Acc & H-score & \cellcolor{gray!25}OSCR & Acc & H-score & \cellcolor{gray!25}OSCR \\

\midrule

NPN~\cite{sheng2024adaptive}  & 15.27 & 15.50 & \cellcolor{gray!25}8.17 & 16.01 & 2.83 & \cellcolor{gray!25}7.02 & 23.62 & 15.39 & \cellcolor{gray!25}7.85 & 16.13 & 12.98 & \cellcolor{gray!25}13.17 &17.76 & 11.68& \cellcolor{gray!25}9.05 \\

BadLabel~\cite{zhang2024badlabel}  & 16.88 & \textbf{16.34} & \cellcolor{gray!25}8.02 & 18.45 & 9.58 & \cellcolor{gray!25}8.28 & \textbf{31.00} & 22.21 & \cellcolor{gray!25}14.98 & 23.59 & 0.00 & \cellcolor{gray!25}10.31 &22.48 & 12.03 & \cellcolor{gray!25} 10.40\\

\midrule 

ODGNet~\cite{bose2023beyond} & \textbf{27.22} & 10.83 & \cellcolor{gray!25}8.97 & 19.70 & 3.97 & \cellcolor{gray!25}7.62 & 25.37 & 24.79 & \cellcolor{gray!25}14.82 & 26.16 & 25.54 & \cellcolor{gray!25}16.48 &24.61 & 16.28 & \cellcolor{gray!25}11.97 \\

MEDIC-cls~\cite{wang2023generalizable} & 15.51 & 10.59 & \cellcolor{gray!25}5.25 & 9.76 & 7.97 & \cellcolor{gray!25}4.36 & 18.00 & 9.34 & \cellcolor{gray!25}11.73 & 20.48  & 0.0 & \cellcolor{gray!25}4.24 & 15.94 & 6.98 & \cellcolor{gray!25}6.40\\

MEDIC-bcls~\cite{wang2023generalizable}  & 15.51 & 8.89 & \cellcolor{gray!25}9.96 & 9.76 & 0.13 & \cellcolor{gray!25}5.19 & 18.00 & 16.43 & \cellcolor{gray!25}9.02 & 20.48 & 0.0 & \cellcolor{gray!25}12.83 &15.94&6.36 & \cellcolor{gray!25}9.25 \\
EBiL-HaDS-cls~\cite{peng2024advancing}  & 18.66 & 0.00 & \cellcolor{gray!25}12.62 & 13.76 &0.0 & \cellcolor{gray!25}5.16 & 20.42 & 21.23 & \cellcolor{gray!25}11.75 & 23.75 & 0.00 & \cellcolor{gray!25}12.62 & 19.15&5.31 & \cellcolor{gray!25}10.54 \\
\bottomrule
\end{tabular}

}
\end{table*}

\begin{table*}[t!]
\caption{Results (\%) of PACS on ViT-Base~\cite{dosovitskiy2021an} using CLIP~\cite{radford2021learning} initialization.
The open-set ratio is $6{:}1$ and asymmetric label noise with a ratio of $50\%$ is selected.}
\label{tab:pacs_vit_50_a_clip}
\centering
\resizebox{1.\linewidth}{!}{
\begin{tabular}{l|ccc|ccc|ccc|ccc|ccc}
\toprule
& \multicolumn{3}{c|}{\textbf{Photo (P)}} & \multicolumn{3}{c|}{\textbf{Art (A)}} & \multicolumn{3}{c|}{\textbf{Cartoon (C)}} & \multicolumn{3}{c|}{\textbf{Sketch (S)}} & \multicolumn{3}{c}{\textbf{Avg}} \\
\textbf{Method} & Acc & H-score & \cellcolor{gray!25}OSCR & Acc & H-score & \cellcolor{gray!25}OSCR & Acc & H-score & \cellcolor{gray!25}OSCR & Acc & H-score & \cellcolor{gray!25}OSCR & Acc & H-score & \cellcolor{gray!25}OSCR \\

\midrule

NPN~\cite{sheng2024adaptive}  & 35.06 & 17.05 & \cellcolor{gray!25}9.11 & 36.90 & 22.20 & \cellcolor{gray!25}12.96 & 20.06 & 12.98 & \cellcolor{gray!25}10.57 & 20.48 & 22.59 & \cellcolor{gray!25}19.35 & 28.13 & 18.71 & \cellcolor{gray!25}13.00 \\

BadLabel~\cite{zhang2024badlabel}  & 36.83 & 35.92 & \cellcolor{gray!25}26.46 & 30.64 & 23.89& \cellcolor{gray!25}13.96 & 32.65 & 25.72 & \cellcolor{gray!25}16.28 & 28.87 & 27.81 & \cellcolor{gray!25}18.89 &  32.25 & 28.34 & \cellcolor{gray!25}18.90\\

\midrule

ODGNet~\cite{bose2023beyond} & 42.00 & 36.96 & \cellcolor{gray!25}32.93 & 34.15 & 32.07& \cellcolor{gray!25}24.77 & 41.10 & 36.93 & \cellcolor{gray!25}26.13 & 34.07 & 29.39 & \cellcolor{gray!25}18.68 &  37.83& 33.84 & \cellcolor{gray!25}25.63\\

MEDIC-cls~\cite{wang2023generalizable} & 47.09 & 47.23 & \cellcolor{gray!25}41.29 & 25.45 & 24.26 & \cellcolor{gray!25}14.58 & 43.58 &37.96  & \cellcolor{gray!25}28.94 & 44.92 & 16.23 & \cellcolor{gray!25}18.73 & 40.26& 31.42 & \cellcolor{gray!25}25.89\\

MEDIC-bcls~\cite{wang2023generalizable}  & 47.09 & 0.00 & \cellcolor{gray!25}21.97 & 25.45 & 6.24 & \cellcolor{gray!25}9.96 & 43.58 & 5.03& \cellcolor{gray!25}28.45 & 44.92 &0.00  & \cellcolor{gray!25}23.50 &40.26& 2.82 & \cellcolor{gray!25}20.97 \\
\bottomrule
\end{tabular}

}
\end{table*}

\begin{table*}[t!]
\caption{Results ($\%$) of DomainNet using ResNet18~\cite{he2016deep}. The target domain is selected as \textit{Clipart}.}
\label{tab:clipart}
\centering
\resizebox{0.95\linewidth}{!}{
\begin{tabular}{l|ccc|ccc|ccc|ccc}
\toprule
& \multicolumn{3}{c|}{\textbf{20\% sym}} & \multicolumn{3}{c|}{\textbf{50\% sym}} & \multicolumn{3}{c|}{\textbf{80\% sym}} & \multicolumn{3}{c}{\textbf{50\% asym}} \\
\textbf{Method} & Acc & H-score & \cellcolor{gray!25}OSCR & Acc & H-score & \cellcolor{gray!25}OSCR & Acc & H-score & \cellcolor{gray!25}OSCR & Acc & H-score & \cellcolor{gray!25}OSCR \\
\midrule
NPN~\cite{sheng2024adaptive} & 53.04 & 51.47 & \cellcolor{gray!25}43.70 & 48.49 & 48.02 & \cellcolor{gray!25}39.24 & 34.62 & 38.14& \cellcolor{gray!25}27.15 & 48.11 & 47.26 & \cellcolor{gray!25}38.41 \\

BadLabel~\cite{zhang2024badlabel}  & 52.25 & 51.82 & \cellcolor{gray!25}44.07 & 50.78 & 49.79 & \cellcolor{gray!25}41.25 & 35.28 & 31.61& \cellcolor{gray!25}27.26 & 13.15 & 18.11 & \cellcolor{gray!25}10.22 \\

\midrule 

ODGNet~\cite{bose2023beyond}  & 53.25 & 53.13 & \cellcolor{gray!25}44.06 & 31.85 & 35.11 & \cellcolor{gray!25}24.59 & 20.36 & 21.84& \cellcolor{gray!25}15.19 & 30.46 & 33.74 & \cellcolor{gray!25}23.25 \\

MEDIC-cls~\cite{wang2023generalizable}  & 38.11 & 42.14 & \cellcolor{gray!25}31.26 & 11.49 & 8.61 & \cellcolor{gray!25}15.24 & 1.14 & 0.32 & \cellcolor{gray!25}0.56 & 13.23 & 17.98 & \cellcolor{gray!25}10.32 \\

MEDIC-bcls~\cite{wang2023generalizable}  & 38.11 & 14.36 & \cellcolor{gray!25}19.12 & 11.49 & 7.47 & \cellcolor{gray!25}5.73 & 1.14 &1.71 & \cellcolor{gray!25}0.53 & 13.23 & 9.89 & \cellcolor{gray!25}6.87 \\

EBiL-HaDS-cls~\cite{peng2024advancing}   & 47.91 & 49.07 & \cellcolor{gray!25}39.36 & 38.33 & 41.51 & \cellcolor{gray!25}30.70 & 9.43 &12.98 & \cellcolor{gray!25}6.63 & 49.43 & 49.52 & \cellcolor{gray!25}40.42 \\
\bottomrule
\end{tabular}}

\vskip-1ex
\end{table*}

\begin{table*}[t!]
\caption{Results ($\%$) of DomainNet using ResNet18~\cite{he2016deep}. The target domain is selected as \textit{Infograph}.}
\label{tab:infograph}
\centering
\resizebox{0.95\linewidth}{!}{
\begin{tabular}{l|ccc|ccc|ccc|ccc}
\toprule
& \multicolumn{3}{c|}{\textbf{20\% sym}} & \multicolumn{3}{c|}{\textbf{50\% sym}} & \multicolumn{3}{c|}{\textbf{80\% sym}} & \multicolumn{3}{c}{\textbf{50\% asym}} \\
\textbf{Method} & Acc & H-score & \cellcolor{gray!25}OSCR & Acc & H-score & \cellcolor{gray!25}OSCR & Acc & H-score & \cellcolor{gray!25}OSCR & Acc & H-score & \cellcolor{gray!25}OSCR \\
\midrule
NPN~\cite{sheng2024adaptive} & \textbf{17.48} & 17.06 & \cellcolor{gray!25}12.15 & \textbf{16.10} & 15.64 & \cellcolor{gray!25}10.81 & 12.44 & \textbf{14.87} & \cellcolor{gray!25}8.41 & 15.49 & 16.79 & \cellcolor{gray!25}9.94 \\

BadLabel~\cite{zhang2024badlabel}  & 15.88 & 15.58 & \cellcolor{gray!25}10.20 & 12.89 & 14.31 & \cellcolor{gray!25}7.69 & 12.67 & 12.70 & \cellcolor{gray!25}7.20 & 5.35 & 6.70 & \cellcolor{gray!25}3.85 \\

\midrule 

ODGNet~\cite{bose2023beyond}  & 17.37 & 20.24 & \cellcolor{gray!25}13.36 & 14.23 & 17.11 & \cellcolor{gray!25}10.71 & 10.49 & 11.56 & \cellcolor{gray!25}6.69 & 14.53 & 17.51 & \cellcolor{gray!25}11.23 \\

MEDIC-cls~\cite{wang2023generalizable}  & 13.38 & 15.50 & \cellcolor{gray!25}9.35 & 7.04 & 9.99 & \cellcolor{gray!25}4.46 & 1.08 & 0.46 & \cellcolor{gray!25}0.77 & 7.92 & 10.75 & \cellcolor{gray!25}5.80 \\

MEDIC-bcls~\cite{wang2023generalizable}  & 13.38 & 3.45 & \cellcolor{gray!25}9.35 & 7.04 & 3.65 & \cellcolor{gray!25}4.10 & 1.08 & 0.24 & \cellcolor{gray!25}0.27 & 7.92 & 4.19 & \cellcolor{gray!25}4.82 \\

EBiL-HaDS-cls~\cite{peng2024advancing}   & 16.51 & 18.16 & \cellcolor{gray!25}11.24 & 8.25 & 8.17 & \cellcolor{gray!25}5.35 & 7.06 & 9.59 & \cellcolor{gray!25}4.81 & 14.25 & 16.19 & \cellcolor{gray!25}9.43 \\
\bottomrule
\end{tabular}}

\vskip-1ex
\end{table*}

\begin{table*}[t!]

\caption{Results ($\%$) of DomainNet using ResNet18~\cite{he2016deep}. The target domain is selected as \textit{Quickdraw}.}
\label{tab:quickdraw}
\centering
\resizebox{0.95\linewidth}{!}{
\begin{tabular}{l|ccc|ccc|ccc|ccc}
\toprule
& \multicolumn{3}{c|}{\textbf{20\% sym}} & \multicolumn{3}{c|}{\textbf{50\% sym}} & \multicolumn{3}{c|}{\textbf{80\% sym}} & \multicolumn{3}{c}{\textbf{50\% asym}} \\
\textbf{Method} & Acc & H-score & \cellcolor{gray!25}OSCR & Acc & H-score & \cellcolor{gray!25}OSCR & Acc & H-score & \cellcolor{gray!25}OSCR & Acc & H-score & \cellcolor{gray!25}OSCR \\
\midrule
NPN~\cite{sheng2024adaptive} & 12.50 & 8.31 & \cellcolor{gray!25}9.00 & 12.55 & 6.38 & \cellcolor{gray!25}8.66 & 5.13 &3.77 & \cellcolor{gray!25}3.45 & 11.59 & 7.14 & \cellcolor{gray!25}8.05 \\

BadLabel~\cite{zhang2024badlabel}  & 11.35 & 5.98 & \cellcolor{gray!25}7.77 & 11.25 & 5.71 & \cellcolor{gray!25}7.91 & 5.61 & 3.11 & \cellcolor{gray!25}3.74 & 3.51 & 2.94 & \cellcolor{gray!25}1.92 \\

\midrule 

ODGNet~\cite{bose2023beyond}  & 13.96 & 10.18 & \cellcolor{gray!25}9.52 & 11.94 & 8.27 & \cellcolor{gray!25}8.00 & 6.16 & 4.90 & \cellcolor{gray!25}4.24 & 9.21 & 6.64 & \cellcolor{gray!25}6.03 \\

MEDIC-cls~\cite{wang2023generalizable}  & 3.20 & 3.23 & \cellcolor{gray!25}2.31 & 4.45 & 3.84 & \cellcolor{gray!25}3.28 & 0.55 & 0.57 & \cellcolor{gray!25}0.33 & 4.55 & 2.23 & \cellcolor{gray!25}3.31 \\

MEDIC-bcls~\cite{wang2023generalizable}  & 3.20 & 2.31 & \cellcolor{gray!25}1.55 & 4.45 & 3.00 & \cellcolor{gray!25}2.21 & 0.55 & 0.89 & \cellcolor{gray!25}0.34 & 4.55 & 4.32 & \cellcolor{gray!25}2.14 \\

EBiL-HaDS-cls~\cite{peng2024advancing}   & 12.48 & 9.47 & \cellcolor{gray!25}8.83 & 6.46 & 6.27 & \cellcolor{gray!25}4.61 & 3.35 & 5.19 & \cellcolor{gray!25}2.36 & 6.46 & 6.29 & \cellcolor{gray!25}4.60 \\
\bottomrule
\end{tabular}}

\vskip-1ex
\end{table*}

ODGNet~\cite{bose2023beyond} achieves closed-set accuracy of $68.88\%$, $60.90\%$, and $17.17\%$; H-score of $40.87\%$, $30.62\%$, and $11.31\%$; and OSCR of $51.07\%$, $43.12\%$, and $9.52\%$ on the DigitsDG dataset, as shown in Table~\ref{tab:dg_20}, Table~\ref{tab:dg_50}, and Table~\ref{tab:dg_80}, respectively. Under asymmetric label noise with a ratio of $50\%$, NPN~\cite{sheng2024adaptive} achieves $58.39\%$, $29.98\%$, and $42.38\%$ in terms of closed-set accuracy, H-score, and OSCR, respectively, as presented in Table~\ref{tab:dg_50a}.

Our proposed approach achieves state-of-the-art performance on the DigitsDG dataset. Specifically, it outperforms ODGNet~\cite{bose2023beyond} by $4.27\%$, $0.98\%$, and $1.11\%$ in terms of OSCR under symmetric label noise with ratios of $20\%$, $50\%$, and $80\%$, respectively, and surpasses NPN~\cite{sheng2024adaptive}  by $6.41\%$ in terms of OSCR under asymmetric label noise with a ratio of $50\%$. 
HyProMeta also consistently outperforms both EBiL-HaDS-cls~\cite{peng2024advancing} and EBiL-HaDS-bcls~\cite{peng2024advancing} across all noise levels on the DigitsDG dataset, achieving the highest average OSCR in every setting. The performance gap is particularly pronounced under high noise conditions, with HyProMeta reaching $10.63\%$ OSCR at $80\%$ symmetric noise, compared to just $7.45\%$ and $7.31\%$ for the EBiL-HaDS variants.

These results further confirm the strong cross-dataset generalizability of our proposed approach in addressing the challenging OSDG-NL task. 

\noindent\textbf{Analysis on benchmark on DomainNet:} 
In order to further explore the OSDG-NL setting on more large-scale domain generalization benchmark with more classes, we delivered the experiments in Table~\ref{tab:clipart}, Table~\ref{tab:infograph}, and Table~\ref{tab:quickdraw}, where we choose \textit{Clipart}, \textit{Infograph}, and \textit{Quickdraw} as the target domain in leave one-out setting, respectively.

Among those test domains, \textit{Quickdraw} and \textit{Infograph} represent for challenging domain generalization settings while \textit{Clipart} preserves less domain generalization difficulty. We adopt these evaluation setting to deliver the performances of the baselines and our proposed approach under different domain generalization difficulties.

The results presented in Tables~\ref{tab:clipart},~\ref{tab:infograph}, and~\ref{tab:quickdraw} provide a comprehensive evaluation of HyProMeta against a variety of baseline methods on the DomainNet dataset under the OSDG-NL setting. Specifically, Table~\ref{tab:clipart} demonstrates performance when the target domain is \textit{Clipart}, while Table~\ref{tab:infograph} evaluates the case when \textit{Infograph} is selected as the target domain. Table~\ref{tab:quickdraw} delivers the performance evaluation of OSDG-NL when \textit{Quickdraw} is selected as the target domain. All of the above-mentioned tables include experiments under different types and levels of label noise, including symmetric noise at $20\%$, $50\%$, and $80\%$, as well as asymmetric noise at a $50\%$ ratio.

From Table~\ref{tab:clipart}, it can be observed that the proposed HyProMeta method consistently outperforms all baseline approaches across nearly all noise configurations when generalizing to the \textit{Clipart} domain, where the rest of the domains are used as source domains for training. For instance, under $20\%$ symmetric label noise, HyProMeta achieves $55.40\%$ accuracy, $53.39\%$ H-score, and $45.54\%$ OSCR, clearly surpassing the closest competitors from both the noisy label learning field (\textit{e.g.}, NPN~\cite{sheng2024adaptive} and BadLabel~\cite{zhang2024badlabel}) and the open-set domain generalization field (\textit{e.g.}, ODGNet~\cite{bose2023beyond} and EBiL-HaDS~\cite{peng2024advancing}). Even under more severe noise conditions such as $80\%$ symmetric noise, HyProMeta still maintains competitive performance, reaching $31.78\%$ OSCR and $42.12\%$ H-score, demonstrating its robustness against high noise corruption and its effectiveness in retaining generalization to unseen classes.
Table~\ref{tab:infograph} presents a more challenging scenario, with \textit{Infograph} as the target domain, which introduces a more substantial visual domain shift. As expected, all methods exhibit a performance drop across all metrics. Nevertheless, HyProMeta continues to demonstrate superior robustness and generalizability. Under $50\%$ asymmetric noise, it achieves the best performance with $15.73\%$ Acc, $19.15\%$ H-score, and $11.61\%$ OSCR. 
Notably, the performance gap brought by a more challenging test domain highlights the difficulty of generalization under compounded domain and label noise conditions.

As shown in Table~\ref{tab:quickdraw}, HyProMeta outperforms all competing methods across nearly all noise levels. Under the $20\%$ symmetric label noise setting, HyProMeta achieves $14.87\%$ Acc and $10.77\%$ OSCR, which are the highest values among all methods. Even under more severe noise, such as the $50\%$ symmetric case, HyProMeta maintains its lead with $13.44\%$ Acc and $9.32\%$ OSCR, indicating strong resilience against noisy supervision. Importantly, in the extremely noisy $80\%$ symmetric scenario, where many baseline methods exhibit performance collapse, HyProMeta still achieves $7.86\%$ accuracy and $5.41\%$ OSCR, significantly outperforming the rest.
In the $50\%$ asymmetric label noise setting, HyProMeta again delivers the best overall performance with $12.27\%$ Acc, $7.65\%$ H-score, and $8.76\%$ OSCR. These results further confirm the robustness and adaptability of the proposed framework in handling both high semantic label confusion and significant visual domain shifts.
Comparatively, most baseline methods struggle with the dual challenges of noisy labels and open-set recognition in the \textit{Quickdraw} domain. Meta-learning-based approaches such as MEDIC-cls~\cite{wang2023generalizable} and MEDIC-bcls~\cite{wang2023generalizable} suffer from severe degradation across all metrics, often falling below $5\%$ accuracy. While some noisy label learning methods like NPN~\cite{sheng2024adaptive} and EBiL-HaDS~\cite{peng2024advancing} show moderate performance under low-noise settings, they deteriorate rapidly as noise increases or becomes asymmetric, indicating limited generalization capability under OSDG-NL settings.

The consistently strong performance of HyProMeta across both target domains and all noise settings supports the conclusion that its design, particularly the hyperbolic prototype-based label correction and prompt-based augmentation, effectively addresses the dual challenges of label noise and open-set domain shifts. In contrast, baseline methods from either the noisy label learning or domain generalization domains tend to fail under extreme conditions, confirming the necessity of a unified framework tailored for the OSDG-NL task.

A more detailed ablation study of each module component, along with an analysis on the learned embedding space and confidence scores, will be presented in the subsequent sections.
\begin{figure*}[!h]
\centering
\begin{subfigure}[b]{0.23\textwidth}
    \centering
    \includegraphics[width=\textwidth]{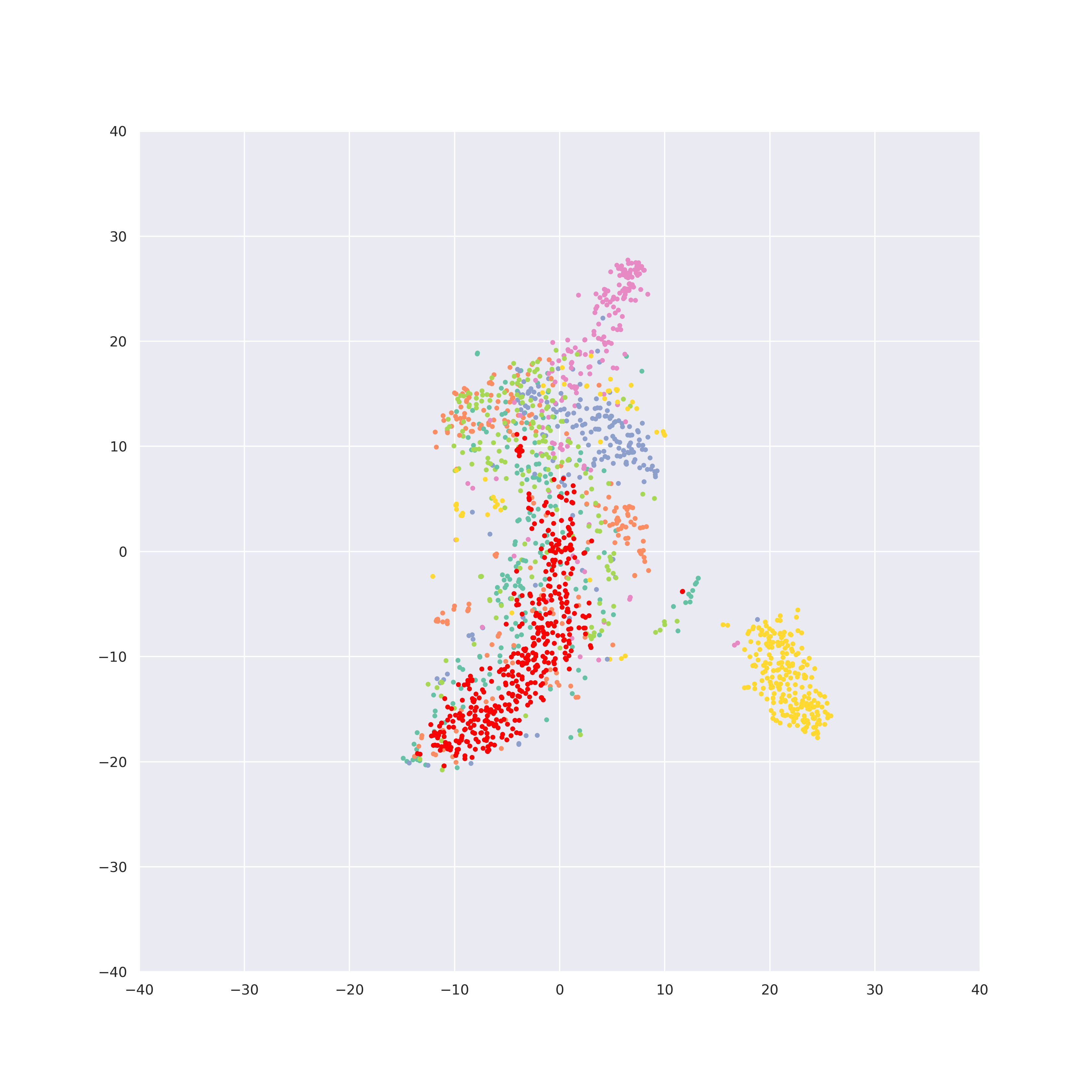}
    \caption{BadLabel~\cite{zhang2024badlabel}, D:Photo, LNSR:20\% sym}
    \label{fig:bad_label_photo_tsne}
\end{subfigure}
\hfill
\begin{subfigure}[b]{0.23\textwidth}
    \centering
    \includegraphics[width=\textwidth]{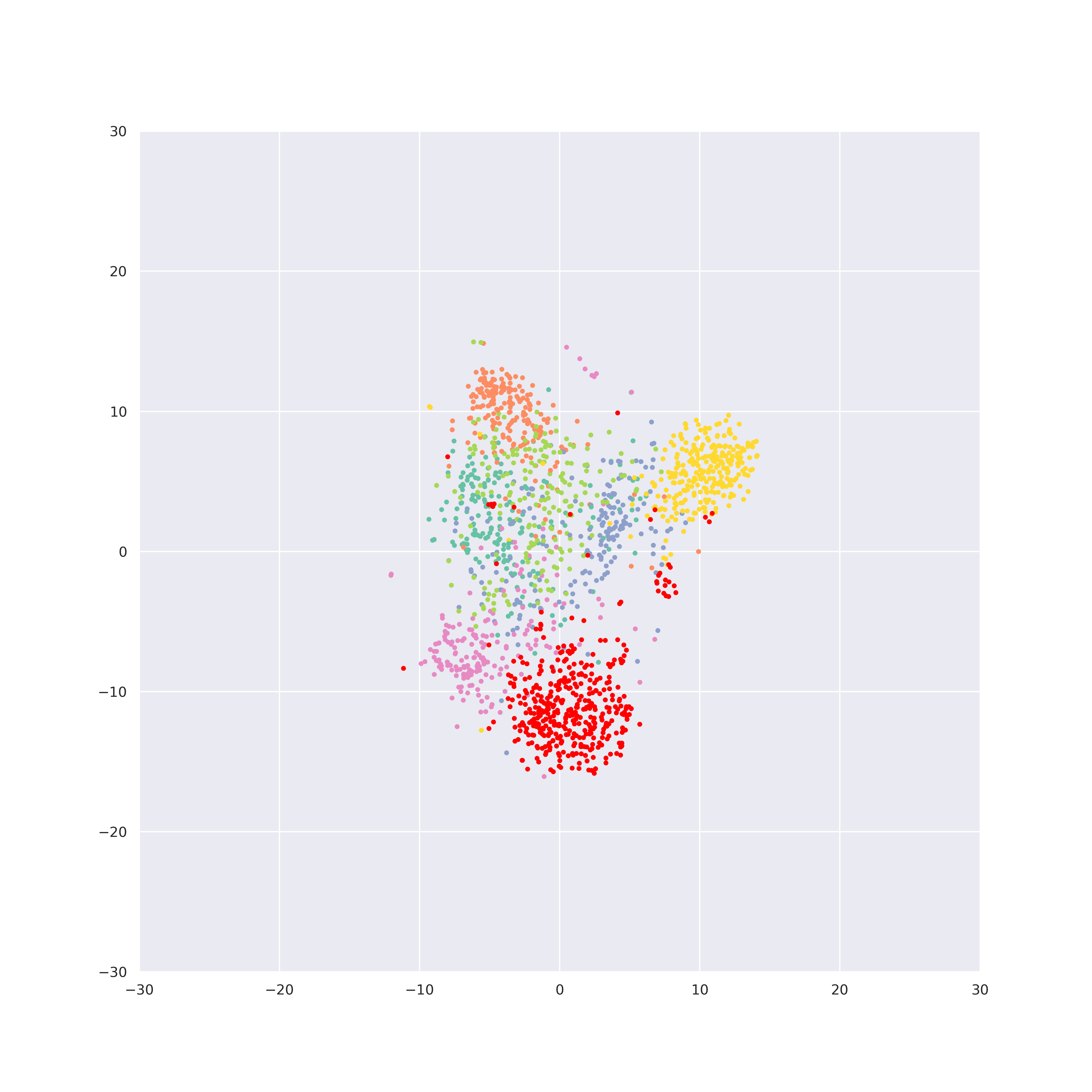}
    \caption{MLDG~\cite{shu2019meta}, D:Photo, LNSR:20\% sym}
    \label{fig:MLDG_photo_tsne}
\end{subfigure}
\hfill
\begin{subfigure}[b]{0.23\textwidth}
    \centering
    \includegraphics[width=\textwidth]{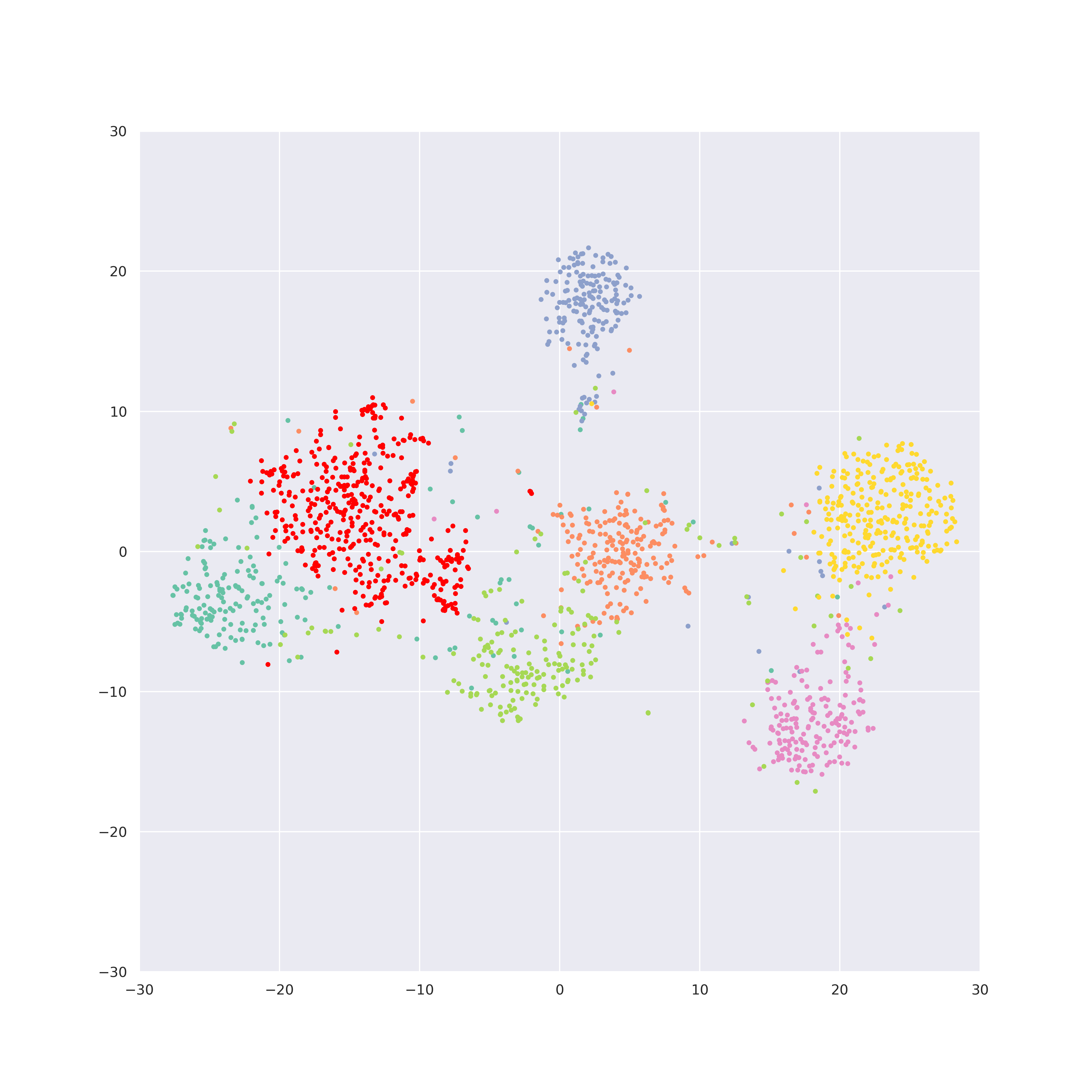}
    \caption{MEDIC~\cite{wang2023generalizable}, D:Photo, LNSR:20\% sym}
    \label{fig:medic_photo_tsne}
\end{subfigure}
\hfill
\begin{subfigure}[b]{0.23\textwidth}
    \centering
    \includegraphics[width=\textwidth]{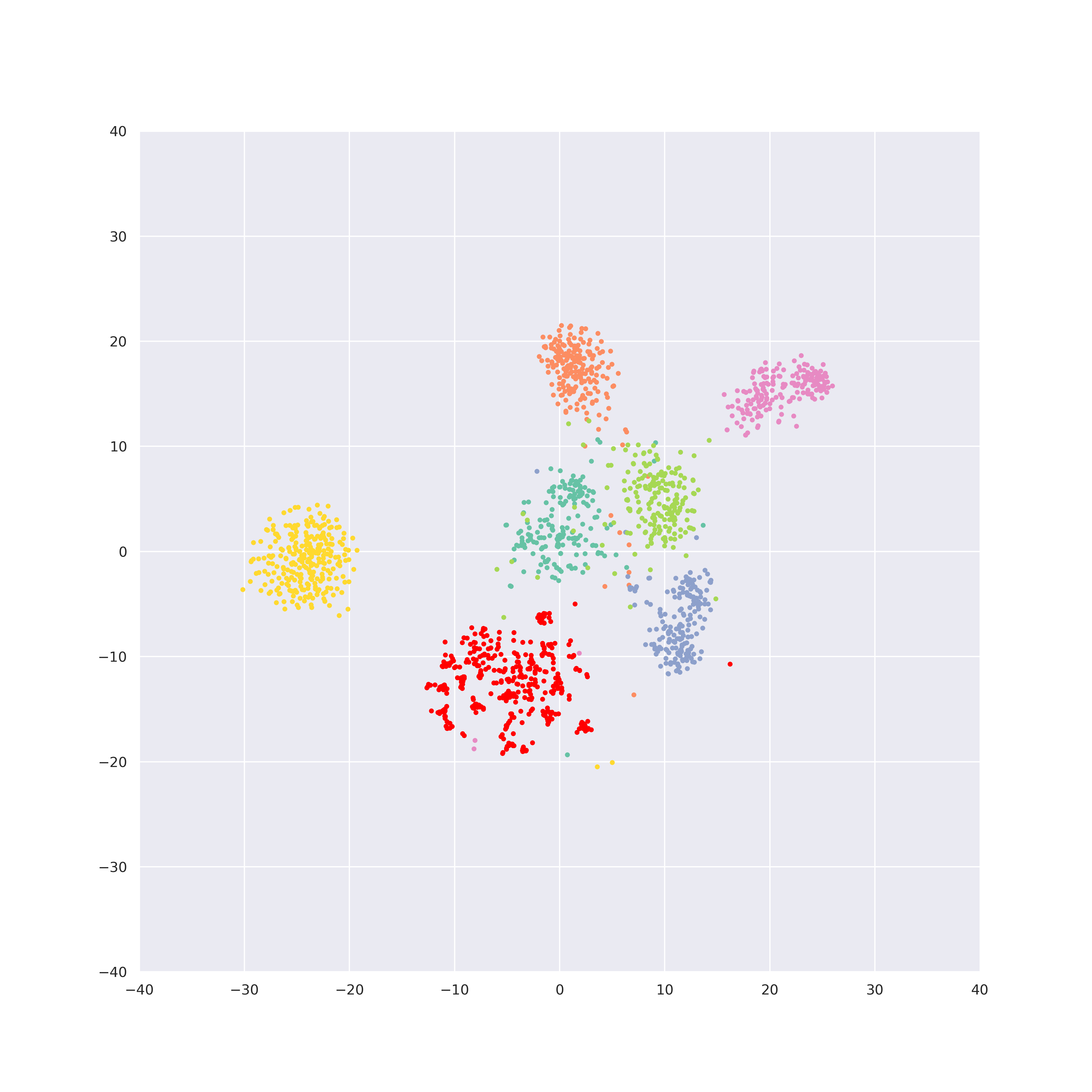}
    \caption{HyProMeta (Ours), D:Photo, LNSR:20\% sym}
    \label{fig:ours_photo_tsne}
\end{subfigure}
\hfill
\begin{subfigure}[b]{0.23\textwidth}
    \centering
    \includegraphics[width=\textwidth]{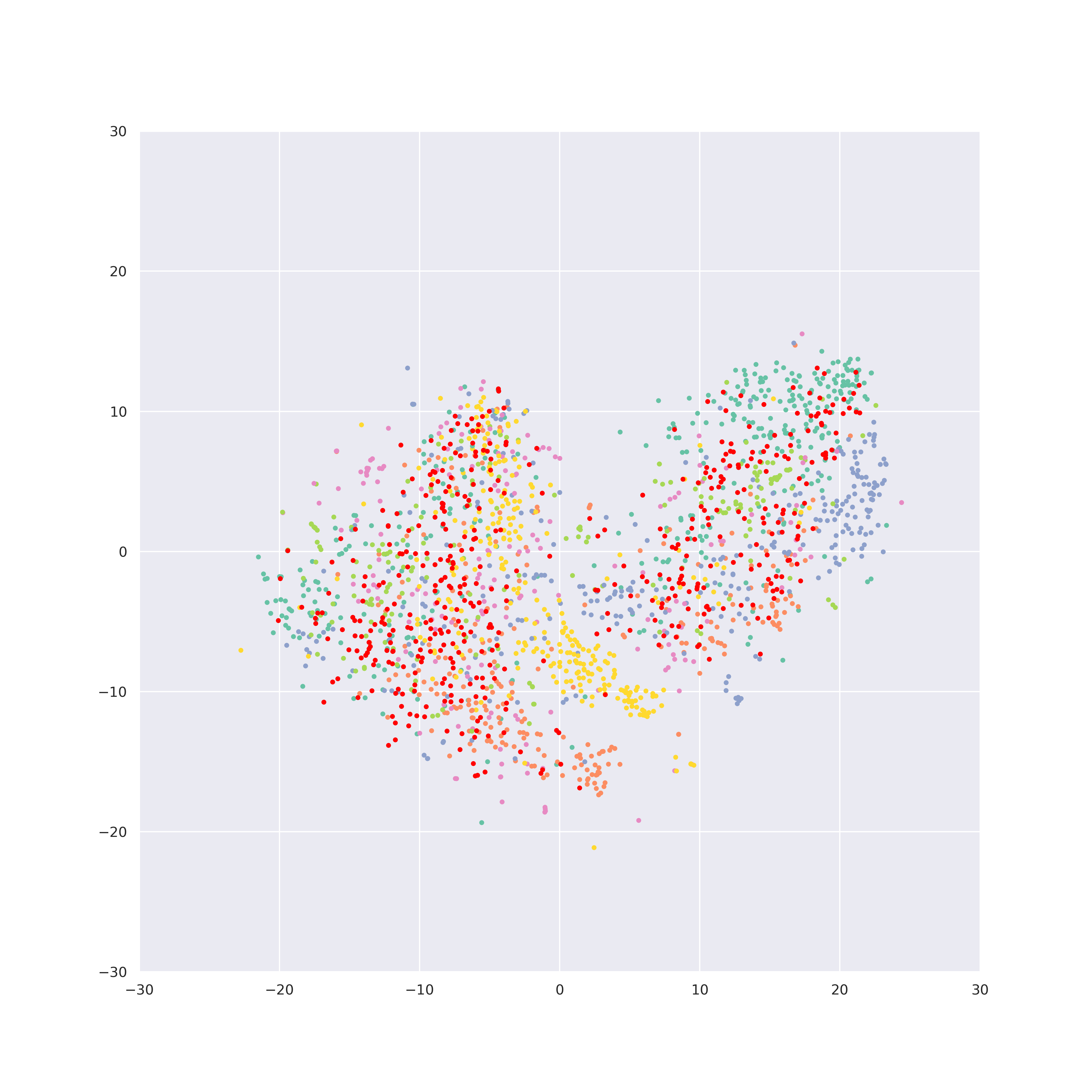}
    \caption{BadLabel~\cite{zhang2024badlabel}, D:Art, LNSR:20\% sym}
    \label{fig:bad_label_art}
\end{subfigure}
\hfill
\begin{subfigure}[b]{0.23\textwidth}
    \centering
    \includegraphics[width=\textwidth]{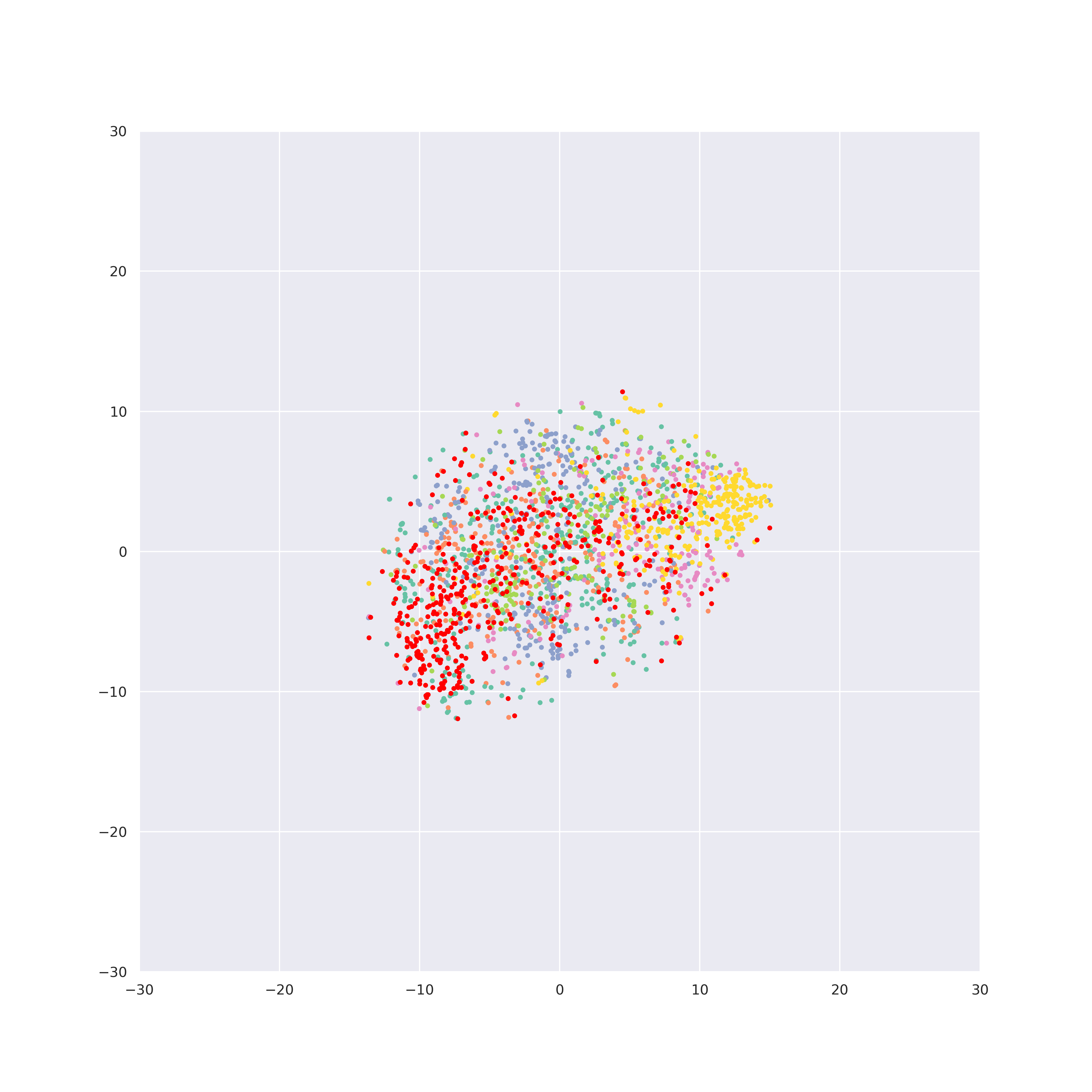}
    \caption{MLDG~\cite{shu2019meta}, D:Art, LNSR:20\% sym}
    \label{fig:mldg_art}
\end{subfigure}
\hfill
\begin{subfigure}[b]{0.23\textwidth}
    \centering
    \includegraphics[width=\textwidth]{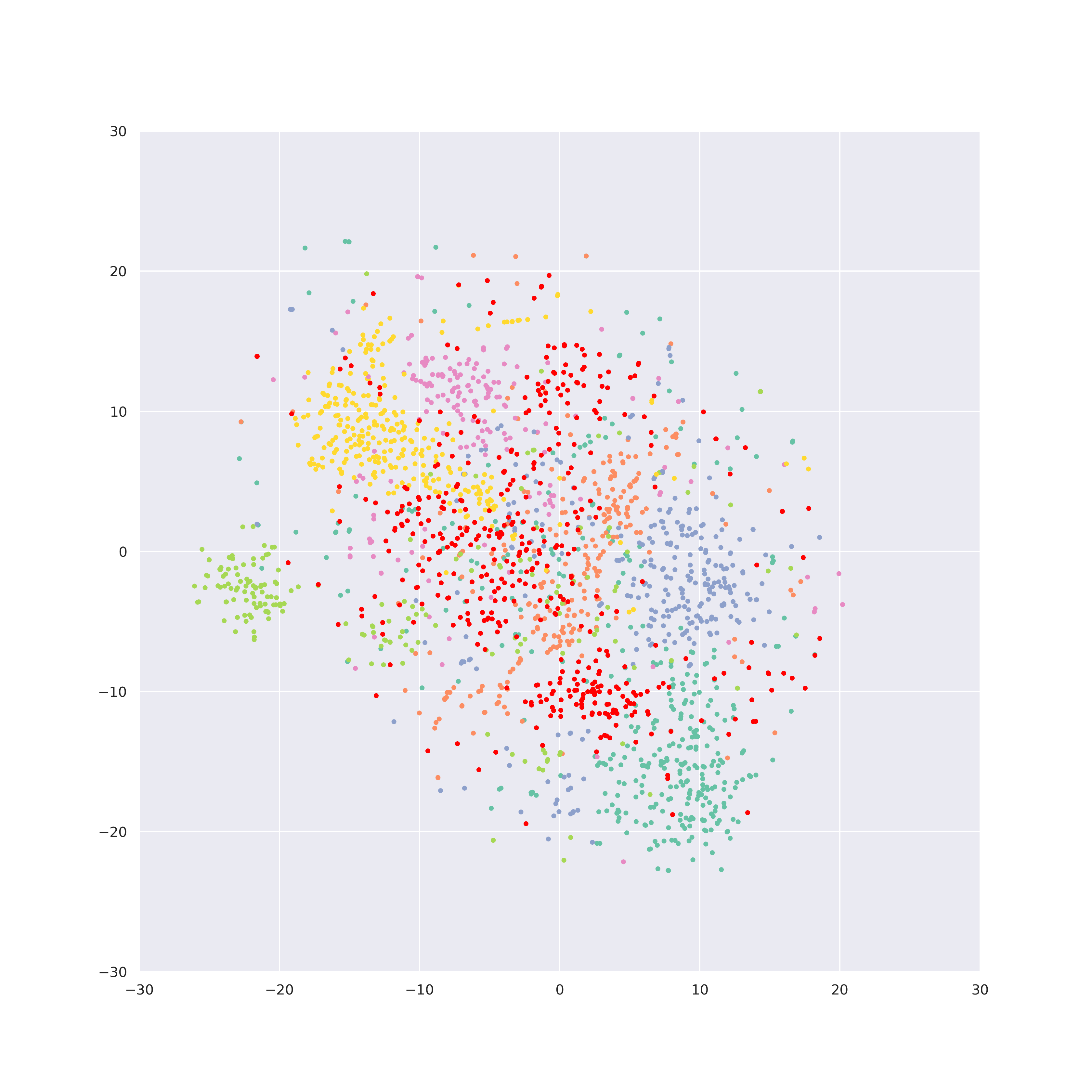}
    \caption{MEDIC~\cite{wang2023generalizable}, D:Art, LNSR:20\% sym}
    \label{fig:medic_art}
\end{subfigure}
\hfill
\begin{subfigure}[b]{0.23\textwidth}
    \centering
    \includegraphics[width=\textwidth]{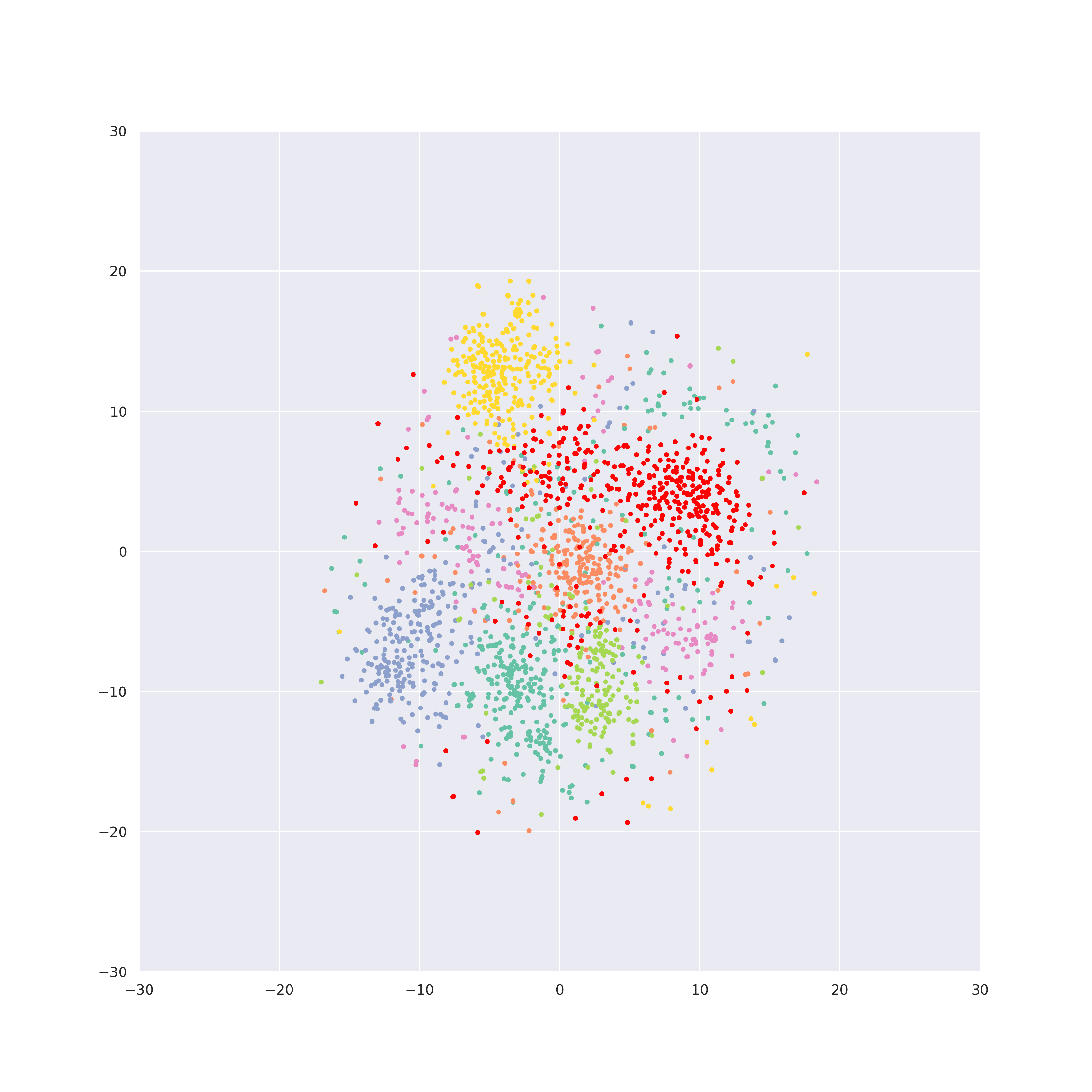}
    \caption{HyProMeta (Ours), D:Art, LNSR:20\% sym}
    \label{fig:ours_art}
\end{subfigure}
\caption{T-SNE visualization~\cite{van2008visualizing} of learned representations on PACS using ResNet18~\cite{he2016deep} under symmetric label noise with a ratio of $20\%$ when we use \textit{Photo} and \textit{Art Painting} as test domains, respectively. D indicates the target domain and LNSR indicates the label noise setting.}
\label{fig:tsne}
\end{figure*}

\begin{table*}
\caption{Evaluation of the noisy/clean set partition on PACS using ResNet18~\cite{he2016deep}, where the target domain is chosen as \textit{Photo}.}
\label{tab:clean_noisy}
\centering
\resizebox{0.95\linewidth}{!}{\begin{tabular}{l|lll|lll|lll|lll} 
\toprule
\textbf{LNSR}  & \multicolumn{3}{c|}{\textbf{20\% sym}} & \multicolumn{3}{c|}{\textbf{50\% sym}} & \multicolumn{3}{c|}{\textbf{80\% sym}} & \multicolumn{3}{c}{\textbf{50\% asym}} \\
\hline
\multicolumn{1}{c|}{\textbf{Domains}} & \multicolumn{1}{c}{\textbf{A}} & \multicolumn{1}{c}{\textbf{C}} & \multicolumn{1}{c|}{\textbf{S}} & \multicolumn{1}{c}{\textbf{A}} & \multicolumn{1}{c}{\textbf{C}} & \multicolumn{1}{c|}{\textbf{S}} & \multicolumn{1}{c}{\textbf{A}} & \multicolumn{1}{c}{\textbf{C}} & \multicolumn{1}{c|}{\textbf{S}} & \multicolumn{1}{c}{\textbf{A}} & \multicolumn{1}{c}{\textbf{C}} & \multicolumn{1}{c}{\textbf{S}}  \\ 
\hline
Mean                                  & 52.96                            & 51.19                                & 59.82                                & 54.28                            & 51.92                                & 56.17                                & 45.26                            & 32.98                                & 39.24                                & 48.17                            & \textbf{70.02}                                & 62.38                                \\
Median                                 & 53.98                            & 55.11                                & 53.41                                & 52.27                            & 56.81                                & 52.84                                & 46.02                            & \textbf{47.16}                                & 47.73                                & 47.73                            & 56.81                                & 51.70                                \\ 
\hline
Ours                                  & \textbf{82.18}                   & \textbf{83.56}                       & \textbf{81.32}                       & \textbf{79.41}                   & \textbf{77.00}                       & \textbf{69.47}                       & \textbf{50.59}                   & 45.07                       & \textbf{49.56}                       & \textbf{69.09}                   & 66.30                       & \textbf{69.66}                       \\
\bottomrule
\end{tabular}}
\end{table*}

\begin{table*}[t!]
\caption{Comparison of asymmetric label noise generated using different textual encoders (\textit{i.e.}, BERT~\cite{BERT} and CLIP~\cite{radford2021learning}) (\%) of PACS on ViT-Base~\cite{dosovitskiy2021an}. The open-set ratio is $6{:}1$ and asymmetric label noise with a ratio of $50\%$ is selected.}

\label{tab:clip_text}
\centering
\resizebox{1.\linewidth}{!}{
\begin{tabular}{l|c|ccc|ccc|ccc|ccc|ccc}
\toprule
& &\multicolumn{3}{c|}{\textbf{Photo (P)}} & \multicolumn{3}{c|}{\textbf{Art (A)}} & \multicolumn{3}{c|}{\textbf{Cartoon (C)}} & \multicolumn{3}{c|}{\textbf{Sketch (S)}} & \multicolumn{3}{c}{\textbf{Avg}} \\
\textbf{Method} &TextEnc. & Acc & H-score & \cellcolor{gray!25}OSCR & Acc & H-score & \cellcolor{gray!25}OSCR & Acc & H-score & \cellcolor{gray!25}OSCR & Acc & H-score & \cellcolor{gray!25}OSCR & Acc & H-score & \cellcolor{gray!25}OSCR \\

\midrule

NPN~\cite{sheng2024adaptive} &\multirow{8}{*}{\begin{sideways}
    BERT~\cite{BERT}
\end{sideways}}  & 27.63 & 8.86 & \cellcolor{gray!25}6.32 & 32.40 & 19.00 & \cellcolor{gray!25}15.72 & 21.66 & 16.87 & \cellcolor{gray!25}14.25 & 20.48 & 26.15 & \cellcolor{gray!25}16.50 & 25.54 & 17.72 & \cellcolor{gray!25}13.20 \\

BadLabel~\cite{zhang2024badlabel}  & & 46.12 & 58.96 & \cellcolor{gray!25}45.77 & 35.33 & 34.30 & \cellcolor{gray!25}34.83 & 26.92 & 39.52 & \cellcolor{gray!25}26.18 & 20.70 & \textbf{31.32} & \cellcolor{gray!25}20.66 & 32.27 & 41.03 & \cellcolor{gray!25}31.86 \\

\cmidrule{1-1}\cmidrule{3-17}

ODGNet~\cite{bose2023beyond}& & 51.13 & 55.83 & \cellcolor{gray!25}48.95 & 49.16 & 45.58 & \cellcolor{gray!25}37.21 & 40.69 & 37.35 & \cellcolor{gray!25}28.85 & 21.76 & 17.55 & \cellcolor{gray!25}12.68 & 40.69 & 39.08 & \cellcolor{gray!25}31.92 \\

MEDIC-cls~\cite{wang2023generalizable}& & 52.10 & 59.31 & \cellcolor{gray!25}48.42 & 46.47 & \textbf{56.57} & \cellcolor{gray!25}44.42 & 35.69 & 29.04 & \cellcolor{gray!25}21.31 & 35.42 & 28.01 & \cellcolor{gray!25}27.79 & 42.42 & 43.23 & \cellcolor{gray!25}35.49 \\

MEDIC-bcls~\cite{wang2023generalizable}& & 52.10 & 49.72 & \cellcolor{gray!25}42.70 & 46.47 & 55.52 & \cellcolor{gray!25}43.88 & 35.69 & 30.26 & \cellcolor{gray!25}20.16 & 35.42 & 24.54 & \cellcolor{gray!25}26.48 & 42.42 & 40.01 & \cellcolor{gray!25}33.31 \\

EBiL-HaDS-cls~\cite{peng2024advancing} & & 54.60 & 52.92 & \cellcolor{gray!25}44.98 & 46.97 & 56.52 & \cellcolor{gray!25}44.29 & 34.66 & 25.52 & \cellcolor{gray!25}22.48 & 25.71 & 26.60 & \cellcolor{gray!25}17.09 & 40.49 & 40.39 & \cellcolor{gray!25}32.21 \\
EBiL-HaDS-bcls~\cite{peng2024advancing} & & 54.60 & 39.01 & \cellcolor{gray!25}29.61 & 46.97 & 54.34 & \cellcolor{gray!25}42.74 & 34.66 & 31.86 & \cellcolor{gray!25}22.47 & 25.71 & 29.67 & \cellcolor{gray!25}16.33 & 40.49 & 38.72 & \cellcolor{gray!25} 27.79\\
\cmidrule{1-1}\cmidrule{3-17}
Ours&  & \textbf{56.87} & \textbf{59.59} & \cellcolor{gray!25}\textbf{53.15} & \textbf{55.97} & 56.31 & \cellcolor{gray!25}\textbf{48.31} & \textbf{48.94} & \textbf{46.85} & \cellcolor{gray!25}\textbf{40.16} & \textbf{38.18} & 31.14 & \cellcolor{gray!25}\textbf{32.15} & \textbf{49.99} & \textbf{48.47} & \cellcolor{gray!25}\textbf{43.44} \\
\midrule

NPN~\cite{sheng2024adaptive} &\multirow{8}{*}{\begin{sideways}
    CLIP~\cite{radford2021learning}
\end{sideways}} & 41.92 & 38.02 & \cellcolor{gray!25}33.40 & 43.40 & 31.50 & \cellcolor{gray!25}36.84 & 27.64 & 24.41 & \cellcolor{gray!25}19.08 & 30.69 & 22.96 & \cellcolor{gray!25}17.28 & 35.91 & 29.22 & \cellcolor{gray!25}26.65\\

BadLabel~\cite{zhang2024badlabel}  &  & 38.77 & 33.07 & \cellcolor{gray!25}29.46 & 32.02 & 34.20 & \cellcolor{gray!25}22.44 & 30.69 & 32.39 & \cellcolor{gray!25}21.17 & 21.41 & 29.44 & \cellcolor{gray!25}19.56 & 30.72 & 32.28 & \cellcolor{gray!25}23.16\\

\cmidrule{1-1}\cmidrule{3-17}

ODGNet~\cite{bose2023beyond}&  & 47.33 & 22.31 & \cellcolor{gray!25}26.80 & 51.09 & 52.26 & \cellcolor{gray!25}43.79 & 43.89 & 46.46 & \cellcolor{gray!25}37.16 & 37.60 & 5.71 & \cellcolor{gray!25}20.16 & 44.98 & 31.69 & \cellcolor{gray!25}31.98\\

MEDIC-cls~\cite{wang2023generalizable}&  & 57.51 & 42.73 & \cellcolor{gray!25}47.71 & 45.97 & 37.90 & \cellcolor{gray!25}35.80 & 39.92 & 38.74 & \cellcolor{gray!25}29.86 & 19.02 & 15.70 & \cellcolor{gray!25}7.98 & 40.61 & 33.77 & \cellcolor{gray!25}30.34\\

MEDIC-bcls~\cite{wang2023generalizable}&  & 57.51 & 25.00 & \cellcolor{gray!25}30.04 & 45.97 & 11.50 & \cellcolor{gray!25}24.04 & 39.92 & 21.22 & \cellcolor{gray!25}16.57 & 19.02 & 6.27 & \cellcolor{gray!25}6.93 & 40.61 & 16.00 & \cellcolor{gray!25}19.40\\

EBiL-HaDS-cls~\cite{peng2024advancing}&  & 57.19 & 42.09 & \cellcolor{gray!25}44.84 & 52.41 & 44.08 & \cellcolor{gray!25}39.49 & 41.21 & 46.20 & \cellcolor{gray!25}35.91 & 23.43 & 22.64 & \cellcolor{gray!25}16.27 & 43.56 & 38.75 & \cellcolor{gray!25}34.13\\
EBiL-HaDS-bcls~\cite{peng2024advancing}&  & 57.19 & 42.30 & \cellcolor{gray!25}40.86 & 52.41 & 44.87 & \cellcolor{gray!25}38.64 & 41.21 & 46.52 & \cellcolor{gray!25}35.74 & 23.43 & \textbf{28.45} & \cellcolor{gray!25}14.45 & 43.56 & 40.54  & \cellcolor{gray!25}32.42\\
\cmidrule{1-1}\cmidrule{3-17}
Ours&  & \textbf{62.52} & \textbf{52.18} & \cellcolor{gray!25}\textbf{52.41} & \textbf{56.35} & \textbf{54.03} & \cellcolor{gray!25}\textbf{47.38} & \textbf{51.62} & \textbf{56.13} & \cellcolor{gray!25}\textbf{46.64} & \textbf{42.21} & 27.90 & \cellcolor{gray!25}\textbf{31.53} & \textbf{53.18} & \textbf{47.56} & \cellcolor{gray!25}\textbf{44.49}\\
\bottomrule
\end{tabular}

}
\end{table*}

\subsection{Analysis on the Effect of Different Pretraining}
\label{sec:pretraining}
In Table~\ref{tab:pretraining_comparison}, we present the impact of three distinct pretraining strategies (\textit{i.e.}, ImageNet pretraining~\cite{deng2009imagenet}, CLIP pretraining~\cite{radford2021learning}, and random initialization). It can be clearly observed that all methods with random initialization consistently underperform compared to their counterparts pretrained on ImageNet~\cite{deng2009imagenet} or CLIP~\cite{radford2021learning} across all considered label noise settings. This highlights the critical role of pretraining for both the proposed model and related baselines in addressing the OSDG-NL task. Pretraining provides a strong initialization by capturing transferable and noise-resilient representations, which is particularly crucial in open-set domain generalization under noisy labels, where clean supervision is limited and domain shifts are significant.

Next, we compare the performance of models initialized with CLIP~\cite{radford2021learning} and ImageNet~\cite{deng2009imagenet} pretraining. HyProMeta achieves superior OSDG-NL results with CLIP~\cite{radford2021learning} initialization under symmetric label noise with a ratio of $20\%$, attaining $67.95\%$, $69.74\%$, and $62.33\%$ in terms of Acc, H-score, and OSCR, respectively. Similarly, under asymmetric label noise with a ratio of $50\%$, CLIP~\cite{radford2021learning} initialization yields $55.39\%$, $52.72\%$, and $47.19\%$ for the same metrics. In contrast, ImageNet pretraining offers more robust performance under higher noise levels, specifically symmetric label noise at $80\%$. This phenomenon is attributed to the nature of the pretraining data: CLIP~\cite{radford2021learning}, pretrained on noisy image-text pairs, provides strong semantic alignment that benefits moderate-noise scenarios, while ImageNet~\cite{deng2009imagenet} pretraining, based on manually curated labels, offers more stable and noise-resilient features under extreme label corruption.

Per-domain experimental results are presented in Table~\ref{tab:pacs_vit_20_clip}, Table~\ref{tab:pacs_vit_50_clip}, Table~\ref{tab:pacs_vit_80_clip}, and Table~\ref{tab:pacs_vit_50_a_clip}, corresponding to different label noise settings considered in this work. It can be observed that for more challenging test domains, \textit{e.g.}, \textit{Sketch}, utilizing CLIP-pretrained weights offers a more effective initialization compared to ImageNet-pretrained weights. For instance, HyProMeta achieves over $20\%$ OSCR improvement on PACS under $20\%$ symmetric label noise when initialized with CLIP. Similar trends are consistently observed across other label noise ratios. This performance gain can be attributed to the fact that CLIP is pretrained on a large-scale, diverse set of image-text pairs, which enables it to capture richer semantic representations. Such representations are especially beneficial in open-set domain generalization with noisy labels, where robust semantic alignment helps mitigate the domain shift and suppress noise-induced confusion.

\subsection{Analysis towards Clean/Noisy Label Partition}
\label{sec:clean_noisy}
In this subsection, we analyze the impact of different clean/noisy label partition strategies by introducing two additional threshold selection variants: \textit{Mean} and \textit{Median}, as summarized in Table~\ref{tab:clean_noisy}. These strategies are designed to determine the threshold for distinguishing label-clean and label-noisy samples based on the distances to their corresponding hyperbolic class prototypes.
Specifically, for the \textit{Mean} variant, we compute the statistical mean of the distances between samples and their class-specific hyperbolic prototype within a given domain. This mean value is then used as the decision threshold: samples with distances below the threshold are regarded as clean, while those above are considered noisy. The \textit{Median} variant follows the same procedure but replaces the mean with the statistical median, which is generally more robust to outliers.
We evaluate the effectiveness of these thresholding strategies by measuring the accuracy of label noise classification across three different training domains, \textit{i.e.}, \textit{Art Painting} (A), \textit{Cartoon} (C), and \textit{Sketch} (S), using the checkpoints which have best validation performances, and the test domain is selected as \textit{Photo}. This evaluation helps to assess how accurately the model can identify noisy labels during training, which is crucial to improve robustness in the OSDG-NL setting.

According to the performance comparisons of different thresholding strategies presented in Table~\ref{tab:clean_noisy}, we observe that our proposed thresholding method significantly outperforms the \textit{Mean} and \textit{Median} baselines. Specifically, our method demonstrates consistently higher accuracy in partitioning label-clean and label-noisy samples across all domains and noise settings.
Notably, even under the most severe label noise condition with an $80\%$ noise ratio, our approach achieves promising label noise classification accuracies of $50.59\%$ (A), $45.07\%$ (C), and $49.56\%$ (S), respectively. These results not only exceed the performance of the simple statistical baselines but also surpass the expected partition quality implied by the injected noise ratios.

These results demonstrate the robustness and effectiveness of our thresholding strategy in accurately identifying clean samples, even under severe label corruption. Despite a high proportion of injected label noise, the majority of samples within each class and domain still retain correct labels. Our method capitalizes on this observation by analyzing the distribution of distances to the class-specific hyperbolic prototypes, where the upper bound of the distance range among the majority of clean samples provides a reliable threshold to separate clean from noisy labels.
In contrast, the \textit{Mean} and \textit{Median} strategies compute thresholds based on the entire set of samples for each category and within each domain, without explicitly accounting for the clean-label majority structure within each class-domain combination. As a result, these methods tend to be influenced by noisy outliers, which skew the threshold and reduce their ability to correctly distinguish between clean and noisy samples. This lack of adaptability to the underlying distribution of clean data leads to their inferior performance in the label partitioning task. Our proposed method, by leveraging class- and domain-aware distance distributions, achieves more accurate and noise-resilient partitioning, which is especially beneficial in the context of open-set domain generalization with noisy labels.

\subsection{Analysis on Different Textual Encoders for Asymmetric Label Noise Generation}
We provide further experimental comparison between different textual encoders, CLIP~\cite{radford2021learning} and BERT~\cite{BERT}, for asymmetric label noise generation in Table~\ref{tab:clip_text}, with a fixed label noise ratio of $50\%$. Asymmetric label noise is generated by computing the cosine similarity between visual features and category name embeddings extracted from the respective textual encoder. The likelihood of one class being mislabeled as another is guided by this semantic similarity, introducing a more realistic and structured noise pattern compared to random corruption.

From the results, we observe that HyProMeta achieves an OSCR of $43.33\%$ when trained on data perturbed by BERT~\cite{BERT}-based asymmetric label noise. In contrast, it reaches $44.49\%$ OSCR when trained with CLIP~\cite{radford2021learning}-based noise, demonstrating that CLIP~\cite{radford2021learning}-based asymmetric label noise generation considers more category alignment from the visual perspective, which alleviates the challenges brought by the label perturbation. This performance gain can be attributed to CLIP~\cite{radford2021learning}'s training paradigm, which aligns visual and textual modalities using a large-scale image-text paired corpus. As a result, CLIP~\cite{radford2021learning} produces more discriminative and semantically aligned textual embeddings for class names, leading to less challenging label noise patterns.
The difference in performance also reflects how the structure of the label noise influences the training dynamics under noisy supervision. While both BERT~\cite{BERT} and CLIP~\cite{radford2021learning} provide contextual embeddings, BERT~\cite{BERT} operates primarily in a pure language space without direct alignment to visual semantics, which may introduce less visually grounded noise transitions. In contrast, CLIP~\cite{radford2021learning}'s joint vision-language pretraining offers embeddings that are more semantically consistent with visual representations, making the resulting noisy labels more amenable to correction and robust feature learning.

However, we could observe that HyProMeta still yields state-of-the-art performances under asymmetric label noise generated from different textual encoders.
These results further support the generalizability of HyProMeta under diverse and challenging asymmetric label noise configurations in the open-set domain generalization setting.

\subsection{Analysis on the Module Ablations}
\label{sec:abl}
The ablation study of the individual components in our HyProMeta framework is presented in Table~\ref{tab:ablation_study_component}. This study is conducted on the PACS dataset across all label noise settings, with ResNet18~\cite{he2016deep} used as the feature learning backbone. 
Two HyProMeta variants are considered in this analysis: \textit{w/o HYB-Meta} and \textit{w/o NCA-Prompt}. 
The \textit{w/o HYB-Meta} variant incorporates only the new category-aware learnable prompt into the training pipeline, whereas the \textit{w/o NCA-Prompt} variant utilizes only the hyperbolic prototype-based label noise-agnostic meta-learning.

The results indicate that removing the hyperbolic prototype-based label noise-agnostic meta-learning (\textit{HYB-Meta}) results in more significant performance degradation compared to removing the new category-aware learnable prompt (\textit{NCA-Prompt}). Specifically, by employing \textit{HYB-Meta}, HyProMeta achieves performance improvements of $7.58\%$, $6.13\%$, $10.07\%$, and $3.57\%$ in terms of OSCR for symmetric label noise ratios ranging from $20\%$ to $80\%$, as well as for an asymmetric label noise ratio of $50\%$ on PACS. These results demonstrate the efficacy of label noise-aware meta-learning guided by hyperbolic prototypes in addressing the challenges of OSDG-NL.

Similarly, the inclusion of \textit{NCA-Prompt} in HyProMeta delivers performance gains of $6.22\%$, $4.77\%$, $6.54\%$, and $2.38\%$ in terms of OSCR under the same noise ratio settings. These improvements underscore the significance of \textit{NCA-Prompt} in facilitating the learning of more generalizable embeddings, which in turn enhance the performance of \textit{HYB-Meta} during the label noise-agnostic learning process.

Overall, the results highlight the complementary roles of \textit{NCA-Prompt} and \textit{HYB-Meta}, where their integration in HyProMeta achieves superior performance. The \textit{NCA-Prompt} contributes to the generalizability of embeddings, thereby augmenting the effectiveness of \textit{HYB-Meta} in handling label noise in the OSDG-NL task.

\subsection{Analysis on t-SNE Visualizations}
\label{sec:tsne}
We provide t-SNE~\cite{van2008visualizing} visualization results on the PACS dataset, comparing our proposed approach with three baseline methods: \textit{BadLabel}~\cite{zhang2024badlabel}, \textit{MLDG}~\cite{li2018learning}, and \textit{MEDIC}~\cite{wang2023generalizable}. The visualizations of the learned embeddings for the test domain \textit{Photo} are shown in Figure~\ref{fig:bad_label_photo_tsne} to Figure~\ref{fig:ours_photo_tsne}, while those for the test domain \textit{Art Painting} are provided in Figure~\ref{fig:bad_label_art} to Figure~\ref{fig:ours_art}.

Our observations reveal that HyProMeta generates more discriminative embeddings compared to the baselines, particularly in distinguishing between unseen categories (represented by red dots) and seen categories (represented by blue dots). This enhanced capability to learn discriminative embeddings is a significant factor contributing to the performance improvements achieved by our method, as demonstrated in the benchmark analysis section.

\subsection{Analysis on the Confidence Score}
\label{sec:conf}
We further analyze the confidence scores of samples from seen and unseen categories to investigate the advantages of our proposed approach in addressing the challenging OSDG-NL task. Comparisons between the leveraged baselines and our method are presented to illustrate the improvements brought by our technique.

\begin{figure*}[!t]
\centering
\begin{subfigure}[b]{0.23\textwidth}
    \centering
    \includegraphics[width=\textwidth]{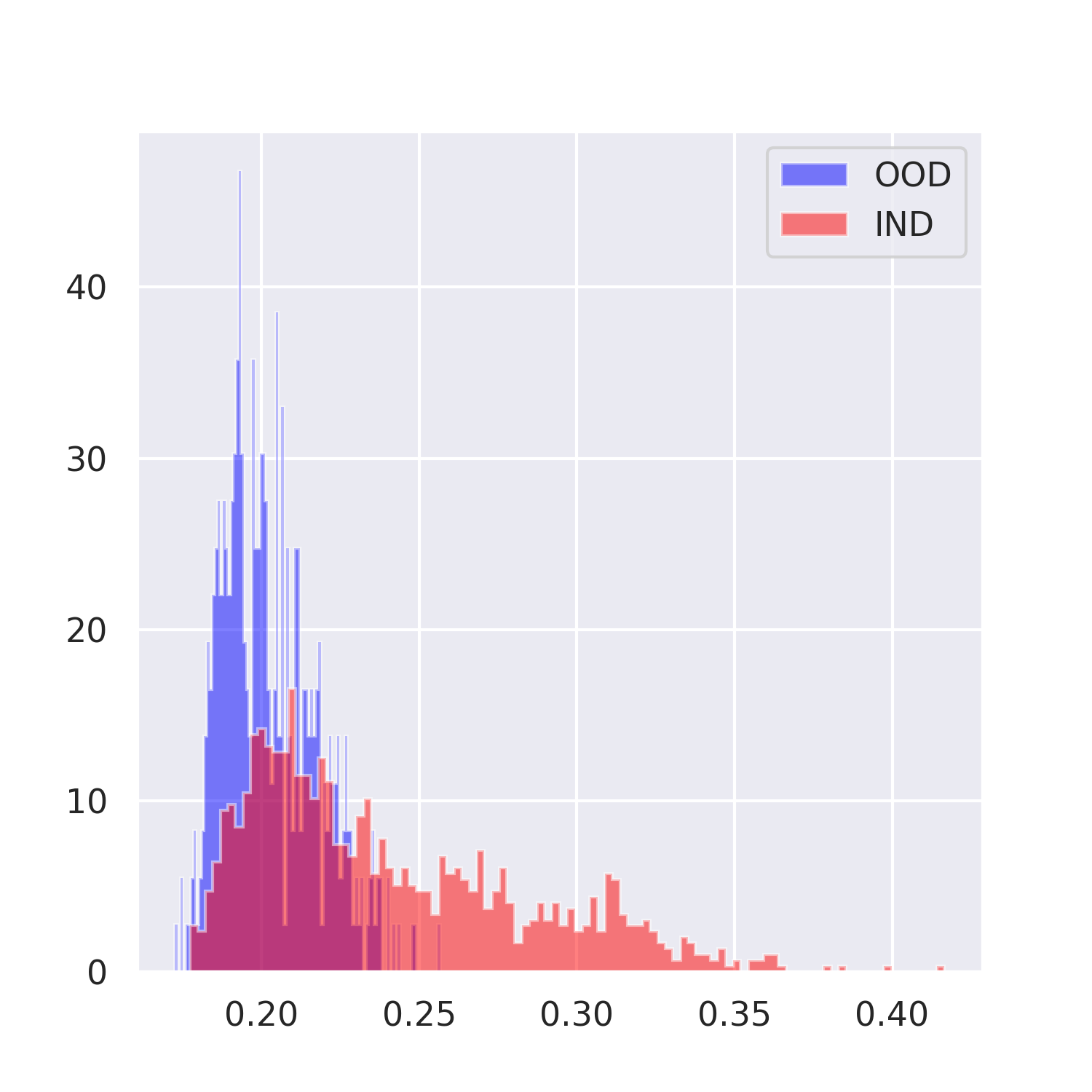}
    \caption{BadLabel~\cite{zhang2024badlabel}, D:Photo, LNSR:50\% sym}
    \label{fig:bad_label_photo}
\end{subfigure}
\hfill
\begin{subfigure}[b]{0.23\textwidth}
    \centering
    \includegraphics[width=\textwidth]{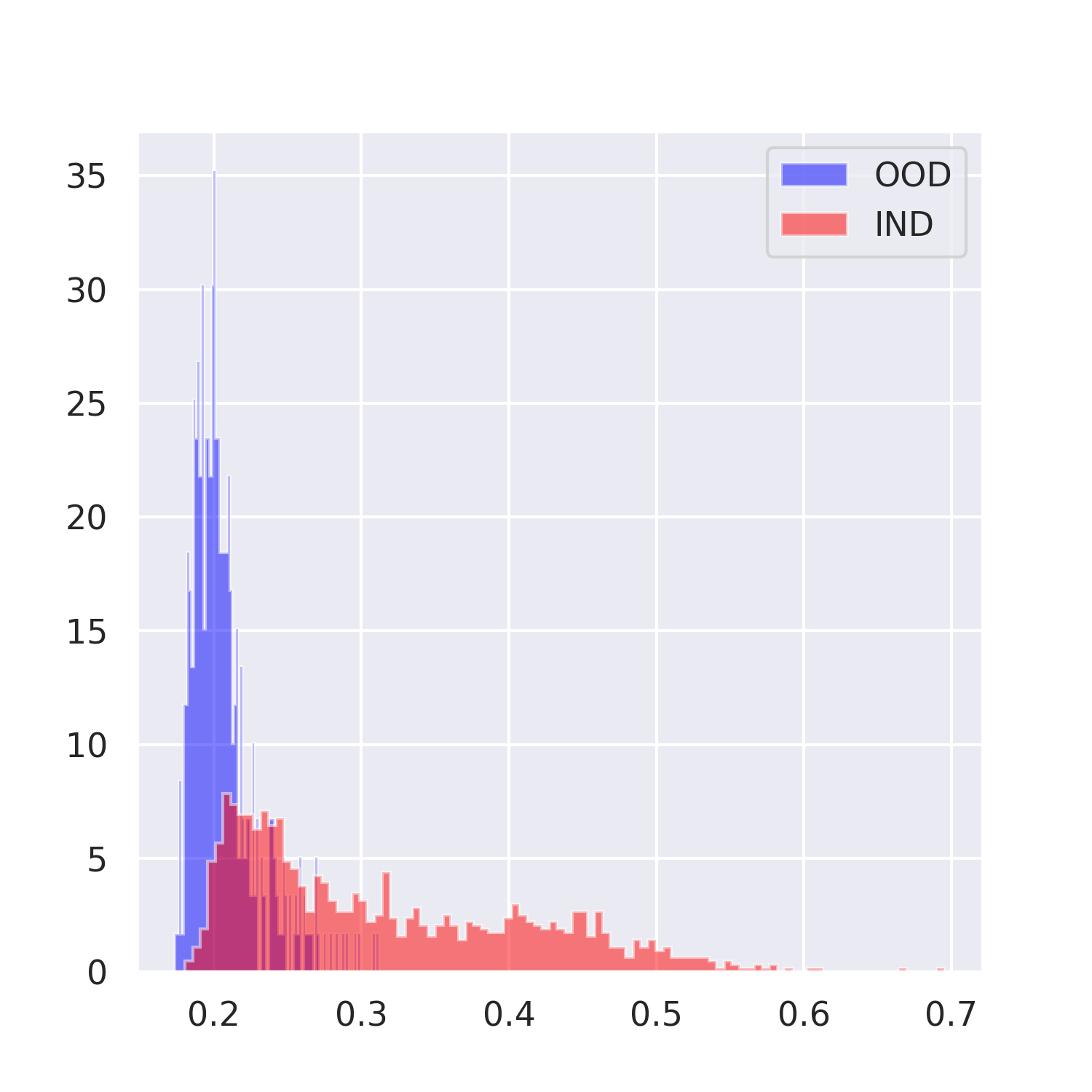}
    \caption{MLDG~\cite{shu2019meta}, D:Photo, LNSR:50\% sym}
    \label{fig:MLDG_photo}
\end{subfigure}
\hfill
\begin{subfigure}[b]{0.23\textwidth}
    \centering
    \includegraphics[width=\textwidth]{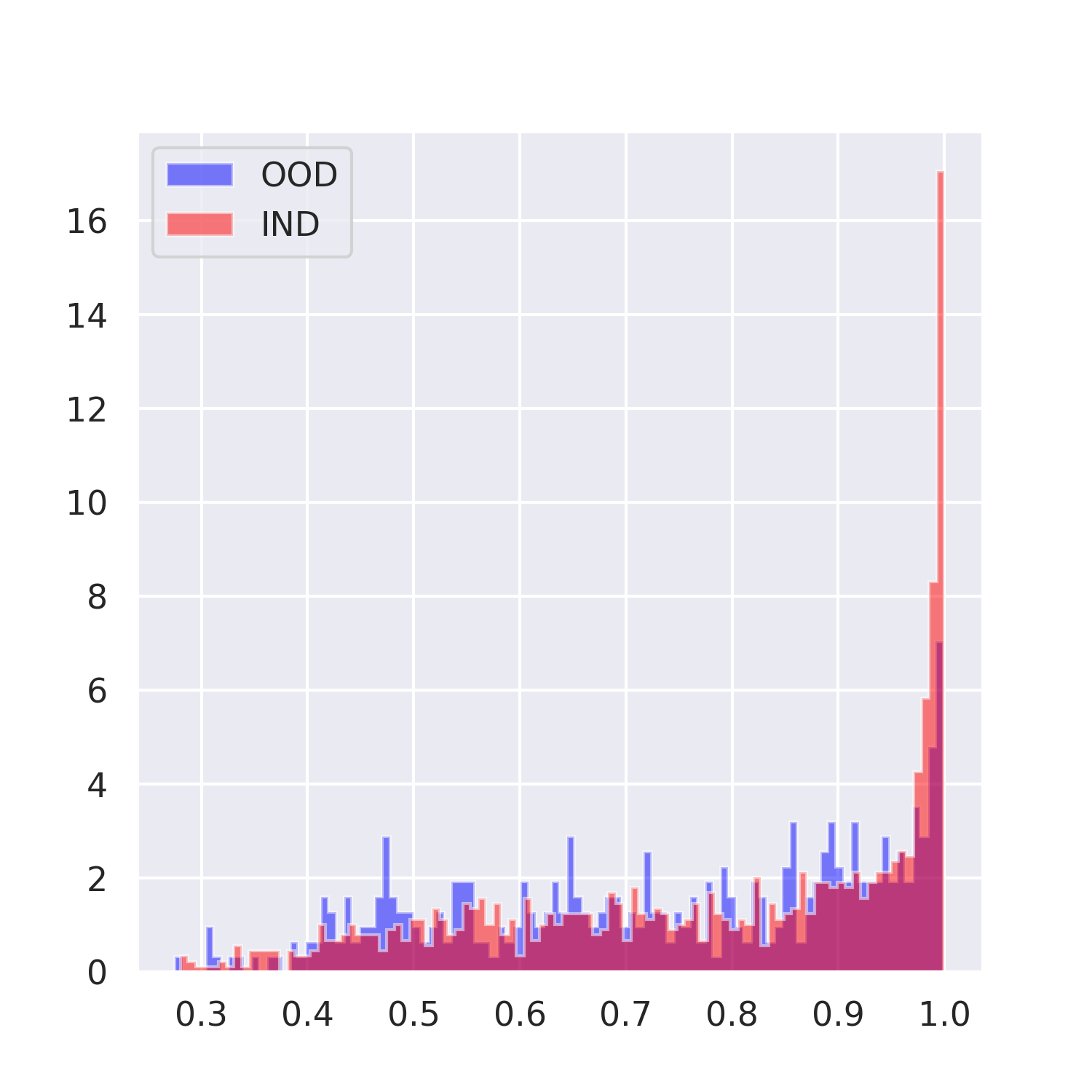}
    \caption{MEDIC~\cite{wang2023generalizable}, D:Photo, LNSR:50\% sym}
    \label{fig:medic_photo}
\end{subfigure}
\hfill
\begin{subfigure}[b]{0.23\textwidth}
    \centering
    \includegraphics[width=\textwidth]{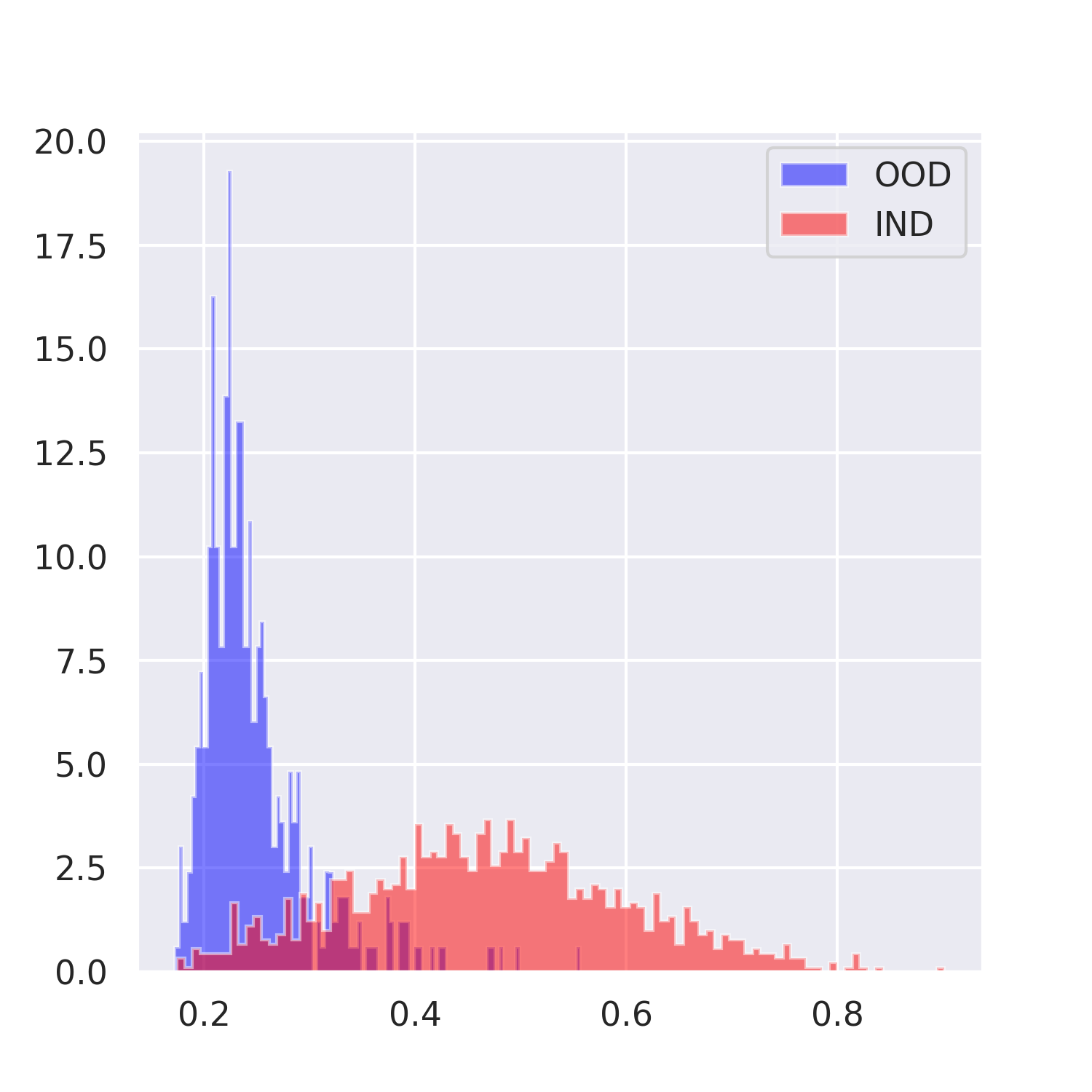}
    \caption{HyProMeta (Ours), D:Photo, LNSR:50\% sym}
    \label{fig:ours_photo}
\end{subfigure}
\hfill
\begin{subfigure}[b]{0.23\textwidth}
    \centering
    \includegraphics[width=\textwidth]{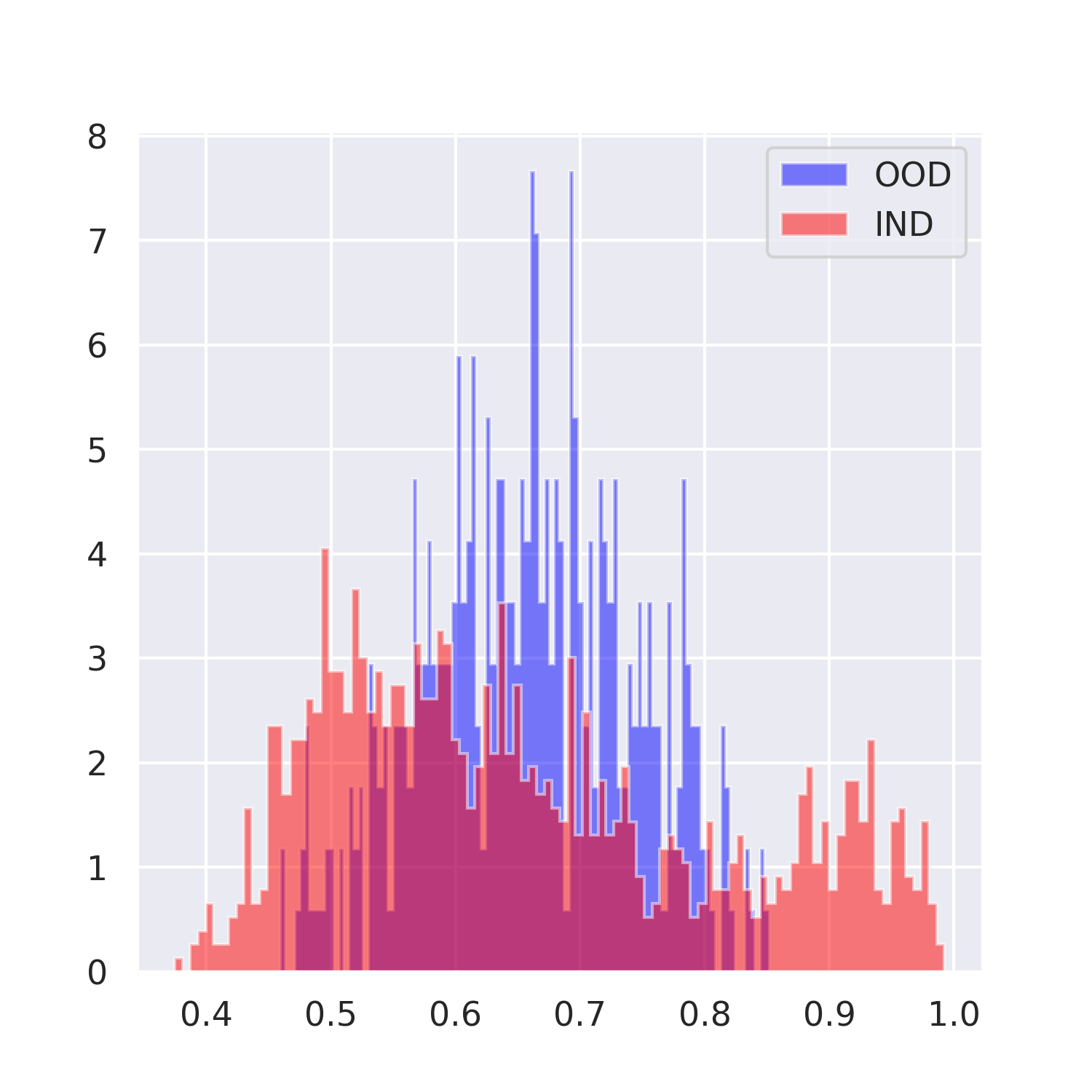}
    \caption{BadLabel~\cite{zhang2024badlabel}, D:Photo, LNSR:50\% asym}
    \label{fig:bad_label_photo_a}
\end{subfigure}
\hfill
\begin{subfigure}[b]{0.23\textwidth}
    \centering
    \includegraphics[width=\textwidth]{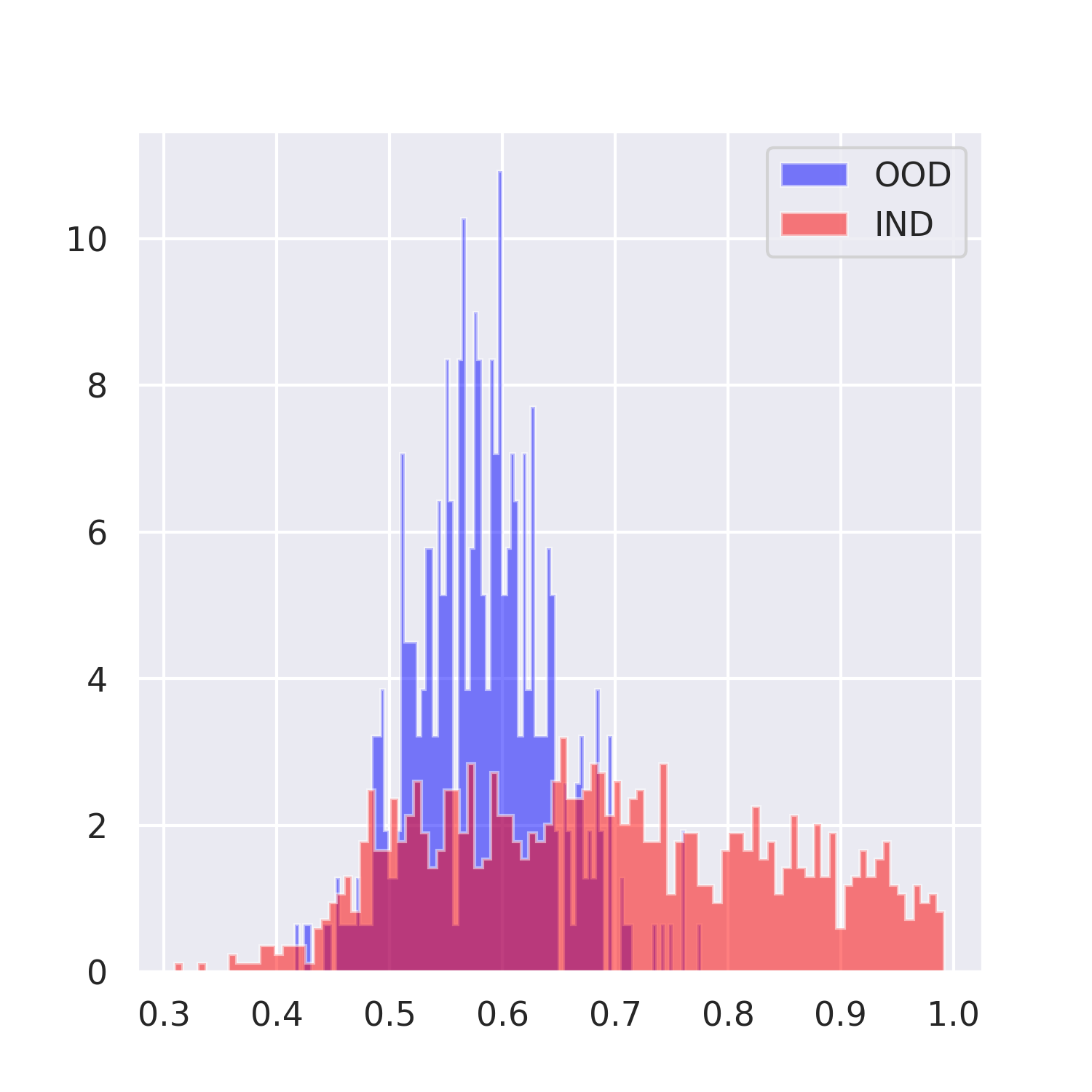}
    \caption{MLDG~\cite{shu2019meta}, D:Photo, LNSR:50\% asym}
    \label{fig:MLDG_photo_a}
\end{subfigure}
\hfill
\begin{subfigure}[b]{0.23\textwidth}
    \centering
    \includegraphics[width=\textwidth]{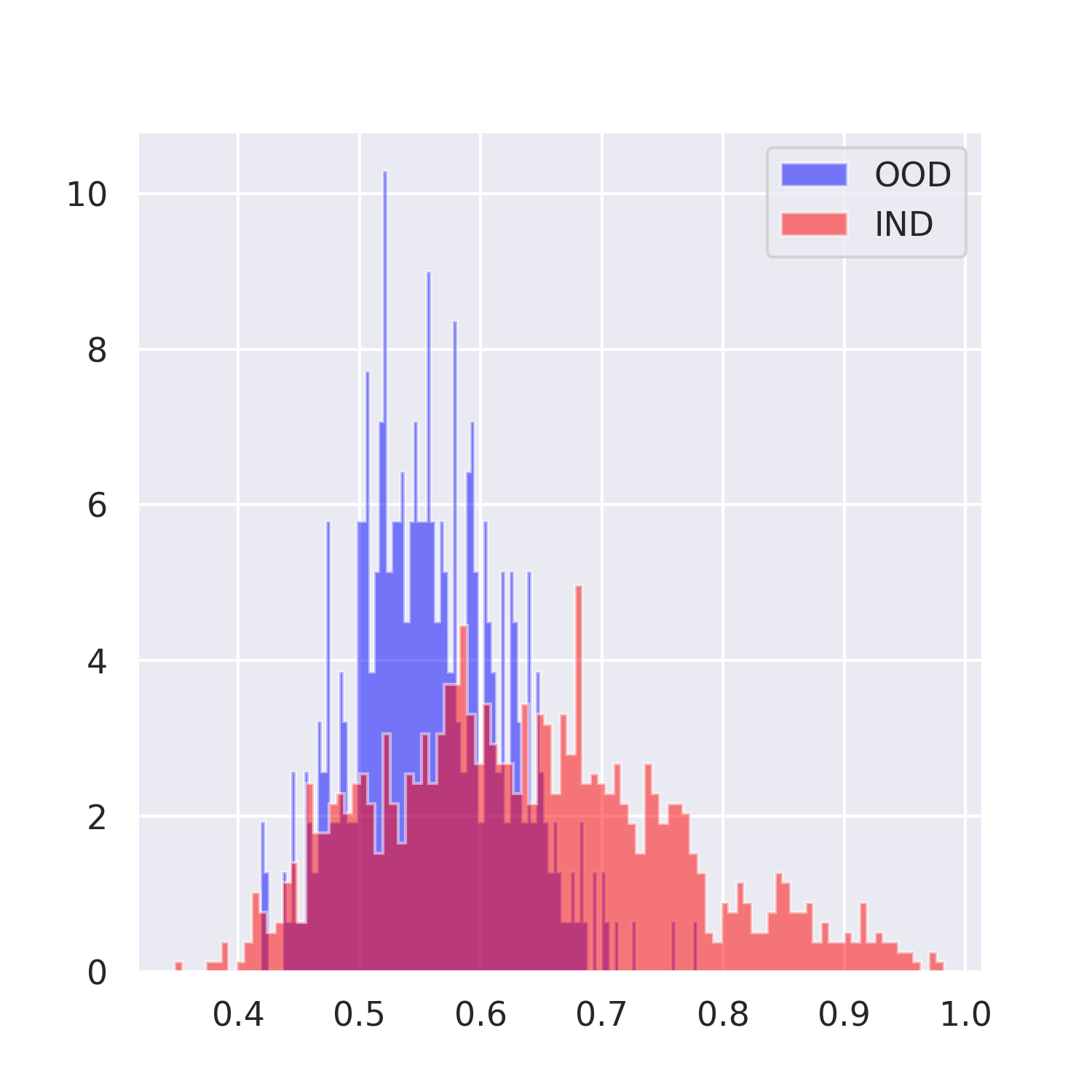}
    \caption{MEDIC~\cite{wang2023generalizable}, D:Photo, LNSR:50\% asym}
    \label{fig:medic_photo_a}
\end{subfigure}
\hfill
\begin{subfigure}[b]{0.23\textwidth}
    \centering
    \includegraphics[width=\textwidth]{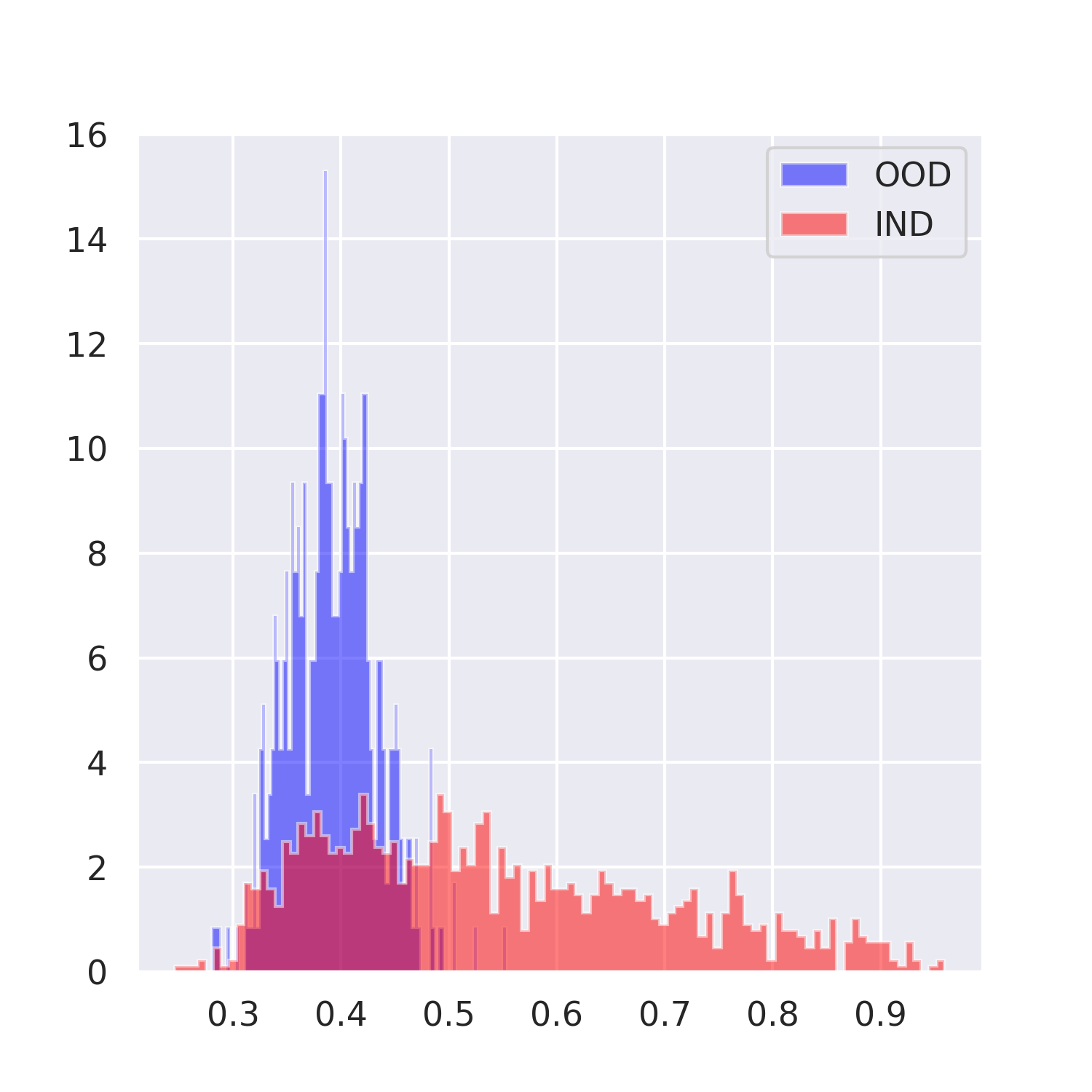}
    \caption{HyProMeta (Ours), D:Photo, LNSR:50\% asym}
    \label{fig:ours_photo_a}
\end{subfigure}

\caption{Confidence score visualization of learned representations on PACS using ResNet18~\cite{he2016deep} under symmetric label noise with a ratio of $50\%$ and asymmetric label noise with a ratio of $50\%$. D indicates the target domain and LNSR indicates the label noise setting.}

\label{fig:conf_score}
\end{figure*}

The visualizations of the confidence score distributions for seen and unseen categories from the test domain \textit{Photo} are shown in Figure~\ref{fig:bad_label_photo} to Figure~\ref{fig:ours_photo} for the symmetric label noise ratio of $50\%$, and in Figure~\ref{fig:bad_label_photo_a} to Figure~\ref{fig:ours_photo_a} for the asymmetric label noise ratio of $50\%$. In these figures, the distributions of confidence scores for seen categories are represented in red, while those for unseen categories are shown in blue.

Our observations indicate that the confidence score distributions obtained by HyProMeta exhibit greater separation between seen and unseen categories. This separation highlights a key advantage of our proposed approach, as it enables the model to achieve a strong awareness of unseen categories and to assign them low confidence scores in the target domain. These findings further demonstrate the effectiveness of HyProMeta in handling the complexities of the OSDG-NL task.

\subsection{Ablation of the NCA-Prompt}
\label{sec:nca}
We compare our proposed NCA-Prompt approach with several alternative variants: \textit{Asynchronous Opt.}, \textit{Adversarial Opt.}, and \textit{Fixed Crop.}. These comparisons are presented in Table~\ref{tab:abl_nca}, with experiments conducted on the ResNet18~\cite{he2016deep} backbone using the PACS dataset under a symmetric label noise ratio of $50\%$.

In the \textit{Asynchronous Opt.} variant, the main network and the prompt is updated alternately, with the main network being trained first while keeping the prompt frozen, followed by updating the prompt with the main network frozen. In the \textit{Adversarial Opt.} variant, a similar alternating update procedure is applied; however, the prompt is optimized adversarially to mislead the main network into incorrect predictions, increasing the complexity of the learned prompt during training. In the \textit{Fixed Crop.} variant, the crop size is fixed, and the prompt is defined only within the fixed crop size. Joint learning of the prompt and backbone is performed using synchronized optimization.

In contrast, our NCA-Prompt utilizes a synchronous optimization strategy to enable the joint learning of both the main network and the prompt dynamically. This approach achieves superior performance compared to the aforementioned variants, as demonstrated in Table~\ref{tab:abl_nca}. These results validate the effectiveness of our synchronous optimization design in enhancing the learning capability of the proposed NCA-Prompt approach.

\subsection{Ablation of HYB-Meta}
\label{sec:hyb}
The ablation study of the HYB-Meta approach is presented in Table~\ref{tab:abl_hyb}. This study is conducted using ResNet18~\cite{he2016deep} as the backbone on the PACS dataset under a label noise ratio of $50\%$. Three variants are examined in this analysis, specifically \textit{w/o label corr.}, \textit{w/o cross domain}, and \textit{w/ euc. prototype}.

\textit{w/o label corr.} refers to the removal of the hyperbolic prototype-based label correction mechanism in HYB-Meta. Without this correction, our approach demonstrates a $2.53\%$ improvement in OSCR, highlighting the critical role of label correction in the meta-learning process for the OSDG-NL task.
\textit{w/o cross domain} indicates that both the meta-train and meta-test partitions are derived from the same source domain, rather than incorporating cross-domain data. Under this setting, HYB-Meta achieves a $4.98\%$ OSCR improvement, emphasizing the benefits of cross-domain data partitioning in enhancing model performance for the OSDG-NL task.
\textit{w/ euc. prototype} involves replacing the category prototypes computed in hyperbolic space with those computed in Euclidean space. This substitution results in a $3.16\%$ OSCR performance drop, demonstrating the effectiveness of hyperbolic category prototypes in leveraging the unique properties of hyperbolic space for OSDG-NL.

The findings from this ablation study underscore the importance of each design component in HYB-Meta, with each contributing significantly to its overall performance in tackling the challenges of the OSDG-NL task.

\subsection{Ablation Regarding the Step Length of the Prototype Calculation}
\label{sec:hyper}
In this section, we present an ablation study on the step length parameter $N_{epoch}$, which determines the frequency of updating the hyperbolic prototypes and corrected labels during meta-learning. The hyperbolic prototype calculation requires determining the categorical centers of each class within each source domain in hyperbolic space. Performing this operation at every step during meta-learning would be computationally expensive. Therefore, we adopt a fixed step length $N_{epoch}$ for these updates.

Table~\ref{tab:abl_n_e} reports the results for five different values of $N_{epoch}$, specifically $N_{epoch} \in \{500, 1000, 1500, 2000, 2500\}$. From these experiments, we observe that setting $N_{epoch} = 500$ yields the best performance for our proposed HyProMeta approach. Consequently, we use $N_{epoch} = 500$ in all subsequent experiments to balance computational efficiency and performance.

\begin{table}[h!]
\caption{Ablation study of $N_{epoch}$ on PACS with ResNet18~\cite{he2016deep} backbone, where Photo is selected as the test domain.}
\label{tab:abl_n_e}
\centering
\begin{tabular}{llll} 
\toprule
$\mathbf{N}_{epoch}$ & \textbf{Acc}            & \textbf{H-score}        & \textbf{OSCR}            \\ 
\hline
500      & \textbf{66.00} & \textbf{76.84} & \textbf{66.00}  \\
1000     & 63.81          & 73.76          & 62.88           \\
1500     & 64.54          & 73.50          & 63.63           \\
2000     & 65.67          & 75.69          & 64.86           \\
2500     & 65.02          & 75.31          & 64.48           \\
\bottomrule
\end{tabular}
\end{table}

\section{Conclusion}
\label{sec6}
This paper presents HyProMeta, a novel framework addressing the challenges of Open-Set Domain Generalization under Noisy Labels (OSDG-NL), an area that remains both critical and underexplored in deep learning. By incorporating hyperbolic prototypes for label noise-aware meta-learning and a learnable prompt to improve generalization, HyProMeta demonstrates superior performance compared to state-of-the-art methods on newly developed OSDG-NL benchmarks based on the PACS and DigitsDG datasets. Comprehensive experimental evaluations validate the framework’s robustness across varying noise levels, showcasing significant advancements in classification accuracy and recognition of unseen categories. This study not only underscores the limitations of existing OSDG and noisy label learning techniques but also lays a foundation for advancing research into effective noise-handling strategies in open-set environments. 
By providing open-source access to the proposed benchmarks and codebase, this work aims to catalyze further exploration in this domain, fostering the development of models with enhanced reliability in real-world applications. The results affirm the potential of HyProMeta to effectively integrate noisy label learning and open-set recognition, achieving improved robustness and generalization in challenging scenarios.

\noindent\textbf{Future work:} 
In the future, we will explore the OSDG-NL task in other downstream tasks, \textit{e.g.}, video classification, semantic segmentation, and multi-modal learning.

\section*{Acknowledgment}
The project served to prepare the SFB 1574 Circular Factory for the Perpetual Product (project ID: 471687386), approved by the German Research Foundation (DFG, German Research Foundation) with a start date of April 1, 2024. 
This work was also partially supported by the SmartAge project sponsored by the Carl Zeiss Stiftung (P2019-01-003; 2021-2026). 
This work was performed on the HoreKa supercomputer funded by the Ministry of Science, Research and the Arts Baden-Württemberg and by the Federal Ministry of Education and Research. The authors also acknowledge support by the state of Baden-Württemberg through bwHPC and the German Research Foundation (DFG) through grant INST 35/1597-1 FUGG. 
This project was also supported partially by the National Natural Science Foundation of China under Grant No. 62473139, partially by the Hunan Provincial Research and Development Project under Grant No. 2025QK3019, and partially by the State Key Laboratory of Autonomous Intelligent Unmanned Systems (the opening project number  ZZKF2025-2-10).
%

\bibliographystyle{plain}
\bibliography{sn-bibliography}

\end{document}